\documentclass[letterpaper]{article} 
\usepackage{aaai2026}  
\usepackage{times}  
\usepackage{helvet}  
\usepackage{courier}  
\usepackage[hyphens]{url}  
\usepackage{graphicx} 
\usepackage{natbib}  
\usepackage{caption} 
\usepackage{cite}
\usepackage{amsmath,amssymb,amsfonts}
\usepackage{textcomp}
\usepackage{multirow}
\usepackage{subcaption}
\usepackage{enumitem}
\usepackage{url}
\usepackage{amsthm}

\newtheorem{corollary}{Corollary} 
\newtheorem{definition}{Definition} 
\usepackage{booktabs}
\usepackage{float}
\usepackage{algorithm}
\usepackage{algorithmic}
\setlist[itemize]{leftmargin=*}
\usepackage{newfloat}
\usepackage{listings}
\DeclareCaptionStyle{ruled}{labelfont=normalfont,labelsep=colon,strut=off} 
\floatstyle{ruled}
\newfloat{listing}{tb}{lst}{}
\floatname{listing}{Listing}
\nocopyright
\title{PrivacyCD: Hierarchical Unlearning for Protecting Student Privacy in Cognitive Diagnosis
}
\author{
    Mingliang Hou\textsuperscript{\rm 1,3}
    Yinuo Wang\textsuperscript{\rm 1},
    Teng Guo\textsuperscript{\rm 1},
    Zitao Liu\textsuperscript{\rm 1}\thanks{The corresponding author.},
    Wenzhou Dou\textsuperscript{\rm 1,3},
    Jiaqi Zheng\textsuperscript{\rm 1},
    Renqiang Luo\textsuperscript{\rm 2},
    Mi Tian\textsuperscript{\rm 3},
    Weiqi Luo\textsuperscript{\rm 1}
}
\affiliations{
    \textsuperscript{\rm 1} Guangdong Institute of Smart Education, Jinan University, Guangzhou, China\\
    \textsuperscript{\rm 2} College of Computer Science and Technology,Jilin University, Changchun, China\\
    \textsuperscript{\rm 3} TAL Education Group, Beijing, China\\

}

\usepackage{bibentry}

\begin{document}

\maketitle

\begin{abstract}
The need to remove specific student data from cognitive diagnosis (CD) models has become a pressing requirement, driven by users' growing assertion of their ``right to be forgotten''. However, existing CD models are largely designed without privacy considerations and lack effective data unlearning mechanisms. Directly applying general-purpose unlearning algorithms is suboptimal, as they struggle to balance unlearning completeness, model utility, and efficiency when confronted with the unique heterogeneous structure of CD models. To address this, our paper presents the first systematic study of the data unlearning problem for CD models, proposing a novel and efficient algorithm: hierarchical importance-guided forgetting (HIF). Our key insight is that parameter importance in CD models exhibits distinct layer-wise characteristics. HIF leverages this via an innovative smoothing mechanism that combines individual and layer-level importance, enabling a more precise distinction of parameters associated with the data to be unlearned. Experiments on three real-world datasets show that HIF significantly outperforms baselines on key metrics, offering the first effective solution for CD models to respond to user data removal requests and for deploying high-performance, privacy-preserving AI systems.

\end{abstract}


\section{Introduction}
Artificial intelligence (AI) is driving a paradigm shift in education towards personalization, with the ultimate goal of providing tailored learning experiences for each student. Central to this vision, cognitive diagnosis (CD) plays an important role~\cite{survey_cd}. CD aims to precisely infer a student's mastery of latent knowledge components (KCs)\footnote{A KC is a generalization of everyday terms like concept, principle, fact, or skill.} by analyzing their response data. Unlike traditional tests that only assess correctness, CD acts like an experienced ``educational doctor'', providing deep insights into the cognitive structure reflected in a student's response patterns and pinpointing deficiencies in their knowledge base~\cite{neuralcd}. This fine-grained diagnosis is a prerequisite for enabling adaptive learning path planning and personalized resource recommendations, thereby transforming personalized education at scale from an ideal into a tangible reality. Recently, deep learning-based CD models have demonstrated immense potential in various intelligent education systems, owing to their superior diagnostic accuracy.

However, despite continuous performance improvements in CD models, a critical problem has been largely overlooked: the protection of student privacy~\cite{survey_mu1,survey_mu2}. The efficacy of CD models is built upon the in-depth analysis of vast amounts of sensitive student data, which can be indirectly linked to students' personal information, posing severe compliance and security risks~\cite{survey_cd}. This challenge is particularly acute in real-world scenarios. On the one hand, regulations such as the General Data Protection Regulation (GDPR) grant users the ``right to be forgotten'', which requires the complete removal of their personal data~\cite{gdpr}. This requires not only deleting records from databases but also ensuring that the model itself unlearns the associated information. On the other hand, models may face malicious probing from competitors. For example, an attacker could use a membership inference attack (MIA) to determine whether a model was trained on an exclusive high-quality dataset, thus spying on business secrets and identifying the sources of its technical barriers~\cite{survey_mia,mia_bwbox}. These imminent threats converge on a pressing technical challenge: how to enable CD models to forget information efficiently and reliably.

To address these challenges, machine unlearning has emerged as a promising solution. Current techniques fall into two main streams: exact unlearning (e.g., SISA)~\cite{mu_sisa}, which relies on data partitioning and sub-model retraining but suffers from high storage overhead and limited scalability; and approximate unlearning, which seeks a balance between efficiency and effectiveness by directly modifying model parameters~\cite{mu_ga1,mu_fisher1,mu_hessian1}. Within approximate methods, techniques based on the Fisher information matrix (FIM) have become central to the field~\cite{mu_ssd}. The FIM provides an efficient way to quantify parameter importance, enabling researchers to identify and precisely modify parameters associated with the data to be unlearned, thus avoiding the need for a full retraining.

Although approximate unlearning methods based on FIM show great promise, their direct application to modern CD models, especially those based on deep neural networks (e.g., NeuralCDM~\cite{neuralcd}, KaNCD~\cite{kancd}, KSCD~\cite{kscd}), presents new and unique challenges. Specifically, prevailing deep learning-based CD models typically consist of multiple heterogeneous components, such as a \texttt{student embedding layer} to represent student abilities, an \texttt{exercise embedding layer} to characterize question properties, and an \texttt{interactive prediction network} to simulate the response process. Within such a complex architecture, the importance signal for a single parameter as computed by the FIM can become ambiguous and noisy. For instance, a parameter identified by the FIM as ``important'' for a student-to-be-forgotten might be just one dimension of an embedding vector that is also critical for representing the general proficiency of many other students. Indiscriminately suppressing such parameters can easily harm innocent representations, leading to a catastrophic degradation of the model's diagnostic performance on the retained data. Therefore, leveraging the importance of parameters in these fine-grained models is a key bottleneck for achieving efficient unlearning. 

To systematically address these problems, this paper introduces PrivacyCD, a comprehensive privacy-preserving framework designed specifically for the CD domain. First, we formalize the machine unlearning problem in the context of CD, specifying its requirements regarding data, models, and evaluation. Second, we design an efficient and analyzable neural CD architecture that provides a structural foundation for subsequent unlearning operations. At its core, we propose a novel unlearning strategy: hierarchical importance-guided forgetting (HIF), which introduces layer-wise priors to smooth the noise in parameter importance, thereby enabling more robust unlearning. Finally, to ensure the effectiveness of the unlearning process, we devise an MIA protocol tailored for student-item-score triplet interaction data, offering a reliable evaluation standard for unlearning research in this field.

The main contributions of this paper can be summarized as follows:
\begin{itemize}
    \item We present the first systematic study of machine unlearning for CD, proposing PrivacyCD, a comprehensive framework that addresses a critical research gap in student privacy.
    
    \item We design a novel, retrain-free algorithm, HIF, which leverages a ``hierarchical wisdom'' to smooth parameter importance estimations, enabling more robust and precise unlearning for complex CD models.
    
    \item Our extensive experiments on three real-world datasets validate that HIF achieves a state-of-the-art balance among unlearning efficacy, model utility, and efficiency, providing a strong baseline and reliable evaluation standards for the field.
\end{itemize}

\section{Related Work}
\subsection{Cognitive Diagnosis Models}
CD models aim to infer a student's mastery of latent KCs by analyzing their response records~\cite{survey_cd,neuralcd}. Traditional CD models are mostly based on psychometric theories, such as item response theory, focusing on modeling students' cognitive traits from a statistical perspective~\cite{irt1,irt2}. Recently, deep learning has introduced a new paradigm to this field. In particular, the neural cognitive diagnosis (NeuralCD) framework leverages the powerful representation capabilities of deep neural networks to directly learn latent cognitive representations from student response data, achieving significant performance improvements~\cite{neuralcd,kancd,kscd,neuralncd}.

Prevailing neural CD models are often constructed based on the multi-layer perceptron (MLP), including models like NeuralCD~\cite{neuralcd}, KaNCD~\cite{kancd}, and KSCD~\cite{kscd}. These models typically consist of key components such as a student embedding layer, an exercise embedding layer, and an interaction network, and are trained using carefully designed loss functions. Although graph neural network-based CD models have shown excellent performance in modeling the complex relationships within students' knowledge structures, their core architecture usually requires constructing an interaction graph (e.g., a bipartite graph) between student response records and exercises~\cite{graphcd_rcd,graphcd_dmccdm,graphcd_disencd,graphcd_mhcd}. This requirement often leads to a significant decrease in training and deployment efficiency. In consideration of this trade-off, MLP-based CD models are more feasible for practical applications. Therefore, this paper adopts an MLP-based architecture as the basis of our research.

\subsection{Machine Unlearning}
Machine unlearning aims to efficiently and reliably remove the influence of specific data from a trained model to meet privacy protection and data compliance requirements~\cite{survey_mu1,survey_mu2}. Existing techniques have primarily evolved along two paths. Early methods, such as exact unlearning, employ strategies like data sharding and sub-model ensembling (e.g., SISA)~\cite{mu_sisa,mu_scrubbing1,mu_scrubbing2}. While these methods provide strong unlearning guarantees, they often come at the cost of high storage overhead and complex system design, limiting their scalability for large-scale models.

To overcome these limitations, approximate unlearning has emerged, striving to strike a balance between computational efficiency and unlearning effectiveness~\cite{mu_ssd}. This path has seen the development of various technical routes.

Perturbation-based methods corrupt the information of data-to-be-forgotten by injecting noise into the model. For instance, some works employ label flipping, where the labels of samples to be unlearned are changed to random incorrect ones, followed by a few epochs of fine-tuning to confuse the model's memory~\cite{mu_relabel}. Other approaches operate directly at the parameter level, such as using gradient ascent to maximize the loss on the unlearning set, forcing the model to forget these data, or using noise injection to add Gaussian noise to relevant parameters, thereby reducing their signal-to-noise ratio~\cite{mu_ga1,mu_ga2}.

In contrast to cruder perturbations, a more refined approach is to precisely calculate and remove the contribution of the data-to-be-forgotten to the model parameters. A significant milestone in this direction was the introduction of the FIM, which can effectively quantify the importance of each parameter with respect to specific data points~\cite{mu_fisher1,mu_fisher2}. FIM-based methods can identify and precisely modify or attenuate parameters strongly associated with the unlearning data without full retraining~\cite{mu_ispf,mu_ssd}. Due to their theoretical grounding and efficiency, they represent one of the most advanced and actively researched directions in the field. Despite significant progress in machine unlearning for data such as images and text, its application to CD remains largely unexplored.
\begin{figure}[t]
\centering
\includegraphics[width=0.9\columnwidth]{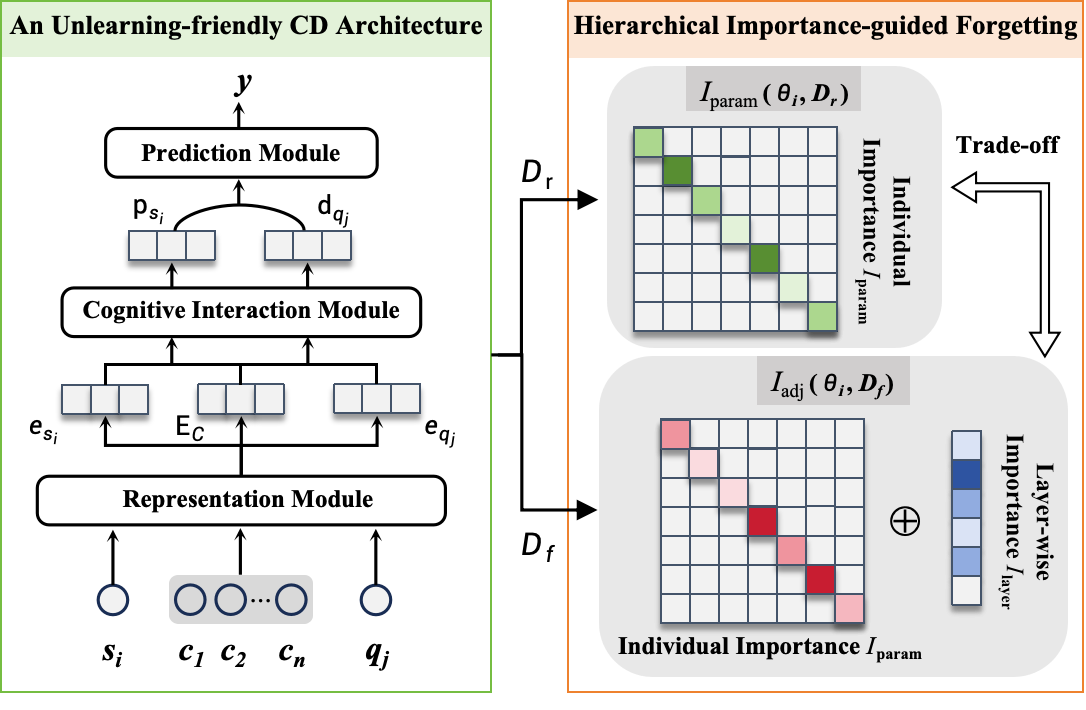} 
\caption{The overall framework of PrivacyCD.}
\label{framework}
\end{figure}
\section{The framework of PrivacyCD}
This section details our PrivacyCD framework. We first formalize the unlearning problem for CD, then present our unlearning-friendly architecture and the core HIF algorithm, concluding with its theoretical analysis.
\subsection{Preliminaries and Problem Formulation}
\begin{definition}
\textbf{Cognitive Diagnosis}: CD aims to infer a student's mastery status over a predefined set of KCs by analyzing their response records. We formalize it as follows: let $\mathcal{S}=\{s_1, ..., s_I\}$ be the set of students, $\mathcal{Q}=\{q_1, ..., q_J\}$ be the set of exercises, and $\mathcal{C}=\{c_1, ..., c_K\}$ be the set of KCs. The relationship between exercises and concepts is described by an expert-annotated Q-matrix, $\mathbf{Q} \in \{0, 1\}^{J \times K}$, where $\mathbf{Q}_{jk}=1$ indicates that mastering KC $c_k$ is required to correctly answer exercise $q_j$. A CD model, $M$, is a function $f_\theta$ parameterized by $\theta$ that takes a student $s_i$ and an exercise $q_j$ as input to predict the probability of a correct response, $p(r_{ij}=1 | s_i, q_j; \theta)$. The model parameters $\theta$ contain representations for each student's proficiency vector across all KCs, $\boldsymbol{\theta}_{s_i} \in \mathbb{R}^K$, and each exercise's attributes (e.g., difficulty, discrimination), $\boldsymbol{\theta}_{q_j}$.
\end{definition}

\begin{definition}
\textbf{Machine Unlearning}: The goal of machine unlearning is to permanently remove the influence of a specific data subset (the forget set, $\mathcal{D}_f$) from a pre-trained model $M_{orig}$ while preserving the knowledge learned from the remaining data (the retain set, $\mathcal{D}_r$). The core objective is to efficiently generate an unlearned model, $M_{unlearn}$, that behaves similarly to a model retrained from scratch on the retain set, $M_{retrain}$, but at a significantly lower computational cost.
\end{definition}

\paragraph{Problem Formulation:}
In the context of CD, we formulate the unlearning problem as follows. Given a complete dataset of response records $\mathcal{D}_{all}$, a Q-matrix $\mathbf{Q}$, and an original CD model $M_{orig}$ (with parameters $\theta_{orig}$) trained on $\mathcal{D}_{all}$. Upon receiving a request to forget all data from a subset of students $\mathcal{S}_f$, we define the forget set $\mathcal{D}_f$ and the retain set $\mathcal{D}_r$. Our goal is to design an unlearning algorithm, $\mathcal{A}_{unlearn}$, that takes $\theta_{orig}$ and $\mathcal{D}_f$ as input to efficiently generate a new set of parameters, $\theta_{unlearn}$. A successful unlearning algorithm must satisfy both \textbf{effectiveness} and \textbf{efficiency}: its computational complexity must be lower than retraining from scratch, and the resulting model $M_{unlearn}$ must not only maintain diagnostic performance on the retained data comparable to the gold-standard model $M_{retrain}$, but also be resilient to MIA, such that an attacker cannot distinguish forgotten data from unseen data.

\subsection{A Neural Architecture for Cognitive Diagnosis}
\label{cd-arc}
To validate the effectiveness of our unlearning algorithm, we select an advanced neural network-based CD architecture as our experimental testbed. The core principle behind our choice is its potential for being ``unlearning-friendly''. The architecture adheres to principles of modularity and representation decoupling, which clearly separate parameters strongly associated with specific individuals (student embeddings) from those that describe general cognitive patterns (interaction networks). This decoupled nature provides an ideal foundation for precise, importance-based unlearning operations, as shown in the left of Figure \ref{framework}.

The selected architecture consists of three main modules. A representation module learns independent embedding vectors for students ($\mathbf{e}_{s_i}$), exercises ($\mathbf{e}_{q_j}$), and KCs ($\mathbf{E}_C$). A cognitive interaction module then dynamically computes a student's proficiency vector $\mathbf{p}_{s_i}$ and an exercise's difficulty vector $\mathbf{d}_{q_j}$ via interaction functions (e.g., $\phi_{\text{prof}}$). Finally, a prediction module integrates this information, calculates a ``cognitive gap'' using the Q-matrix, $\mathbf{g}_{ij} = (\mathbf{p}_{s_i} - \mathbf{d}_{q_j}) \odot \mathbf{Q}_{j,:}$, and feeds it into a shared feed-forward network (FFN) to produce the final prediction probability. This design localizes a student's personal information primarily within their corresponding embedding vector $\mathbf{e}_{s_i}$, providing a clear target for our HIF algorithm.

\subsection{Hierarchical Importance-guided Forgetting}

Building upon the neural CD architecture previously described, this section details our core unlearning algorithm: HIF (hierarchical importance-guided forgetting).

To precisely remove the influence of specific student data without compromising the model's diagnostic capabilities, we first need a tool to quantify memory. In neural networks, the strength of a parameter's memory of specific data can be measured by the impact this data has on the parameter's updates. Theoretically, this is precisely captured by the second-order derivatives of the loss function with respect to the parameters—the Hessian matrix. However, computing the full Hessian for a deep CD model is computationally infeasible. We, therefore, introduce the FIM as an efficient and effective approximation of the Hessian. Near a local minimum of the loss function, the FIM is equivalent to the expectation of the Hessian ($\mathbb{E}[\mathbf{H}] = \mathbf{F}$). Critically, the diagonal elements of the FIM, which represent the importance of each parameter, can be efficiently computed using first-order gradients. Specifically, for a model parameter $\theta_i$ and a dataset $\mathcal{D}$, its importance $I(\theta_i, \mathcal{D})$ is defined as the expected squared gradient of the loss function over the dataset:
\begin{equation}
    I(\theta_i, \mathcal{D}) = \mathbb{E}_{x \in \mathcal{D}} \left[ \left( \frac{\partial \ell(\theta; x)}{\partial \theta_i} \right)^2 \right]
\end{equation}
This provides a practical tool to measure the sensitivity of each parameter to the forget set $\mathcal{D}_f$ and the retain set $\mathcal{D}_r$, respectively.

Despite the utility of the FIM for quantifying parameter importance, its direct application to CD models faces the core challenge of unstable individual importance estimates. Due to training noise, a single parameter can exhibit a spuriously high importance value, leading naive unlearning methods to harm the model's generalizable knowledge. To address this, we propose the HIF algorithm, as shown in the right of Figure \ref{framework}. Our key insight is to leverage the hierarchical structure of parameter importance within CD models. HIF employs an innovative smoothing mechanism that combines the unstable individual parameter importance (a microscopic signal) with the more robust collective layer-wise importance (a macroscopic signal). This hierarchical wisdom yields a more accurate importance metric, enabling efficient unlearning while maximally preserving the model's overall diagnostic utility.

First, we calculate the individual importance ($I_{\text{param}}$) for each parameter $\theta_i$, which is its FIM value concerning the forget set $\mathcal{D}_f$ and the retain set $\mathcal{D}_r$. This reflects the micro-level association strength of each parameter with the data.

However, micro-level information alone is insufficient. We then compute the layer-wise importance ($I_{\text{layer}}$). For each functional layer $L_j$ in the model (e.g., the student embedding layer or the first layer of the interaction network), we calculate the collective average importance of all its parameters with respect to the forget set $\mathcal{D}_f$:
\begin{equation}
    I_{\text{layer}}(L_j, \mathcal{D}_f) = \frac{1}{|L_j|} \sum_{\theta_i \in L_j} I_{\text{param}}(\theta_i, \mathcal{D}_f),
\end{equation}
where $|L_j|$ is the number of parameters in layer $j$. This macroscopic metric offers greater statistical stability and reflects the overall sensitivity of an entire functional module (e.g., the student proficiency representation module) to the data being forgotten.

Next, we perform importance smoothing to correct the microscopic individual estimations with the macroscopic hierarchical wisdom. Using a smoothing hyperparameter $\beta \in [0, 1]$, we create a weighted fusion of the individual and layer-wise importances to obtain a more robust adjusted importance, $I_{\text{adj}}$:
\begin{equation}
    I_{\text{adj}}(\theta_i, \mathcal{D}_f) = (1 - \beta) \cdot I_{\text{param}}(\theta_i, \mathcal{D}_f) + \beta \cdot I_{\text{layer}}(L_j, \mathcal{D}_f).
\end{equation}
This smoothing mechanism effectively suppresses anomalous spikes in individual parameter importance caused by training noise, while preserving the genuine and significant unlearning signals reflected collectively by the entire functional layer.

Finally, based on this more reliable importance metric, we proceed with selective parameter attenuation. We identify parameters that are over-specialized to the forget set using a threshold hyperparameter $\alpha$, selecting those that satisfy the condition $I_{\text{adj}}(\theta_i, \mathcal{D}_f) > \alpha \cdot I_{\text{param}}(\theta_i, \mathcal{D}_r)$. For these selected parameters, we attenuate them precisely according to their adjusted relative importance, controlled by an unlearning strength hyperparameter $\lambda$:
\begin{equation}
    \theta_i' \leftarrow \theta_i \cdot \left(1 - \lambda \cdot \min\left(\frac{I_{\text{adj}}(\theta_i, \mathcal{D}_f)}{I_{\text{param}}(\theta_i, \mathcal{D}_r)}, 1\right)\right).
\end{equation}
\subsection{Theoretical Analysis of HIF}
HIF's hierarchical smoothing mechanism is grounded in established statistical principles. We frame importance estimation as a statistical inference problem, where the empirical importance $I_{\text{param}}$ is a noisy estimate of the true importance $\mu$. HIF's smoothing technique is analogous to a James-Stein shrinkage estimator~\cite{js1,js2}, which is designed to reduce this estimation noise. This connection leads to the following corollary, theoretically substantiating HIF's advantages (see \textbf{Appendix A} for the detailed proof).

\begin{corollary}[Superiority of HIF's Importance Estimation]
\label{cor:hif_superiority}
\textit{Under standard assumptions of estimation noise within a functional layer of a CD model (e.g., the student embedding layer), the total mean squared error of the adjusted importance estimates produced by HIF is lower than that of the naive, individual parameter importance estimates, provided a suitable smoothing factor $\beta \in (0, 1)$ is chosen. That is:}
\begin{equation}
    \sum_{i \in L_j} \mathbb{E}[(I_{\text{adj}}(\theta_i) - \mu_i)^2] < \sum_{i \in L_j} \mathbb{E}[(I_{\text{param}}(\theta_i) - \mu_i)^2]
\end{equation}
\end{corollary}

This corollary implies that HIF's importance estimation is statistically more accurate. A more accurate estimation leads to more reliable decisions in identifying which parameters to modify, thereby reducing two critical types of errors: (1) mistakenly altering parameters crucial for general knowledge (false positives), and (2) failing to modify parameters that have indeed memorized the forgotten data (false negatives). Consequently, HIF is theoretically better positioned to strike an optimal balance between unlearning effectiveness and the preservation of model utility.
\begin{table}[ht]
\centering
\renewcommand{\arraystretch}{1.2} 
\begin{tabular}{ccccc}
\hline
\textbf{Dataset} & \textbf{\#Student} & \textbf{\#Question} & \textbf{\#Response} & \textbf{\#KC} \\ \hline
Math1            & 4,209              & 20                  & 84,180              & 11            \\
Math2            & 3,911              & 20                  & 78,220              & 16            \\
Frcsub           & 536                & 20                  & 10,720              & 8             \\ \hline
\end{tabular}
\caption{Statistics of the datasets used in our experiments.}
\label{dataset}
\end{table}

\section{Experiments}
In this section, we conduct a series of extensive experiments to comprehensively evaluate the effectiveness of our proposed PrivacyCD framework and its core HIF algorithm. 

\subsection{Experimental Settings}
\subsubsection{Datasets}We evaluate our proposed method on three public, real-world educational datasets: Math1, Math2, and FrcSub~\cite{survey_cd}. Math1 and Math2 are derived from two large-scale high school mathematics exams, covering a broad range of KCs in algebra and geometry. FrcSub focuses on the cognitive skills of middle school students in the specific domain of fraction subtraction. All datasets include student response records and the corresponding Q-matrix information. The detailed statistics for these three datasets are presented in Table \ref{dataset}. A more comprehensive description of each dataset can be found in \textbf{Appendix B.1}.

\subsubsection{Baselines}To comprehensively evaluate the performance of our proposed PrivacyCD, we compare it against a series of representative baseline methods as described below:
\begin{itemize}
    \item \textbf{Original Model \& Retrain Model:} Both models are based on the neural CD architecture introduced in Section \ref{cd-arc}. The original model, trained on the entire dataset, serves as the upper bound for model utility on the retain set. The retrain model, trained from scratch solely on the retain set, acts as the \textbf{gold standard} for both unlearning effectiveness and model utility.

    \item \textbf{Hessian-based Unlearning (Hessian):} An approximate unlearning method based on second-order information. This approach efficiently estimates the diagonal of the Hessian matrix using the Hutchinson algorithm and leverages it as a measure of parameter importance to guide the unlearning process~\cite{mu_hessian1,mu_hessian2}.

    \item \textbf{Gradient Ascent (GradAsc):} A perturbation-based approximate unlearning algorithm. It actively corrupts the model's memory of the forget set by performing gradient ascent to maximize the loss on these specific data points~\cite{mu_ga1,mu_ga2}.

    \item \textbf{FIM-based Unlearning (FIM):} This represents a major class of state-of-the-art approximate unlearning methods based on the FIM. The core idea of this class of methods is to utilize the diagonal of the FIM to estimate the importance of individual parameters with respect to the data to be forgotten, and then use this importance to guide parameter modifications~\cite{mu_fisher1,mu_fisher2,mu_ssd}.
\end{itemize}
\begin{table*}[ht]
\centering
\renewcommand{\arraystretch}{1.4}
\resizebox{\textwidth}{!}{%
\begin{tabular}{cc|ccccc|ccccc|ccccc}
\hline
\multicolumn{2}{c|}{\textbf{Dataset}} &
  \multicolumn{5}{c|}{\textbf{Math1}} &
  \multicolumn{5}{c|}{\textbf{Math2}} &
  \multicolumn{5}{c}{\textbf{Frcsub}} \\ \hline
\multicolumn{1}{c|}{\multirow{2}{*}{\textbf{Unlearning Ratio}}} &
  \multirow{2}{*}{\textbf{Model}} &
  \multicolumn{2}{c}{\textbf{Utility}} &
  \multicolumn{2}{c}{\textbf{MIA}} &
  \multirow{2}{*}{\textbf{\begin{tabular}[c]{@{}c@{}}RTRR\end{tabular}}} &
  \multicolumn{2}{c}{\textbf{Utility}} &
  \multicolumn{2}{c}{\textbf{MIA}} &
  \multicolumn{1}{l|}{\multirow{2}{*}{\textbf{\begin{tabular}[c]{@{}l@{}}RTRR\end{tabular}}}} &
  \multicolumn{2}{c}{\textbf{Time}} &
  \multicolumn{2}{c}{\textbf{MIA}} &
  \multicolumn{1}{l}{\multirow{2}{*}{\textbf{\begin{tabular}[c]{@{}l@{}}RTRR\end{tabular}}}} \\ \cline{3-6} \cline{8-11} \cline{13-16}
\multicolumn{1}{c|}{} &
   &
  \textbf{AUC} &
  \textbf{ACC} &
  \textbf{MIA\_AUC} &
  \textbf{MIA\_ACC} &
   &
  \textbf{AUC} &
  \textbf{ACC} &
  \textbf{MIA\_AUC} &
  \textbf{MIA\_ACC} &
  \multicolumn{1}{l|}{} &
  \textbf{Rate} &
  \textbf{ACC} &
  \textbf{MIA\_AUC} &
  \textbf{MIA\_ACC} &
  \multicolumn{1}{l}{} \\ \hline
\multicolumn{1}{c|}{\multirow{6}{*}{\textbf{10\%}}} &
  M\_orig &
  0.7828 &
  0.7126 &
  0.7731 &
  0.7025 &
  * &
  0.7946 &
  0.7169 &
  0.603 &
  0.5686 &
  * &
  0.8644 &
  0.8004 &
  0.7604 &
  0.6989 &
  * \\
\multicolumn{1}{c|}{} &
  M\_retrain &
  * &
  * &
  \textit{0.4902} &
  \textit{0.4889} &
  * &
  * &
  * &
  \textit{0.5069} &
  \textit{0.5039} &
  * &
  * &
  * &
  \textit{0.5138} &
  \textit{0.5077} &
  * \\
\multicolumn{1}{c|}{} &
  Hessian &
  0.6972 &
  0.6267 &
  0.6282 &
  0.5747 &
  \textbf{91.00\%} &
  0.6517 &
  0.5636 &
  0.5006 &
  0.5059 &
  \textbf{94.77\%} &
  0.8413 &
  0.7824 &
  \underline{0.4954} &
  \underline{0.5011} &
  \underline{96.04\%} \\
\multicolumn{1}{c|}{} &
  GradAsc &
  0.7821 &
  0.6913 &
  0.7487 &
  0.6854 &
  \underline{90.06\%} &
  \underline{0.7947} &
  \underline{0.714} &
  0.5992 &
  0.5678 &
  \underline{96.38\%} &
  \textbf{0.8629} &
  \textbf{0.7979} &
  0.7328 &
  0.6725 &
  \textbf{97.44\%} \\
\multicolumn{1}{c|}{} &
  FIM &
  \textbf{0.7827} &
  \textbf{0.7126} &
  \underline{0.5864} &
  \underline{0.5708} &
  75.88\% &
  \textbf{0.7946} &
  \textbf{0.7169} &
  \underline{0.5112} &
  \underline{0.5031} &
  92.57\% &
  0.8577 &
  0.793 &
  0.3798 &
  0.4505 &
  93.53\% \\
\multicolumn{1}{c|}{} &
  PravicyCD &
  \underline{0.7826} &
  \underline{0.712} &
  \textbf{0.5591} &
  \textbf{0.5731} &
  74.67\% &
  \textbf{0.7946} &
  \textbf{0.7169} &
  \textbf{0.5063} &
  \textbf{0.5004} &
  92.54\% &
  \underline{0.8589} &
  \underline{0.8004} &
  \textbf{0.5172} &
  \textbf{0.4967} &
  93.51\% \\ \hline
\multicolumn{1}{c|}{\multirow{6}{*}{\textbf{5\%}}} &
  M\_orig &
  0.7869 &
  0.7139 &
  0.7826 &
  0.7146 &
  * &
  0.7984 &
  0.718 &
  0.5984 &
  0.5657 &
  * &
  0.8633 &
  0.8009 &
  0.7268 &
  0.6323 &
  * \\
\multicolumn{1}{c|}{} &
  M\_retrain &
  * &
  * &
  \textit{0.4878} &
  \textit{0.4774} &
  * &
  * &
  * &
  \textit{0.5027} &
  \textit{0.5067} &
  * &
  * &
  * &
  \textit{0.5322} &
  \textit{0.5471} &
  * \\
\multicolumn{1}{c|}{} &
  Hessian &
  0.6814 &
  0.6335 &
  0.616 &
  0.5855 &
  \underline{92.53\%} &
  0.6648 &
  0.6235 &
  0.4901 &
  0.4618 &
  \underline{96.02\%} &
  0.7755 &
  0.6573 &
  \textbf{0.5667} &
  \textbf{0.435} &
  \underline{97.47\%} \\
\multicolumn{1}{c|}{} &
  GradAsc &
  0.7867 &
  0.7048 &
  0.7424 &
  0.6843 &
  \textbf{92.67\%} &
  \underline{0.7984} &
  \underline{0.7175} &
  0.584 &
  0.5492 &
  \textbf{97.73\%} &
  \textbf{0.8627} &
  \textbf{0.8091} &
  0.6997 &
  0.6368 &
  \textbf{98.15\%} \\
\multicolumn{1}{c|}{} &
  FIM &
  \underline{0.787} &
  \underline{0.7139} &
  \underline{0.5653} &
  \underline{0.5529} &
  77.12\% &
  0.7936 &
  0.7138 &
  \underline{0.5355} &
  \underline{0.5114} &
  93.84\% &
  \underline{0.8565} &
  \underline{0.8019} &
  0.3961 &
  0.435 &
  94.19\% \\
\multicolumn{1}{c|}{} &
  PravicyCD &
  \textbf{0.7869} &
  \textbf{0.7138} &
  \textbf{0.5435} &
  \textbf{0.5301} &
  76.45\% &
  \textbf{0.7984} &
  \textbf{0.718} &
  \textbf{0.5041} &
  \textbf{0.4996} &
  93.98\% &
  0.793 &
  0.6415 &
  \underline{0.538} &
  \underline{0.3767} &
  94.29\% \\ \hline
\multicolumn{1}{c|}{\multirow{6}{*}{\textbf{1\%}}} &
  M\_orig &
  0.7883 &
  0.7164 &
  0.6574 &
  0.626 &
  * &
  0.798 &
  0.7168 &
  0.5647 &
  0.5582 &
  * &
  0.8613 &
  0.8042 &
  0.8511 &
  0.7381 &
  * \\
\multicolumn{1}{c|}{} &
  M\_retrain &
  * &
  * &
  \textit{0.4743} &
  \textit{0.5039} &
  * &
  * &
  * &
  \textit{0.4745} &
  \textit{0.506} &
  * &
  * &
  * &
  \textit{0.4875} &
  \textit{0.5476} &
  * \\
\multicolumn{1}{c|}{} &
  Hessian &
  0.7066 &
  0.6616 &
  \textbf{0.4918} &
  \textbf{0.4921} &
  \underline{83.90\%} &
  0.5611 &
  0.5653 &
  0.51 &
  0.4849 &
  \underline{96.48\%} &
  0.8019 &
  0.7269 &
  \underline{0.5} &
  \underline{0.5114} &
  \underline{95.36\%} \\
\multicolumn{1}{c|}{} &
  GradAsc &
  \underline{0.7883} &
  \underline{0.7171} &
  0.6222 &
  0.6024 &
  \textbf{93.53\%} &
  \underline{0.798} &
  \underline{0.7169} &
  0.554 &
  0.5422 &
  \textbf{97.53\%} &
  \textbf{0.8613} &
  \textbf{0.8042} &
  0.8511 &
  0.7381 &
  \textbf{98.26\%} \\
\multicolumn{1}{c|}{} &
  FIM &
  \textbf{0.7883} &
  \textbf{0.7164} &
  0.5273 &
  0.5709 &
  74.51\% &
  0.7695 &
  0.7024 &
  \underline{0.4763} &
  \underline{0.4739} &
  92.49\% &
  \underline{0.8454} &
  \underline{0.7805} &
  \textbf{0.4648} &
  \textbf{0.5476} &
  93.18\% \\
\multicolumn{1}{c|}{} &
  PravicyCD &
  \textbf{0.7883} &
  \textbf{0.7164} &
 \underline{0.486} &
  \underline{0.5354} &
  74.02\% &
  \textbf{0.798} &
  \textbf{0.7168} &
  \textbf{0.4749} &
  \textbf{0.506} &
  92.49\% &
  0.8428 &
  0.7877 &
  0.5727 &
  0.5714 &
  93.16\% \\ \hline
\end{tabular}%
}
\caption{Overall performance comparison on three datasets with three unlearning ratios. Results for the Retrain model are shown in \textit{italics} and serve as the gold standard for unlearning. For the MIA\_AUC and MIA\_ACC metrics, results closer to the italicized values are better. The best-performing results are marked in \textbf{bold}, and the second-best are \underline{underlined}. The AUC, ACC, and RTRR of the retrain and original models are not meaningful for comparison and are thus marked with an asterisk (*).}
\label{overallres}
\end{table*}
\subsubsection{Evaluation Metrics}We comprehensively evaluate all methods from three perspectives: model utility, unlearning efficacy, and efficiency.
\begin{itemize}
    \item \textbf{Model Utility:} We use the area under the curve (AUC) and accuracy (ACC), the standard metrics in CD studies~\cite{survey_cd,neuralcd,kancd,kscd,graphcd_rcd}, to evaluate the diagnostic performance retained on the retain set's test data after unlearning.

    \item \textbf{Unlearning Efficacy:} We use MIA to quantify the thoroughness of the unlearning process~\cite{survey_mia,survey_mu1,survey_mu2}. The core mechanism of an MIA is to train an attack classifier that attempts to determine whether a given data point was part of a model's training set by analyzing the model's predictive behavior on that point~\cite{mia_bwbox,mia_bc1}. If unlearning is successful, the model's behavior on a forgotten (member) data point should be indistinguishable from its behavior on an unseen (non-member) data point. We have designed a specific MIA protocol for CD data; the complete procedure is detailed in \textbf{Appendix B.2}. We use \textbf{MIA\_AUC} and \textbf{MIA\_ACC} as the primary evaluation metrics. The closer these values are to the level of the retrain model (the gold standard), the better the unlearning effect.

    \item \textbf{Efficiency:} We measure efficiency using the relative time reduction rate (RTRR), calculated as $(1 - \frac{T_{\text{unlearn}}}{T_{\text{retrain}}}) \times 100\%$, which quantifies the time saved compared to a full retraining.
\end{itemize}
The full reproducible code for our experiments is publicly available on GitHub\footnote{\url{https://anonymous.4open.science/r/privacycd-8783}}. For more implementation details, please refer to \textbf{Appendix B.3}.
\subsection{Overall Performance Comparison}
The overall performance comparison across all experimental settings is presented in Table~\ref{overallres}. The analysis clearly indicates that our proposed PrivacyCD framework achieves the strongest overall performance in balancing model utility, unlearning efficacy, and efficiency. Across all datasets and unlearning ratios, PrivacyCD not only incurs the minimal performance loss in model utility (AUC/ACC) but also achieves unlearning efficacy (MIA metrics) closest to the gold-standard M\_retrain among all approximate methods. In particular, the effectiveness of the HIF is validated by directly comparing PrivacyCD with the naive FIM method. For instance, on the Math2 dataset with a 5\% unlearning ratio, PrivacyCD achieves a substantially better MIA\_AUC (0.5041) than FIM (0.5355)—bringing it closer to the gold standard (0.5027)—while maintaining similar model utility. This confirms that our core design enables more reliable unlearning without compromising performance.

Furthermore, all approximate methods demonstrate significant computational efficiency gains over full retraining (all RTRR $>$ 74\%), highlighting their practical value. However, the baselines exhibit clear shortcomings: GradAsc suffers from poor unlearning efficacy, the Hessian-based method severely degrades model utility, and while FIM serves as a strong baseline, its overall performance is consistently surpassed by PrivacyCD. These comparisons collectively validate that PrivacyCD is the most effective solution to date for the unlearning problem in this domain.
\begin{figure*}[ht]
		\centering
		\begin{minipage}[t]{0.31\textwidth}
			\centering
			\includegraphics[width=5.5cm]{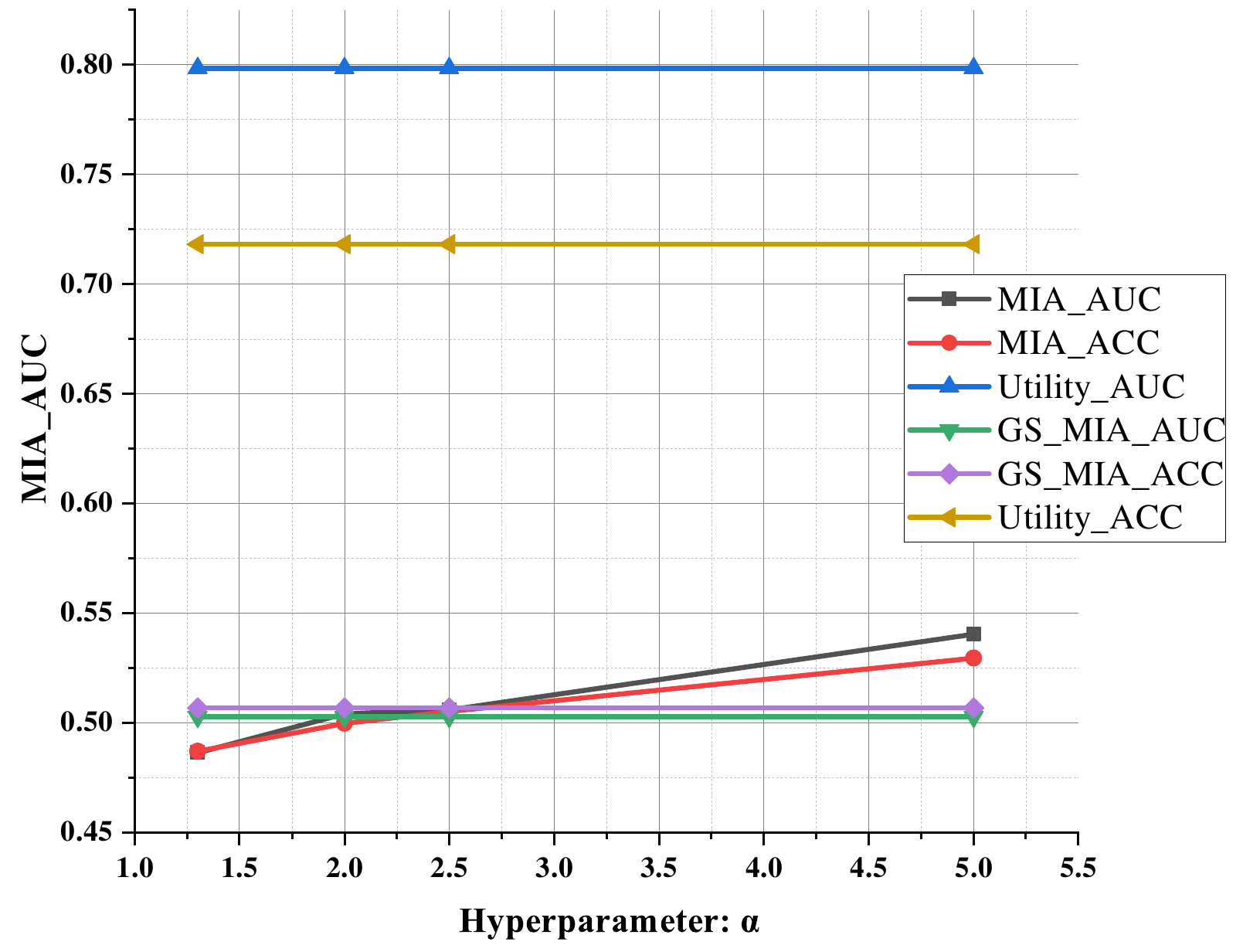}
			\subcaption{Sensitivity to $\alpha$}
		\end{minipage}
		\begin{minipage}[t]{0.31\textwidth}
			\centering
			\includegraphics[width=5.5cm]{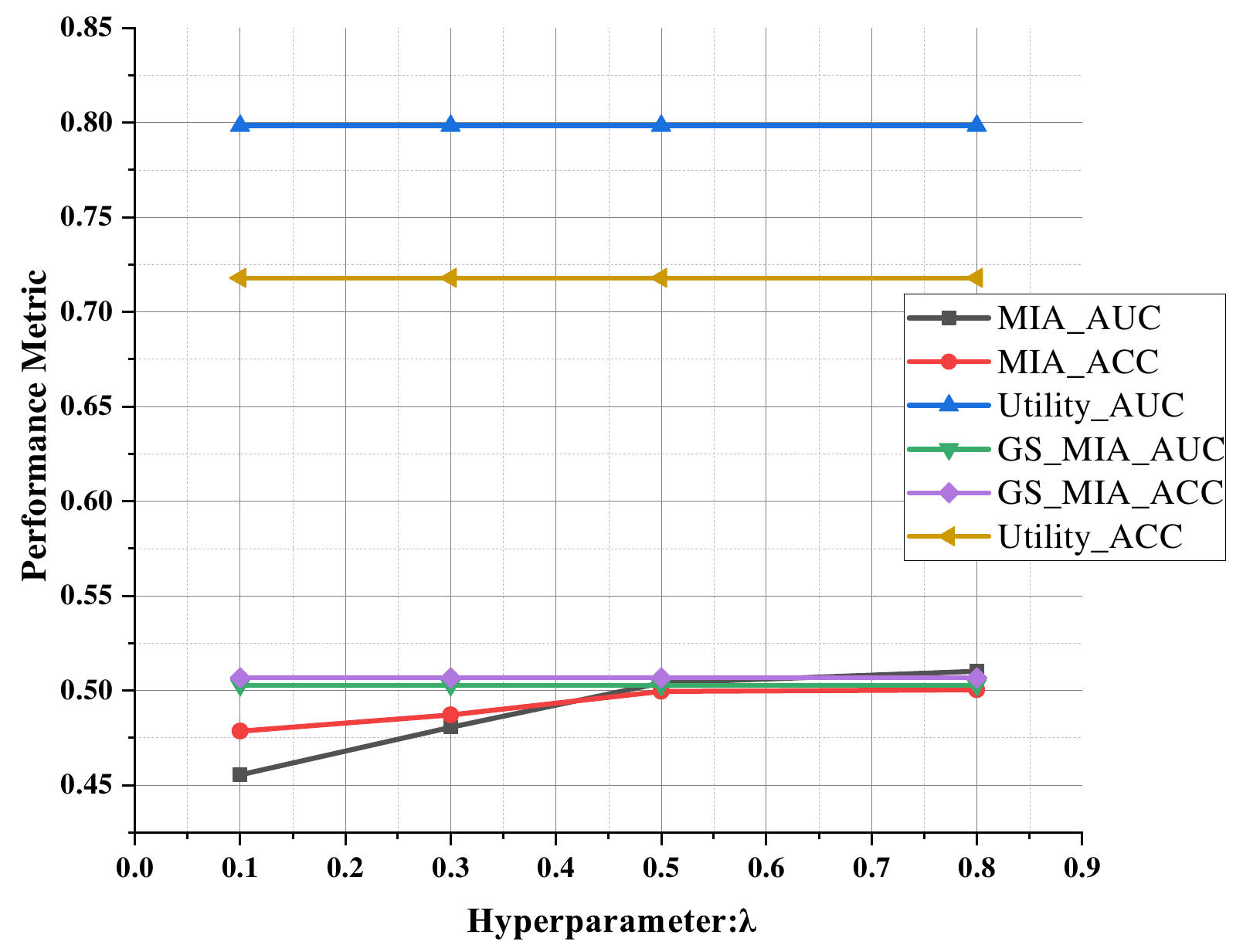}
			\subcaption{Sensitivity to $\lambda$}
		\end{minipage}
		\begin{minipage}[t]{0.31\textwidth}
			\centering
			\includegraphics[width=5.5cm]{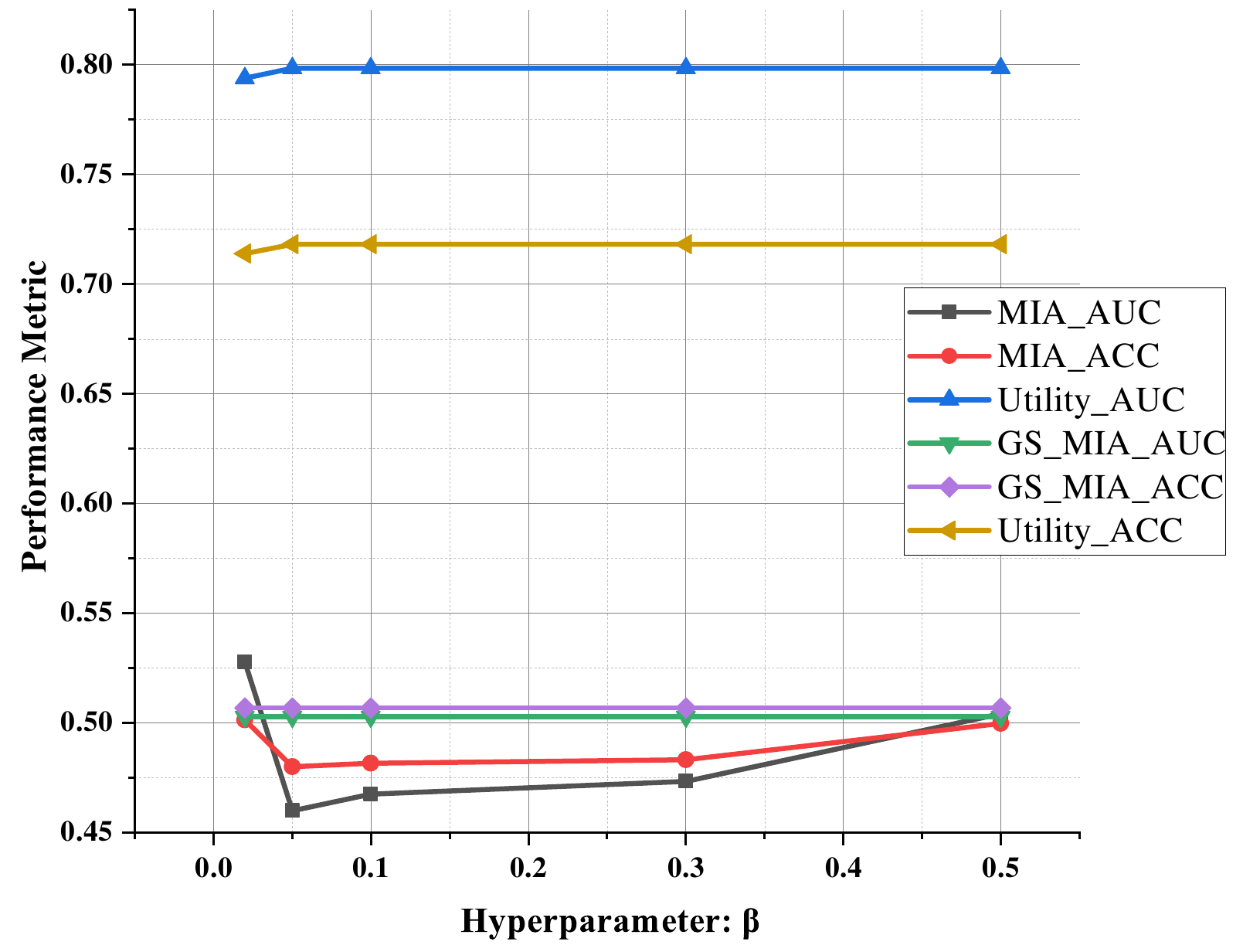}
			\subcaption{Sensitivity to $\beta$}
		\end{minipage}
		\caption{Parameter sensitivity analysis of our HIF algorithm on the Math2 dataset with a 5\% unlearning ratio. We vary one hyperparameter while keeping the others fixed at their optimal values, observing the impact on model utility (AUC, ACC) and unlearning efficacy (MIA\_AUC, MIA\_ACC).}
		\label{fig:sensitivity_analysis}
\end{figure*}
\begin{figure*}[ht]
		\centering
		\begin{minipage}[t]{0.31\textwidth}
			\centering
			\includegraphics[width=5.5cm]{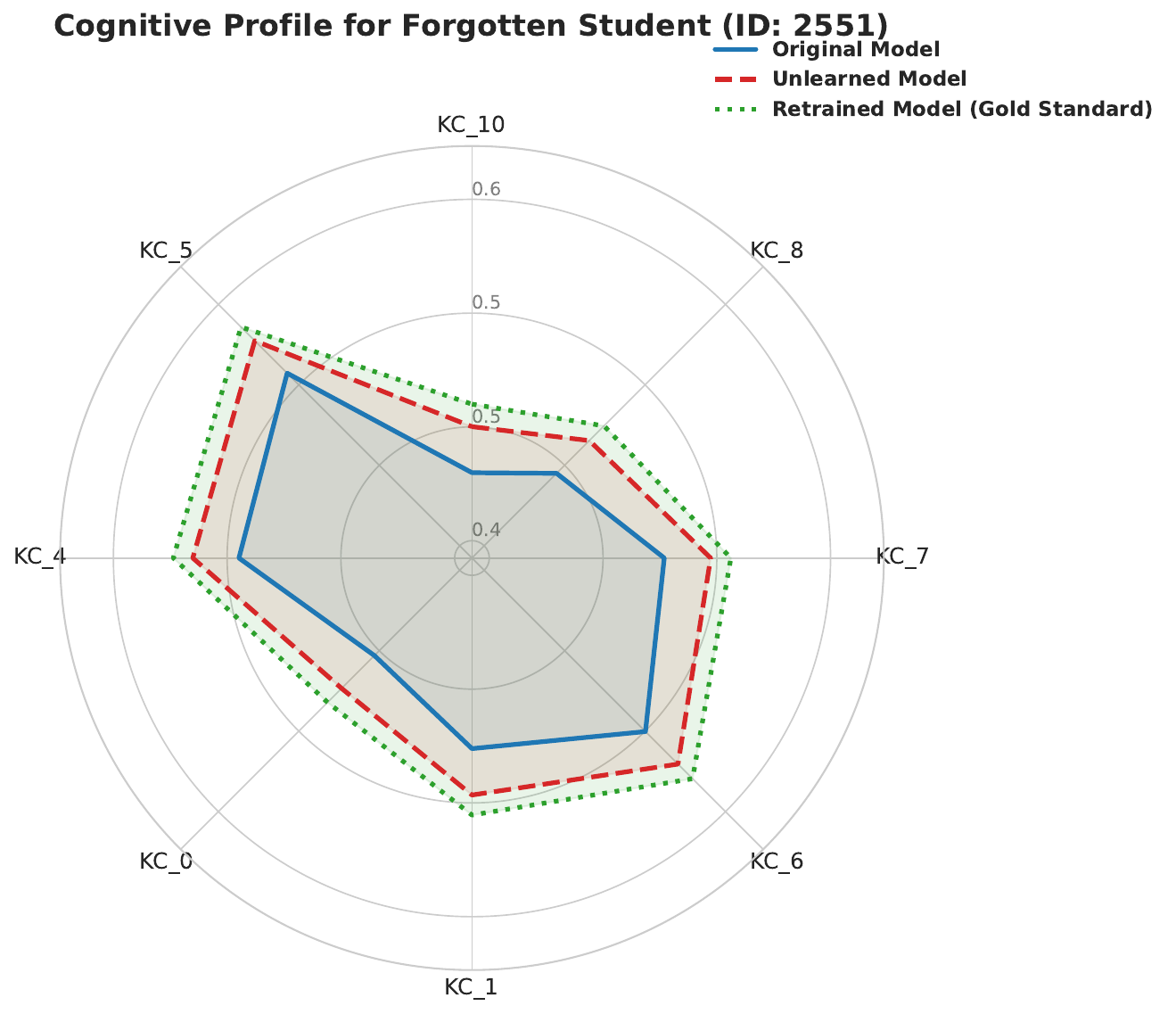}
			\subcaption{Student ID: 2551}
		\end{minipage}
		\begin{minipage}[t]{0.31\textwidth}
			\centering
			\includegraphics[width=5.5cm]{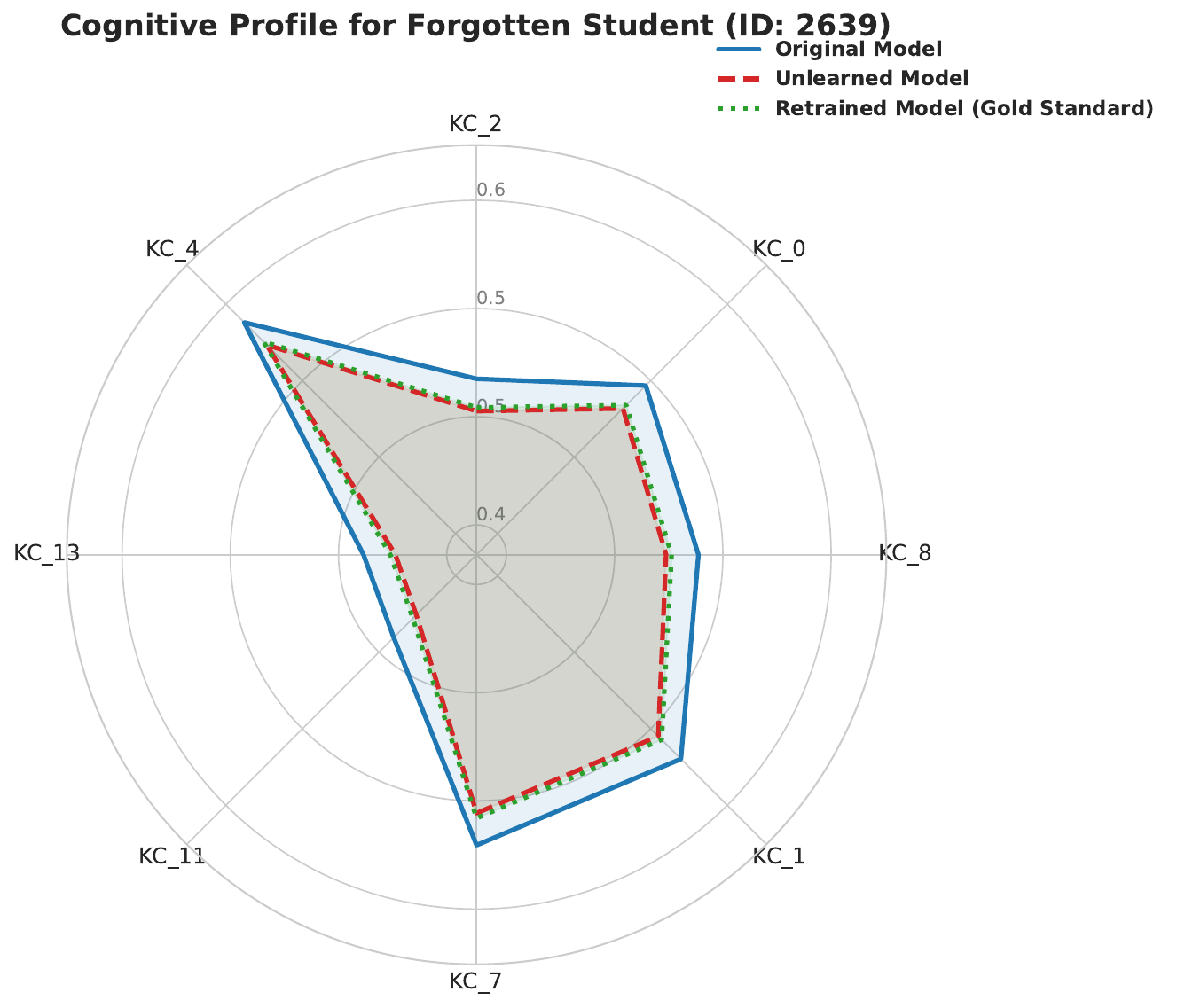}
			\subcaption{Student ID: 2639}
		\end{minipage}
		\begin{minipage}[t]{0.31\textwidth}
			\centering
			\includegraphics[width=5.5cm]{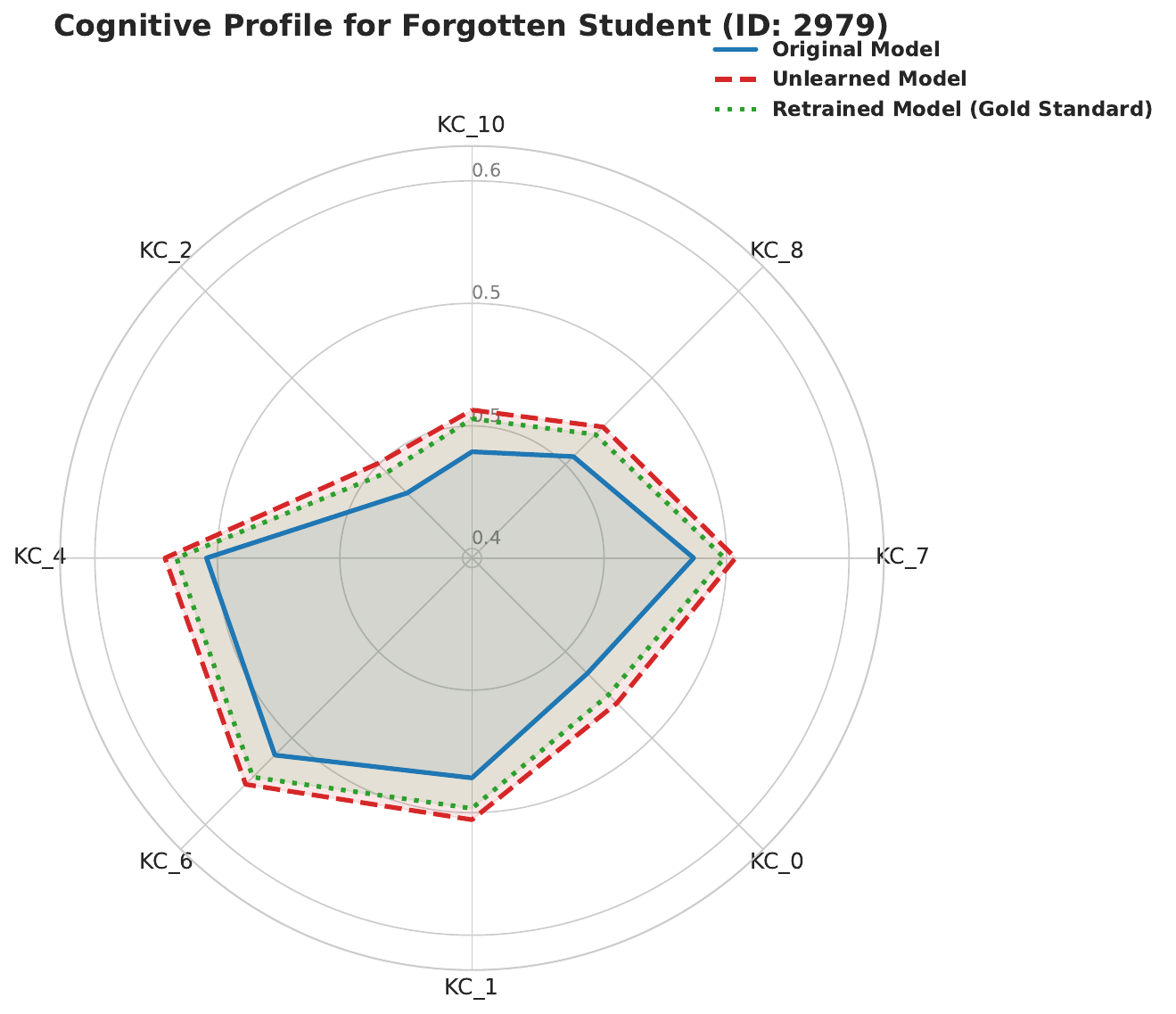}
			\subcaption{Student ID: 2979}
	   \end{minipage}
		\caption{Case study of cognitive profiles for three randomly selected forgotten students from the Math2 dataset (5\% unlearning ratio). The profiles are diagnosed by the original model (blue, solid), our unlearned model with HIF (red, dashed), and the retrained model (green, dotted), which serves as the gold standard.}
		\label{fig:qualitative analysis}
\end{figure*}

In summary, the experimental results provide compelling evidence that our proposed PrivacyCD framework achieves the best overall balance across the three critical dimensions of unlearning efficacy, model utility, and computational efficiency.
\subsection{Ablation Study and Parameter Analysis}
To gain a deeper understanding of our proposed PrivacyCD framework and its core HIF algorithm, this section presents two key analyses. First, we conduct an ablation study to verify the generalizability of the HIF algorithm and the superiority of our chosen architecture. Second, we perform a sensitivity analysis on HIF's three key hyperparameters.

\begin{table}[ht]
\centering
\renewcommand{\arraystretch}{1.2}
\resizebox{0.45\textwidth}{!}{%
\begin{tabular}{cccccc}
\hline
\textbf{Dataset}                     & \multicolumn{5}{c}{\textbf{Math2}}                                            \\ \hline
\multicolumn{1}{c|}{\multirow{2}{*}{\textbf{Model}}} &
  \multicolumn{2}{c}{\textbf{Utility}} &
  \multicolumn{2}{c}{\textbf{MIA}} &
  \multirow{2}{*}{\textbf{\begin{tabular}[c]{@{}c@{}}RTRR\end{tabular}}} \\ \cline{2-5}
\multicolumn{1}{c|}{}                & \textbf{AUC} & \textbf{ACC} & \textbf{MIA\_AUC} & \textbf{MIA\_ACC} &         \\ \hline
\multicolumn{1}{c|}{M\_orig}         & 0.7747       & 0.7066       & 0.7967            & 0.7474            & *       \\
\multicolumn{1}{c|}{M\_retrain}      & *            & *            & \textit{0.4979}   & \textit{0.5043}   & *       \\
\multicolumn{1}{c|}{NueralCD w/ HIF} & 0.7735       & 0.7022       & 0.5431            & 0.5736            & 88.03\% \\ \hline
\multicolumn{1}{c|}{M\_orig}         & 0.7997       & 0.7216       & 0.8258            & 0.7608            & *       \\
\multicolumn{1}{c|}{M\_retrain}      & *            & *            & \textit{0.4790}    & \textit{0.4925}   & *       \\
\multicolumn{1}{c|}{KSCD w/ HIF}     & 0.7997       & 0.7216       & 0.5665            & 0.5681            & 77.59\% \\ \hline
\end{tabular}%
}
\caption{Ablation study results of the HIF algorithm on two different CD architectures (NeuralCDM and KSCD), exemplified on the Math2 dataset with a 5\% unlearning ratio.}
\label{abres}
\end{table}
\subsubsection{Ablation Study: HIF's Performance on Different CD Models}
Our HIF algorithm features a core innovation: the hierarchical smoothing mechanism. Its effectiveness can be validated by directly comparing the results of PrivacyCD (the full HIF) with those of FIM in Table~\ref{overallres}, which serves as a key ablation analysis. To further investigate the generalizability of HIF as a universal unlearning strategy and to validate our hypothesis regarding the impact of model architecture on unlearning performance, we conducted an additional architecture-level ablation study. Specifically, we applied the HIF algorithm to two other prevailing neural CD models: NeuralCDM and KSCD. As a representative example, Table~\ref{abres} presents the performance of HIF on these two architectures on the Math2 dataset with a 5\% unlearning ratio. The complete results across all datasets and unlearning ratios are detailed in \textbf{Appendix C.1}. The representative results in Table~\ref{abres} clearly validate the generalizability and effectiveness of our HIF algorithm. On both the NeuralCDM and KSCD architectures, HIF successfully and significantly reduces the initially high MIA metrics to levels much closer to the ideal unlearning state, demonstrating that HIF is not limited to a single model structure but can be broadly applied to prevailing neural CD models. 

\subsubsection{Parameter Sensitivity Analysis}
The performance of the HIF algorithm is jointly controlled by three key hyperparameters: the selection threshold $\alpha$, the unlearning strength $\lambda$, and the hierarchical smoothing factor $\beta$. To investigate their individual effects, we conducted a series of sensitivity analyses on the Math2 dataset with a 5\% unlearning ratio. The complete results across all datasets and unlearning ratios are detailed in \textbf{Appendix C.2}. In each experiment, we fixed two of the parameters to their optimal values while varying the third to observe the resulting trends in model performance.

As depicted in Figure~\ref{fig:sensitivity_analysis}, our sensitivity analysis confirms the predictable behavior of the HIF algorithm. The results clearly show the expected trade-offs: increasing the selection threshold $\alpha$ or decreasing the unlearning strength $\lambda$ weakens the unlearning effect (i.e., increases MIA metrics), and vice versa. The most critical finding comes from the analysis of the smoothing factor $\beta$ (Figure~\ref{fig:sensitivity_analysis} (c)). As $\beta$ increases from 0 to a small positive value, the MIA metrics significantly drop to their optimal point while model utility remains almost unaffected. This provides strong empirical evidence for the effectiveness of our core hierarchical wisdom concept: a moderate degree of layer-wise smoothing leads to a more thorough unlearning process without compromising model performance.

\subsection{Qualitative Analysis}
To intuitively illustrate HIF's unlearning effectiveness, we conduct a case study by selecting a random forgotten student from the Math2 dataset (5\% unlearning ratio) and comparing their cognitive profiles generated by three key models: M\_orig, our unlearned model M\_unlearn, and the gold-standard M\_retrain. By benchmarking the profile from M\_unlearn against that of M\_orig (the memorized state) and M\_retrain (the ideal forgotten state), we can visually assess the completeness of the unlearning process. More comprehensive case studies are detailed in \textbf{Appendix C.3}.

Figure~\ref{fig:qualitative analysis} intuitively demonstrates the effectiveness of HIF through a three-student case study. In the figure, the original model (blue, solid line) exhibits a distinct, overfitted cognitive profile for each forgotten student. In contrast, the gold-standard retrained model (green, dotted line) produces a more contracted profile, representing an uninformative diagnosis of a stranger. Critically, the profiles generated by our unlearned model (red, dashed line) closely match those of the retrained model in all cases. This visual evidence strongly supports that HIF successfully restores the model to a state nearly identical to one that has never seen the student's data, achieving thorough and reliable unlearning.
\section{Conclusion}
In this paper, we presented the first systematic study on machine unlearning for CD models. We introduced PrivacyCD, a comprehensive framework featuring a novel, unlearning-friendly neural architecture and our core algorithm, HIF. HIF's theoretically-grounded hierarchical smoothing mechanism leverages the layer-wise structure of CD models to achieve more robust parameter importance estimation. Extensive experiments on three real-world datasets demonstrate that HIF significantly outperforms existing baselines, achieving a state-of-the-art balance between unlearning efficacy, model utility, and efficiency. Our work provides a practical and effective solution for deploying privacy-preserving AI in education and offers key insights into the co-design of unlearnable models and algorithms.
\bibliography{aaai2026}
\appendix
\section{Proof of Corollary 1}
\label{appendix:proof_detailed}

Here we provide a detailed proof for the superiority of HIF's importance estimation mechanism as stated in Corollary 1, including the full derivation of the mean squared error (MSE).

Let $\mu_i$ be the ``true'', unobserved importance of a parameter $\theta_i$ with respect to the forget set $\mathcal{D}_f$. The empirically computed importance using the FIM, $I_{\text{param}}(\theta_i, \mathcal{D}_f)$, is modeled as a noisy observation of $\mu_i$:
\begin{equation}
    I_{\text{param}}(\theta_i, \mathcal{D}_f) = \mu_i + \epsilon_i
\end{equation}
where $\epsilon_i$ is a random noise term with $\mathbb{E}[\epsilon_i] = 0$. For a given functional layer $L_j$ with $|L_j| = p$ parameters, we make the following assumptions:
\begin{enumerate}
    \item The noise terms $\epsilon_1, \dots, \epsilon_p$ are independent and identically distributed (i.i.d.) with variance $\sigma^2$.
    \item The true importance values $\mu_1, \dots, \mu_p$ can be viewed as samples drawn from a common prior distribution with mean $\bar{\mu} = \frac{1}{p}\sum_{k=1}^{p} \mu_k$.
\end{enumerate}

The naive estimator uses $I_{\text{param}}(\theta_i)$ directly. The MSE for a single parameter is:
\begin{equation}
\begin{aligned}
    \text{MSE}(I_{\text{param}}(\theta_i)) &= \mathbb{E}[(I_{\text{param}}(\theta_i) - \mu_i)^2] \\
    &= \mathbb{E}[(\mu_i + \epsilon_i - \mu_i)^2]\\
    &= \mathbb{E}[\epsilon_i^2] \\
    &= \sigma^2
\end{aligned}
\end{equation}    
The total MSE for all $p$ parameters in layer $L_j$ is:
\begin{equation}
    \sum_{i=1}^{p} \text{MSE}(I_{\text{param}}(\theta_i)) = p \sigma^2
\end{equation}

The HIF estimator uses the adjusted importance $I_{\text{adj}}(\theta_i) = (1-\beta)I_{\text{param}}(\theta_i) + \beta I_{\text{layer}}$, where $I_{\text{layer}} = \frac{1}{p}\sum_{k=1}^{p} I_{\text{param}}(\theta_k)$. Let's derive its MSE.
Let $\bar{\epsilon} = \frac{1}{p}\sum_{k=1}^{p} \epsilon_k$. Then $I_{\text{layer}} = \bar{\mu} + \bar{\epsilon}$. The MSE for a single parameter $\theta_i$ is:
\begin{align}
    \text{MSE}(I_{\text{adj}}(\theta_i)) &= \mathbb{E}\left[ \left( (1-\beta)I_{\text{param}}(\theta_i) + \beta I_{\text{layer}} - \mu_i \right)^2 \right] \\
    &= \mathbb{E}\left[ \left( (1-\beta)(\mu_i + \epsilon_i) + \beta (\bar{\mu} + \bar{\epsilon}) - \mu_i \right)^2 \right] \label{eq:substitute} \\
    &= \mathbb{E}\left[ \left( (\mu_i - \beta\mu_i + \epsilon_i - \beta\epsilon_i) + (\beta\bar{\mu} + \beta\bar{\epsilon}) - \mu_i \right)^2 \right] \\
    &= \mathbb{E}\left[ \left( \beta(\bar{\mu} - \mu_i) + (1-\beta)\epsilon_i + \beta\bar{\epsilon} \right)^2 \right] \label{eq:rearrange}
\end{align}
Expanding the square, we get three squared terms and three cross-product terms. The expectation of the cross-product terms is zero due to the independence of the noise terms and the fact that $\mathbb{E}[\epsilon_i]=0$. For example, $\mathbb{E}[(\bar{\mu} - \mu_i)\epsilon_i] = (\bar{\mu} - \mu_i)\mathbb{E}[\epsilon_i] = 0$. Similarly, $\mathbb{E}[\epsilon_i \bar{\epsilon}] = \mathbb{E}[\epsilon_i \frac{1}{p}\sum_k \epsilon_k] = \frac{1}{p}\mathbb{E}[\epsilon_i^2] = \frac{\sigma^2}{p}$ for $i=k$, and 0 otherwise. A more careful expansion shows that the cross terms' expectations are zero. We are left with the sum of the expectations of the squared terms:
\begin{equation}
    \text{MSE}(I_{\text{adj}}(\theta_i)) = \mathbb{E}[ \beta^2(\bar{\mu} - \mu_i)^2 ] + \mathbb{E}[ (1-\beta)^2\epsilon_i^2 ] + \mathbb{E}[ \beta^2\bar{\epsilon}^2 ]
\end{equation}
Since the true values $\mu_i$ are fixed, and given that $\mathbb{E}[\epsilon_i^2]=\sigma^2$ and $\mathbb{E}[\bar{\epsilon}^2] = \text{Var}(\bar{\epsilon}) = \frac{\sigma^2}{p}$ (as $\epsilon_i$ are i.i.d.), the expression becomes:
\begin{equation}
    \text{MSE}(I_{\text{adj}}(\theta_i)) = \beta^2(\bar{\mu} - \mu_i)^2 + (1-\beta)^2\sigma^2 + \beta^2\frac{\sigma^2}{p}
\end{equation}
Now, we sum the MSEs over all $p$ parameters in the layer:
\begin{align}
    \sum_{i=1}^{p} \text{MSE}(I_{\text{adj}}(\theta_i)) &= \sum_{i=1}^{p} \left( \beta^2(\bar{\mu} - \mu_i)^2 + (1-\beta)^2\sigma^2 + \beta^2\frac{\sigma^2}{p} \right) \\
    &= \beta^2 \sum_{i=1}^{p} (\bar{\mu} - \mu_i)^2 + p(1-\beta)^2\sigma^2 + p\left(\beta^2\frac{\sigma^2}{p}\right) \\
    &= \beta^2 \sum_{i=1}^{p} (\bar{\mu} - \mu_i)^2 + p(1-2\beta+\beta^2)\sigma^2 + \beta^2\sigma^2 \\
    &= \beta^2 \left( \sum_{i=1}^{p} (\bar{\mu} - \mu_i)^2 + p\sigma^2 + \sigma^2 \right) - 2p\beta\sigma^2 + p\sigma^2 \label{eq:total_mse}
\end{align}
This total MSE is a quadratic function of $\beta$. To find the $\beta$ that minimizes this function, we can take the derivative with respect to $\beta$ and set it to zero. This leads to the optimal shrinkage factor, $\beta^*$, known as the James-Stein estimator:
\begin{equation}
\begin{aligned}
    \beta^* &= \frac{p\sigma^2}{\sum_{i=1}^{p} (\bar{\mu} - \mu_i)^2 + p\sigma^2 + \sigma^2} \\
    & \approx \frac{(p-2)\sigma^2}{\sum_{k=1}^{p}(I_{\text{param}}(\theta_k) - I_{\text{layer}})^2}
\end{aligned}
\end{equation}
The approximation on the right uses the sample variances to estimate the true variances. The James-Stein theorem proves that for $p \ge 3$, if we choose a $\beta$ in the range $(0, 2\beta^*)$, the total MSE of the adjusted estimator will be strictly lower than the total MSE of the naive estimator ($p\sigma^2$). Therefore, by selecting a suitable $\beta \in (0, 1)$, it is possible to reduce the total estimation error:
\begin{equation}
    \sum_{i=1}^{p} \mathbb{E}[(I_{\text{adj}}(\theta_i) - \mu_i)^2] < \sum_{i=1}^{p} \mathbb{E}[(I_{\text{param}}(\theta_i) - \mu_i)^2]
\end{equation}
This holds because the significant reduction in the variance term of the total MSE outweighs the small bias introduced by shrinking the estimates toward the mean.

The derivation shows that the total MSE of HIF's importance estimates is provably lower than that of the naive FIM approach for any suitable choice of $\beta \in (0,1)$ when $p \ge 3$. This provides a more accurate and reliable basis for unlearning decisions, reducing both false positives and false negatives, thus leading to a better trade-off between unlearning effectiveness and model utility preservation. This completes the proof of Corollary 1.

\section{Supplementary Information on Experimental Settings}
\subsection{Datasets}
A detailed description of the datasets used in this paper is as follows:

\begin{itemize}
    \item \textbf{Math1 \& Math2:} These two datasets are derived from two large-scale high school final examinations, representing more complex and realistic diagnostic scenarios.
    \begin{itemize}
        \item The \textbf{Math1} dataset contains 84,180 response records from 4,209 students on 20 questions, which are associated with 11 KCs in the domain of algebra.
        \item The \textbf{Math2} dataset contains 78,220 response records from 3,911 students on 20 questions, covering 16 KCs spanning both algebra and geometry.
    \end{itemize}
    A key characteristic of these datasets is that the exercises often require the simultaneous mastery of multiple KCs, making them ideal for evaluating model performance in complex educational assessment settings.

    \item \textbf{FrcSub:} This is a classic benchmark dataset in the CD field, originally introduced by Tatsuoka. It contains the response records of 536 middle school students on 20 fraction subtraction problems. The value of this dataset lies in its fine-grained Q-matrix, which defines the 8 underlying cognitive skills (e.g., simplifying fractions, converting whole numbers to fractions) required to perform fraction subtraction. Due to its focused domain and well-defined skill set, FrcSub is widely used to evaluate the fine-grained diagnostic capabilities of CD models.
\end{itemize}
\begin{figure*}[ht]
\centering
\includegraphics[width=0.85\textwidth]{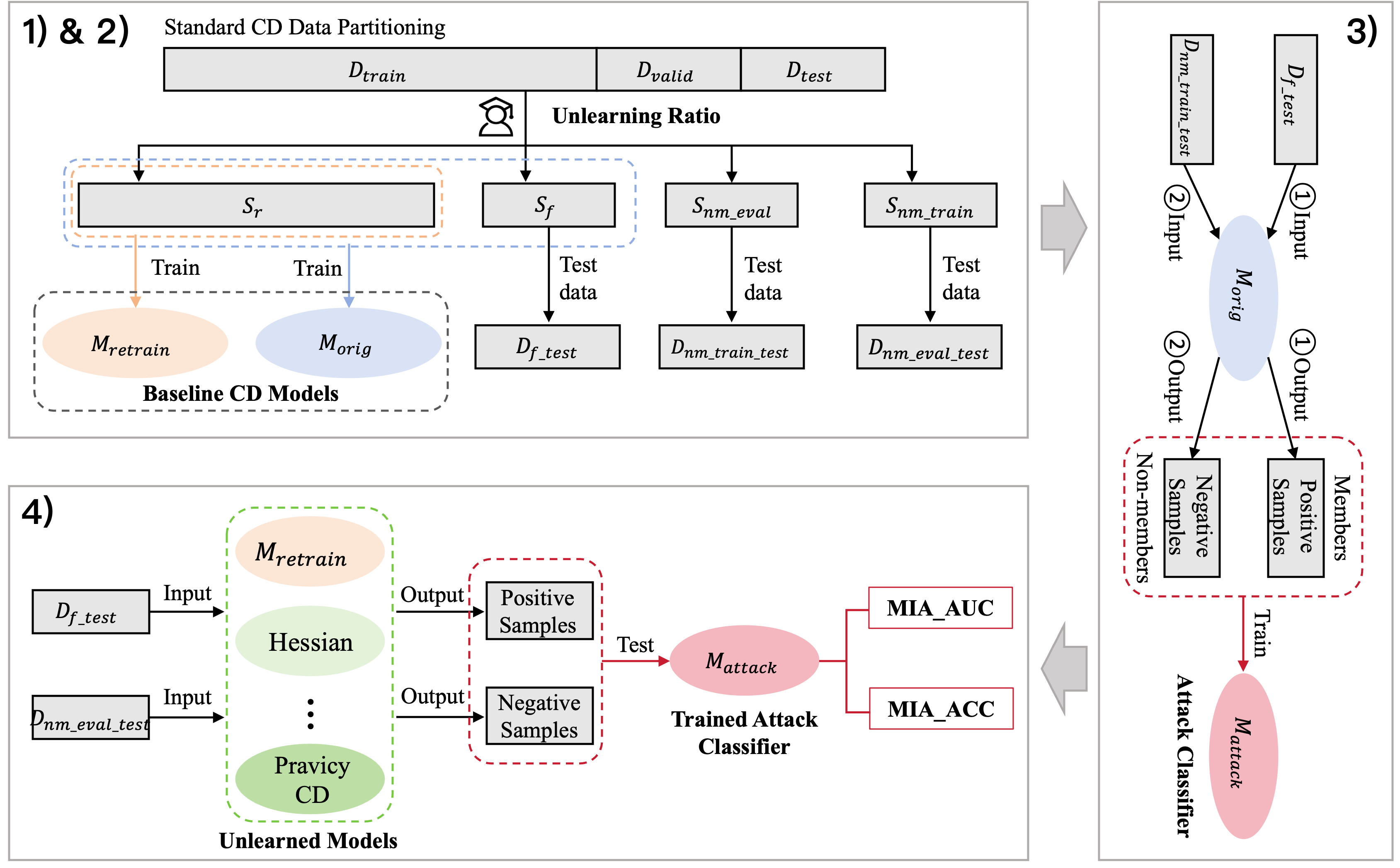} 
\caption{The overall pipeline of MIA Protocol.}
\label{pipe_mia}
\end{figure*}
\subsection{Evaluation Metrics}

This section provides a detailed explanation of the evaluation metrics used in our experiments, including the specific definitions and calculation procedures for model utility and unlearning efficacy metrics.

\subsubsection{B.2.1 Model Utility Metrics}
To measure the cognitive diagnostic capability retained by the model after an unlearning operation, we employ ACC and AUC.

\begin{itemize}
    \item \textbf{ACC:} In the context of cognitive diagnosis, ACC measures the correctness of the model's predictions on student responses. We binarize the model's continuous probability outputs using a threshold of 0.5 (a prediction of $\ge 0.5$ is classified as a correct response, i.e., 1, while $< 0.5$ is classified as incorrect, i.e., 0). We then calculate the proportion of predictions that match the true response labels. ACC provides a direct and intuitive measure of the model's fundamental performance in correctly identifying right and wrong answers.

    \item \textbf{AUC:} Unlike ACC, AUC is a threshold-independent evaluation metric. It measures the probability that the model will rank a randomly chosen positive instance (a correct student response) higher than a randomly chosen negative instance (an incorrect student response). In CD, a high AUC value signifies that the model's output probabilities have strong ranking power, meaning the model is more confident (i.e., assigns a higher probability) to the responses that the student actually answered correctly. Compared to ACC, AUC provides a better assessment of the quality of the model's predictive confidence and is more robust in scenarios with imbalanced class distributions.
\end{itemize}

\subsubsection{B.2.2 MIA Protocol for Unlearning Efficacy}
 We have designed a rigorous MIA protocol to quantify the completeness of the unlearning process. This protocol is built upon a standard data split for the CD task, which is then re-partitioned to create the necessary datasets for the attack, as shown in Figure \ref{pipe_mia}.

\textbf{Prerequisite: Standard CD Data Partitioning.} For each dataset, we first follow the standard procedure for CD tasks. All student response records are randomly shuffled and partitioned, for example, in a 6:2:2 ratio, into a training set ($\mathcal{D}_{train}$), a validation set ($\mathcal{D}_{valid}$), and a test set ($\mathcal{D}_{test}$). All subsequent MIA-related operations are based on these pre-established partitions.

\textbf{The MIA Protocol in Detail.} Our MIA protocol consists of four core steps:

\begin{enumerate}[label=\arabic*)]
    \item \textbf{Constructing MIA-Specific Datasets:} To build the datasets required for training and evaluating the attack model, we re-partition the standard splits \textit{at the student level}. Specifically, we define a hyperparameter, the unlearning ratio (e.g., 10\%, 5\%, or 1\%), which determines the proportion of students in the forget and non-member sets. We partition the entire population of students into three disjoint groups:
    \begin{itemize}
        \item \textbf{Forget Students ($\mathcal{S}_f$):} A fraction of the total students corresponding to the unlearning ratio.
        \item \textbf{Non-member Train Students ($\mathcal{S}_{nm\_train}$):} Another fraction of students, also determined by the unlearning ratio.
        \item \textbf{Non-member Evaluation Students ($\mathcal{S}_{nm\_eval}$):} A third fraction of students, also sized according to the unlearning ratio.
        \item \textbf{Retain Students ($\mathcal{S}_r$):} The remaining students.
    \end{itemize}
    Based on these student groups, we derive the MIA data subsets from the original $\mathcal{D}_{train}$, $\mathcal{D}_{valid}$, and $\mathcal{D}_{test}$. For example, the training part of the forget set, $\mathcal{D}_{f\_train}$, consists of all records from students in $\mathcal{S}_f$ that are present in the original $\mathcal{D}_{train}$.

    \item \textbf{Training Baseline CD Models:}
    \begin{itemize}
        \item \textbf{Original Model ($M_{orig}$):} This model simulates a deployed system containing the data to be unlearned. It is trained on the union of the training and validation sets of both the retain students ($\mathcal{S}_r$) and the forget students ($\mathcal{S}_f$).
        \item \textbf{Retrain Model ($M_{retrain}$):} This model serves as the gold standard for unlearning. It is trained from scratch only on the training and validation sets of the retain students ($\mathcal{S}_r$).
    \end{itemize}

    \item \textbf{Training the Attack Classifier:} The attack classifier learns to distinguish between data that a model has been trained on (members) and data it has not seen (non-members). We use the original model, $M_{orig}$, to generate the training data for this classifier.
    \begin{itemize}
        \item \textbf{Positive Samples (Members):} We feed the test data of the forget students ($\mathcal{D}_{f\_test}$) into $M_{orig}$ to obtain its predictive outputs (e.g., probability vectors). These samples are labeled as ``members'' (label 1).
        \item \textbf{Negative Samples (Non-members):} We feed the test data of the non-member train students ($\mathcal{D}_{nm\_train\_test}$) into $M_{orig}$ to obtain its predictions. These samples are labeled as ``non-members'' (label 0).
        \item \textbf{Training:} Using these (prediction, label) pairs, we train an attack classifier (e.g., XGBoost classifier).
    \end{itemize}

    \item \textbf{Evaluating Unlearning Efficacy:} We use the non-member evaluation students ($\mathcal{S}_{nm\_eval}$) as a clean control group of non-members to evaluate the effectiveness of different unlearning algorithms.
    \begin{itemize}
        \item For each unlearned model $M_{unlearn}$ to be evaluated (including $M_{retrain}$, HIF, etc.), we generate test data for the attack:
            \begin{itemize}
                \item \textit{Test Positive Samples:} Obtain predictions from $M_{unlearn}$ on the test data of the forget students ($\mathcal{D}_{f\_test}$).
                \item \textit{Test Negative Samples:} Obtain predictions from $M_{unlearn}$ on the test data of the non-member evaluation students ($\mathcal{D}_{nm\_eval\_test}$).
            \end{itemize}
        \item \textbf{Evaluation:} We feed this test data into the trained attack classifier and compute its \textbf{MIA\_AUC} and \textbf{MIA\_ACC}.
        \item \textbf{Interpretation:} For the original model $M_{orig}$, we expect MIA metrics to be significantly higher than 0.5, indicating a privacy leak. For an ideal unlearned model like $M_{retrain}$, the metrics should be close to 0.5, indicating that the attack classifier cannot distinguish members from non-members. An effective unlearning algorithm should yield MIA metrics that are substantially lower than those for $M_{orig}$ and as close as possible to those for $M_{retrain}$.
    \end{itemize}
\end{enumerate}
\subsubsection{B.2.3 Rationale and Rigor of the MIA Protocol}
We designed the specific MIA protocol described above with careful consideration for the unique characteristics of data in the CD domain. Unlike tasks such as computer vision, which often deal with independent and identically distributed (i.i.d.) samples (e.g., individual images), the core data in CD consists of response record sequences that are \textbf{bound to specific student identities}. The goal of unlearning is to erase all traces of a particular student. Consequently, our data partitioning must be performed at the student level to ensure the validity of the evaluation.

Furthermore, our protocol design strictly adheres to three fundamental principles of rigorous MIA evaluation to ensure the fairness of our experiments and the reliability of our conclusions:
\begin{enumerate}[label=\arabic*)]
    \item \textbf{Principle of Single Exposure for Member Data:} The original model, $M_{orig}$, is exposed to the forget set, $\mathcal{D}_f$, only once during its initial training phase. This ensures that our attack simulates a realistic, one-shot inference attempt on a deployed model.

    \item \textbf{Principle of Purity for Non-Member Data:} Neither the original model $M_{orig}$ nor the retrain model $M_{retrain}$ ever encounters any data from the non-member sets ($\mathcal{D}_{nm\_train}$ and $\mathcal{D}_{nm\_eval}$) during their training. This guarantees that the ``non-member'' status is pure and uncontaminated.

    \item \textbf{Principle of Separation for Training and Evaluation:} This is a critical aspect of our protocol's design. We partition the non-member set into two distinct subsets: a \textbf{non-member train set} ($\mathcal{S}_{nm\_train}$) and a \textbf{non-member evaluation set} ($\mathcal{S}_{nm\_eval}$). The former is used exclusively for \textit{training} the attack classifier, while the latter is used exclusively for evaluating its final performance. Failing to make this separation and using the same non-member data for both training and testing the attacker would lead to overly optimistic results, as it would not reflect the unlearning algorithm's true privacy-preserving capabilities against genuinely unseen data. This strict separation ensures that our evaluation of unlearning efficacy is unbiased and rigorous.
\end{enumerate}
\subsection{Implementation Details}
This section provides detailed implementation settings for our experiments to ensure full reproducibility.
\subsubsection{B.3.1 CD Model Architecture and Hyperparameter Tuning}
All our experiments are based on the neural CD architecture described in Section 3.2. To ensure that our models achieve their optimal performance, we employed \textbf{Weights \& Biases (WanDB)}\footnote{https://wandb.ai/site/}, a leading platform for experiment tracking and hyperparameter optimization, to conduct a systematic hyperparameter search. The tuning process focused on key hyperparameters, including the embedding dimension, the hidden layer architecture of the MLP in the interaction module, learning rate, batch size, and dropout rates. By leveraging WandB's Bayesian search sweeps, we automatically explored the vast parameter space and selected an optimal configuration for each dataset based on the best performance on the validation set. All models, including the original ($M_{orig}$) and retrain ($M_{retrain}$) models, were subsequently trained using their respective optimal configurations. The complete scripts for our WanDB hyperparameter sweeps and the detailed experimental logs are available in our public code repository to ensure full reproducibility.
\begin{table*}[ht]
\centering
\renewcommand{\arraystretch}{1.5}
\resizebox{\textwidth}{!}{%
\begin{tabular}{cc|ccccc|ccccc|ccccc}
\hline
\multicolumn{2}{c|}{\textbf{Dataset}} &
  \multicolumn{5}{c|}{\textbf{Math1}} &
  \multicolumn{5}{c|}{\textbf{Math2}} &
  \multicolumn{5}{c}{\textbf{Frcsub}} \\ \hline
\multicolumn{1}{c|}{\multirow{2}{*}{\textbf{Unlearning Ratio}}} &
  \multirow{2}{*}{\textbf{Model}} &
  \multicolumn{2}{c}{\textbf{Utility}} &
  \multicolumn{2}{c}{\textbf{MIA}} &
  \multirow{2}{*}{\textbf{\begin{tabular}[c]{@{}c@{}}Relative Time \\ Reduction Rate\end{tabular}}} &
  \multicolumn{2}{c}{\textbf{Utility}} &
  \multicolumn{2}{c}{\textbf{MIA}} &
  \multirow{2}{*}{\textbf{\begin{tabular}[c]{@{}c@{}}Relative Time\\ Reducation Rate\end{tabular}}} &
  \multicolumn{2}{c}{\textbf{Time}} &
  \multicolumn{2}{c}{\textbf{MIA}} &
  \multirow{2}{*}{\textbf{\begin{tabular}[c]{@{}c@{}}Relative Time\\ Reduction Rate\end{tabular}}} \\ \cline{3-6} \cline{8-11} \cline{13-16}
\multicolumn{1}{c|}{} &
   &
  \textbf{AUC} &
  \textbf{ACC} &
  \textbf{MIA\_AUC} &
  \textbf{MIA\_ACC} &
   &
  \textbf{AUC} &
  \textbf{ACC} &
  \textbf{MIA\_AUC} &
  \textbf{MIA\_ACC} &
   &
  \textbf{Rate} &
  \textbf{ACC} &
  \textbf{MIA\_AUC} &
  \textbf{MIA\_ACC} &
   \\ \hline
\multicolumn{1}{c|}{\multirow{3}{*}{\textbf{10\%}}} &
  M\_orig &
  0.7726 &
  0.6944 &
  0.869 &
  0.8125 &
  * &
  0.7821 &
  0.7124 &
  0.8471 &
  0.7886 &
  * &
  0.8625 &
  0.7954 &
  0.8136 &
  0.7407 &
  * \\
\multicolumn{1}{c|}{} &
  M\_retrain &
  * &
  * &
  \textit{0.4768} &
  \textit{0.4823} &
  * &
  * &
  * &
  \textit{0.5049} &
  \textit{0.5086} &
  * &
  * &
  * &
  \textit{0.4868} &
  \textit{0.4879} &
  * \\
\multicolumn{1}{c|}{} &
  NeuralCD w/ HIF &
  0.7726 &
  0.6944 &
  0.6394 &
  0.6417 &
  93.76\% &
  0.7814 &
  0.7114 &
  0.5833 &
  0.6141 &
  84.27\% &
  0.8209 &
  0.7639 &
  0.6621 &
  0.5868 &
  85.25\% \\ \hline
\multicolumn{1}{c|}{\multirow{3}{*}{\textbf{5\%}}} &
  M\_orig &
  0.7791 &
  0.7014 &
  0.8194 &
  0.7605 &
  * &
  0.7747 &
  0.7066 &
  0.7967 &
  0.7474 &
  * &
  0.8623 &
  0.8019 &
  0.7891 &
  0.7085 &
  * \\
\multicolumn{1}{c|}{} &
  M\_retrain &
  * &
  * &
  \textit{0.4734} &
  \textit{0.4891} &
  * &
  * &
  * &
  \textit{0.4979} &
  \textit{0.5043} &
  * &
  * &
  * &
  \textit{0.4644} &
  \textit{0.4484} &
  * \\
\multicolumn{1}{c|}{} &
  NeuralCD w/ HIF &
  0.7791 &
  0.7014 &
  0.6162 &
  0.6275 &
  90.29\% &
  0.7735 &
  0.7022 &
  0.5431 &
  0.5736 &
  88.03\% &
  0.8054 &
  0.7291 &
  0.4563 &
  0.4753 &
  86.80\% \\ \hline
\multicolumn{1}{c|}{\multirow{3}{*}{\textbf{1\%}}} &
  M\_orig &
  0.7791 &
  0.7029 &
  0.7302 &
  0.6811 &
  * &
  0.7767 &
  0.7061 &
  0.7255 &
  0.6667 &
  * &
  0.8580 &
  0.793 &
  0.7943 &
  0.7143 &
  * \\
\multicolumn{1}{c|}{} &
  M\_retrain &
  * &
  * &
  \textit{0.4999} &
  \textit{0.4961} &
  * &
  * &
  * &
  \textit{0.5144} &
  \textit{0.5341} &
  * &
  * &
  * &
  \textit{0.5568} &
  \textit{0.5238} &
  * \\
\multicolumn{1}{c|}{} &
  NeuralCD w/ HIF &
  0.7791 &
  0.7029 &
  0.5175 &
  0.5315 &
  89.76\% &
  0.7631 &
  0.6864 &
  0.557 &
  0.5703 &
  85.29\% &
  0.7589 &
  0.671 &
  0.6159 &
  0.5476 &
  83.80\% \\ \hline
\end{tabular}%
}
\caption{Ablation study results for our HIF algorithm when applied to the NeuralCD architecture. The table shows performance across three datasets and three unlearning ratios. }
\label{ab_neuralcd}
\end{table*}

\begin{table*}[ht]
\centering
\renewcommand{\arraystretch}{1.5}
\resizebox{\textwidth}{!}{%
\begin{tabular}{cc|ccccc|ccccc|ccccc}
\hline
\multicolumn{2}{c|}{\textbf{Dataset}} &
  \multicolumn{5}{c|}{\textbf{Math1}} &
  \multicolumn{5}{c|}{\textbf{Math2}} &
  \multicolumn{5}{c}{\textbf{Frcsub}} \\ \hline
\multicolumn{1}{c|}{\multirow{2}{*}{\textbf{Unlearning Ratio}}} &
  \multirow{2}{*}{\textbf{Model}} &
  \multicolumn{2}{c}{\textbf{Utility}} &
  \multicolumn{2}{c}{\textbf{MIA}} &
  \multirow{2}{*}{\textbf{\begin{tabular}[c]{@{}c@{}}Relative Time \\ Reduction Rate\end{tabular}}} &
  \multicolumn{2}{c}{\textbf{Utility}} &
  \multicolumn{2}{c}{\textbf{MIA}} &
  \multirow{2}{*}{\textbf{\begin{tabular}[c]{@{}c@{}}Relative Time\\ Reducation Rate\end{tabular}}} &
  \multicolumn{2}{c}{\textbf{Time}} &
  \multicolumn{2}{c}{\textbf{MIA}} &
  \multirow{2}{*}{\textbf{\begin{tabular}[c]{@{}c@{}}Relative Time\\ Reduction Rate\end{tabular}}} \\ \cline{3-6} \cline{8-11} \cline{13-16}
\multicolumn{1}{c|}{} &
   &
  \textbf{AUC} &
  \textbf{ACC} &
  \textbf{MIA\_AUC} &
  \textbf{MIA\_ACC} &
   &
  \textbf{AUC} &
  \textbf{ACC} &
  \textbf{MIA\_AUC} &
  \textbf{MIA\_ACC} &
   &
  \textbf{Rate} &
  \textbf{ACC} &
  \textbf{MIA\_AUC} &
  \textbf{MIA\_ACC} &
   \\ \hline
\multicolumn{1}{c|}{\multirow{3}{*}{\textbf{10\%}}} &
  M\_orig &
  0.7884 &
  0.7151 &
  0.9522 &
  0.9041 &
  * &
  0.8013 &
  0.7233 &
  0.8537 &
  0.7996 &
  * &
  0.8623 &
  0.7960 &
  0.7860 &
  0.6989 &
  * \\
\multicolumn{1}{c|}{} &
  M\_retrain &
  * &
  * &
  \textit{0.4827} &
  \textit{0.4885} &
  * &
  * &
  * &
  \textit{0.5304} &
  \textit{0.5180} &
  * &
  * &
  * &
  \textit{0.5060} &
  \textit{0.5297} &
  * \\
\multicolumn{1}{c|}{} &
  KSCD w/ HIF &
  0.7884 &
  0.7151 &
  0.6813 &
  0.6885 &
  80.80\% &
  0.8013 &
  0.7227 &
  0.5821 &
  0.6067 &
  84.27\% &
  0.8607 &
  0.7942 &
  0.5075 &
  0.5319 &
  92.27\% \\ \hline
\multicolumn{1}{c|}{\multirow{3}{*}{\textbf{5\%}}} &
  M\_orig &
  0.7869 &
  0.7131 &
  0.8924 &
  0.8538 &
  * &
  0.7997 &
  0.7216 &
  0.8258 &
  0.7608 &
  * &
  0.8617 &
  0.7918 &
  0.7260 &
  0.6502 &
  * \\
\multicolumn{1}{c|}{} &
  M\_retrain &
  * &
  * &
  \textit{0.4917} &
  \textit{0.4969} &
  * &
  * &
  * &
  \textit{0.4790} &
  \textit{0.4925} &
  * &
  * &
  * &
  \textit{0.4607} &
  \textit{0.4574} &
  * \\
\multicolumn{1}{c|}{} &
  KSCD w/ HIF &
  0.7869 &
  0.7131 &
  0.756 &
  0.7317 &
  90.29\% &
  0.7997 &
  0.7216 &
  0.5665 &
  0.5681 &
  77.59\% &
  0.8610 &
  0.7928 &
  0.5112 &
  0.5022 &
  93.18\% \\ \hline
\multicolumn{1}{c|}{\multirow{3}{*}{\textbf{1\%}}} &
  M\_orig &
  0.7723 &
  0.7009 &
  0.818 &
  0.7756 &
  * &
  0.8075 &
  0.728 &
  0.8162 &
  0.747 &
  * &
  0.8582 &
  0.7899 &
  0.8045 &
  0.7143 &
  * \\
\multicolumn{1}{c|}{} &
  M\_retrain &
  * &
  * &
  \textit{0.4738} &
  \textit{0.5000} &
  * &
  * &
  * &
  \textit{0.4538} &
  \textit{0.4699} &
  * &
  * &
  * &
  \textit{0.5216} &
  \textit{0.5000} &
  * \\
\multicolumn{1}{c|}{} &
  KSCD w/ HIF &
  0.7723 &
  0.7009 &
  0.607 &
  0.6063 &
  84.94\% &
  0.7994 &
  0.7220 &
  0.4261 &
  0.4458 &
  81.10\% &
  0.8514 &
  0.7859 &
  0.5034 &
  0.5000 &
  93.81\% \\ \hline
\end{tabular}%
}
\caption{Ablation study results for our HIF algorithm when applied to the KSCD architecture. The table shows performance across three datasets and three unlearning ratios.}
\label{ab_kscd}
\end{table*}
\subsubsection{B.3.2 Unlearning Hyperparameters Tuning}
The hyperparameter tuning for all unlearning algorithms was conducted on top of the optimal CD model configurations determined in the previous step. Unlike standard model training, the objective of unlearning is multi-faceted: it requires simultaneously maintaining high model utility on the retain set (high AUC/ACC) and ensuring unlearning completeness (MIA metrics close to the gold standard). The multi-objective nature of this task makes it less suitable for automated tuning platforms like WandB. Therefore, we developed custom tuning scripts to perform a grid search to identify the best-performing hyperparameter combination for each method on the validation set.

The specific search spaces for each method are detailed below:
\begin{itemize}
    \item \textbf{Gradient Ascent (GradAsc):} The unlearning learning rate was chosen from $\{10^{-5}, 5 \times 10^{-5}, 10^{-4}\}$, and the number of unlearning steps was chosen from $\{1, 3, 5\}$.
    
    \item \textbf{Hessian-based:} We tuned its two key parameters: the number of samples for Hessian estimation was searched in $\{10, 20, 40\}$, and the number of batches for Hessian-Vector Product (HVP) computation was selected from $\{1, 2\}$.
    
    \item \textbf{FIM Method:} This is a prevailing approximate unlearning technique. Its core idea is to quantify the importance of each parameter with respect to the forgotten data at an individual level using the FIM. We performed a grid search for its two key hyperparameters: the selection threshold $\alpha$ over $\{1.3, 2.0, 2.5, 5.0\}$ and the unlearning strength $\lambda$ over $\{0.1, 0.3, 0.5, 0.8\}$.
    
    \item \textbf{HIF (Ours):} Our proposed HIF algorithm pioneers a different approach that originates from the hierarchical structure of the model, leveraging a collective wisdom to guide the unlearning process. Its performance is jointly governed by three distinct hyperparameters: a selection threshold $\alpha$, an unlearning strength $\lambda$, and the core smoothing factor $\beta$, which controls the intensity of the collective information. We performed a joint grid search for these parameters over the ranges $\alpha \in \{1.3, 2.0, 2.5, 5.0\}$, $\lambda \in \{0.1, 0.3, 0.5, 0.8\}$, and $\beta \in \{0.02, 0.05, 0.1, 0.3, 0.5\}$.
\end{itemize}
For each method, we iterated through all possible hyperparameter combinations and selected the configuration that achieved the best trade-off between model utility (AUC/ACC as close as possible to the original model) and unlearning efficacy (MIA metrics as close as possible to the retrain model).
\begin{figure*}[ht]
		\centering
		\begin{minipage}[t]{0.31\textwidth}
			\centering
			\includegraphics[width=5.5cm]{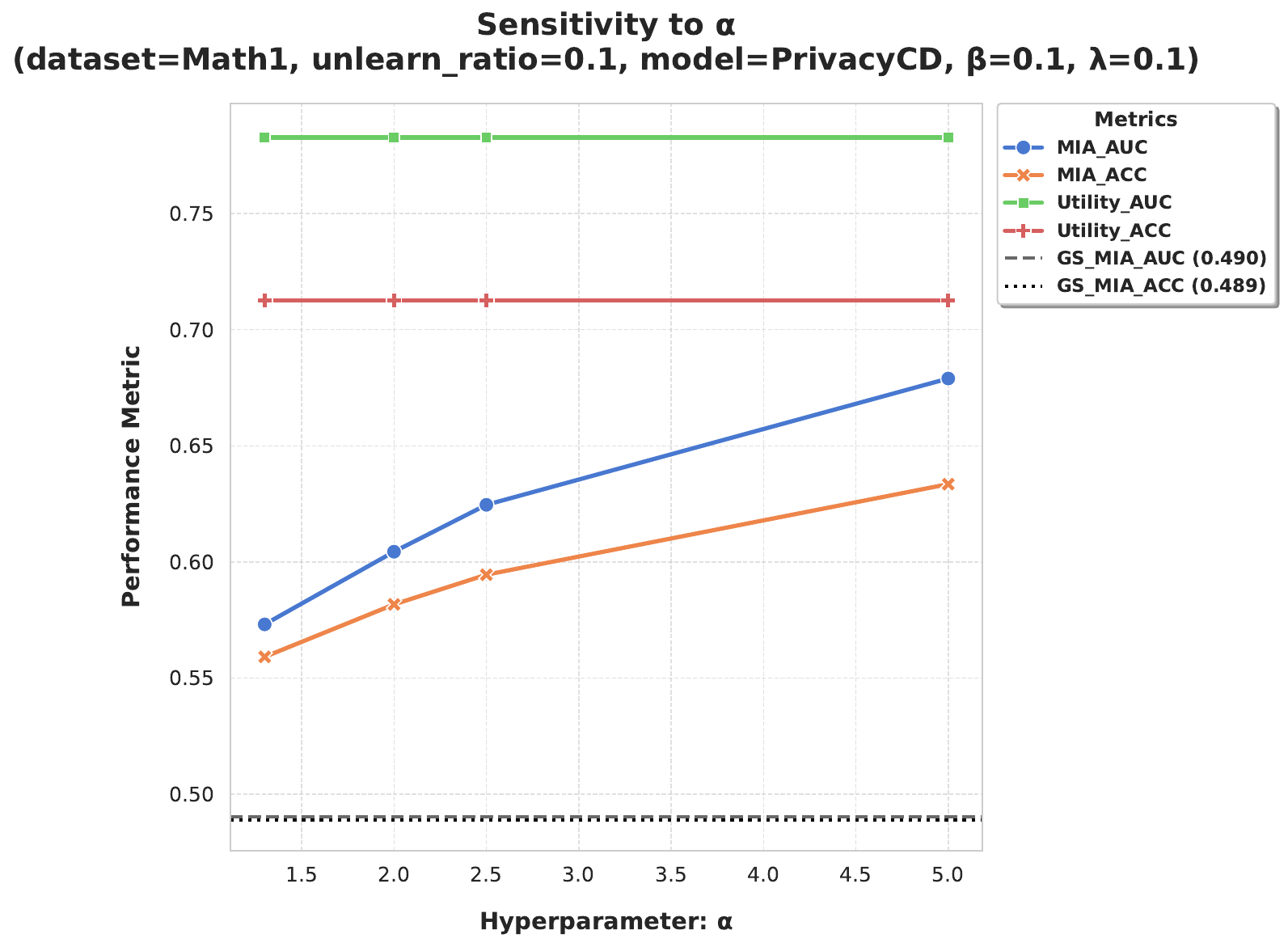}
			\subcaption{Sensitivity to $\alpha$}
		\end{minipage}
		\begin{minipage}[t]{0.31\textwidth}
			\centering
			\includegraphics[width=5.5cm]{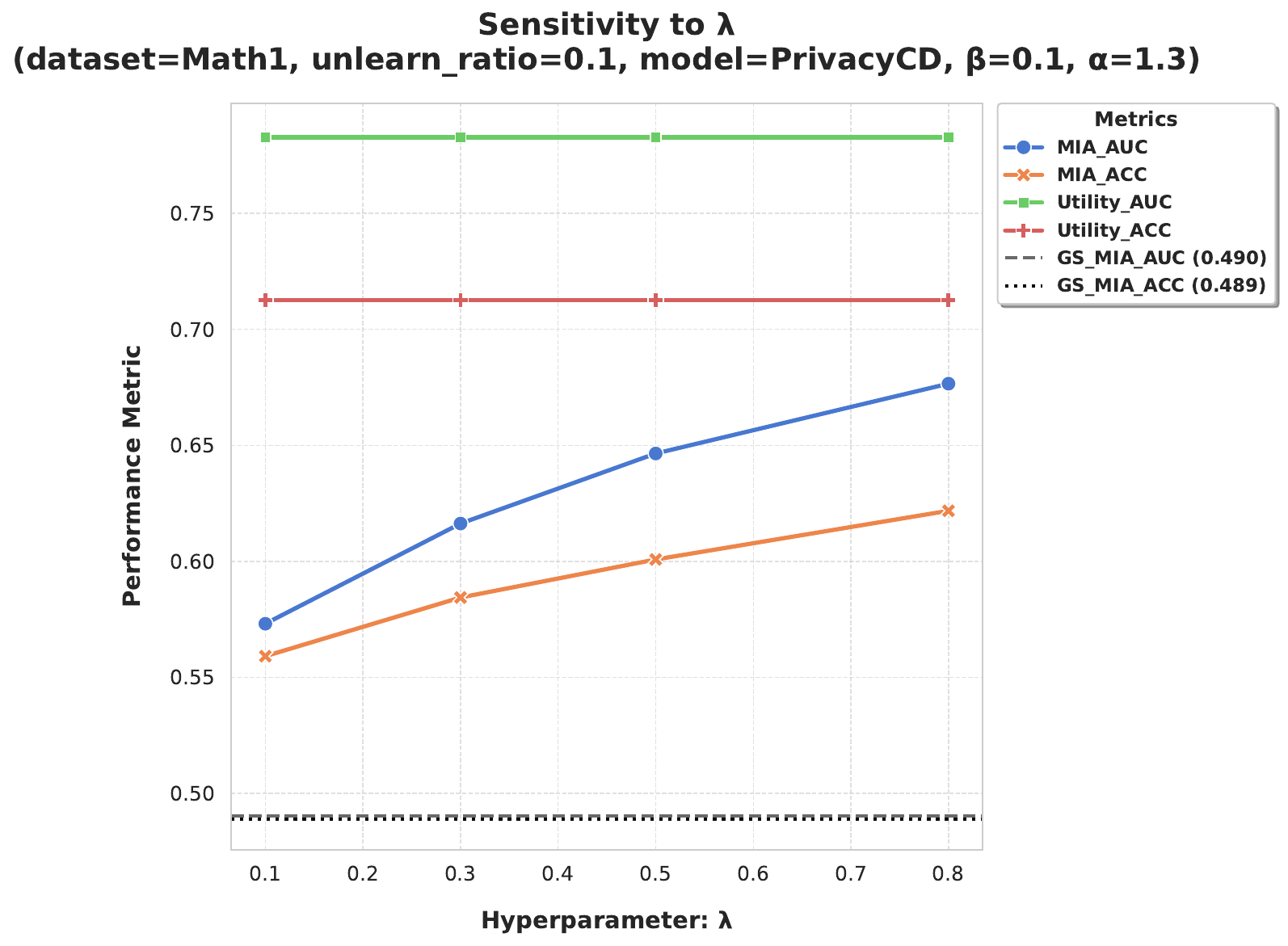}
			\subcaption{Sensitivity to $\lambda$}
		\end{minipage}
		\begin{minipage}[t]{0.31\textwidth}
			\centering
			\includegraphics[width=5.5cm]{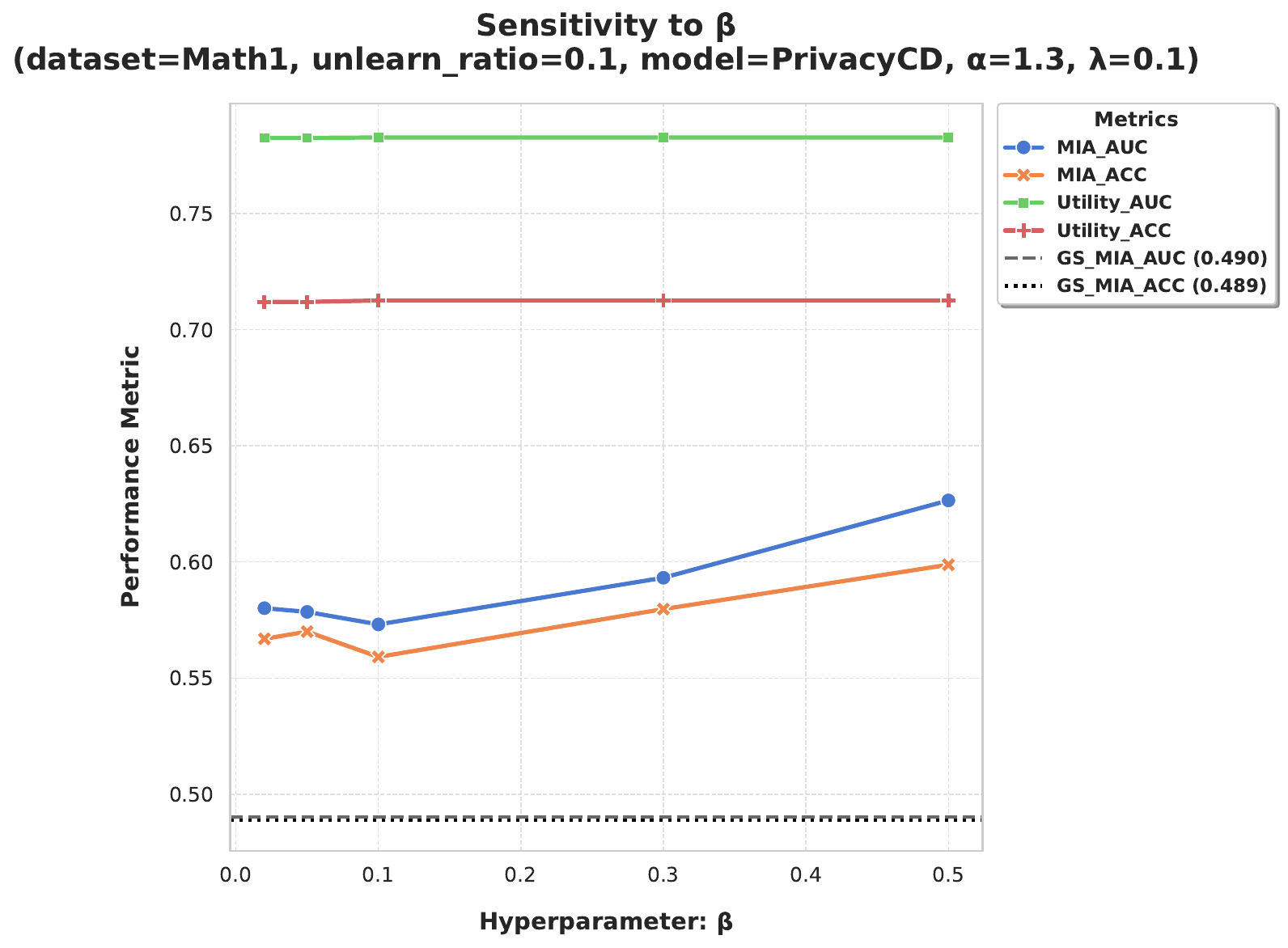}
			\subcaption{Sensitivity to $\beta$}
		\end{minipage}
		\caption{Parameter sensitivity analysis of our HIF algorithm on the Math1 dataset with a 10\% unlearning ratio.}
		\label{sa_math1_01}
\end{figure*}
\begin{figure*}[ht]
		\centering
		\begin{minipage}[t]{0.31\textwidth}
			\centering
			\includegraphics[width=5.5cm]{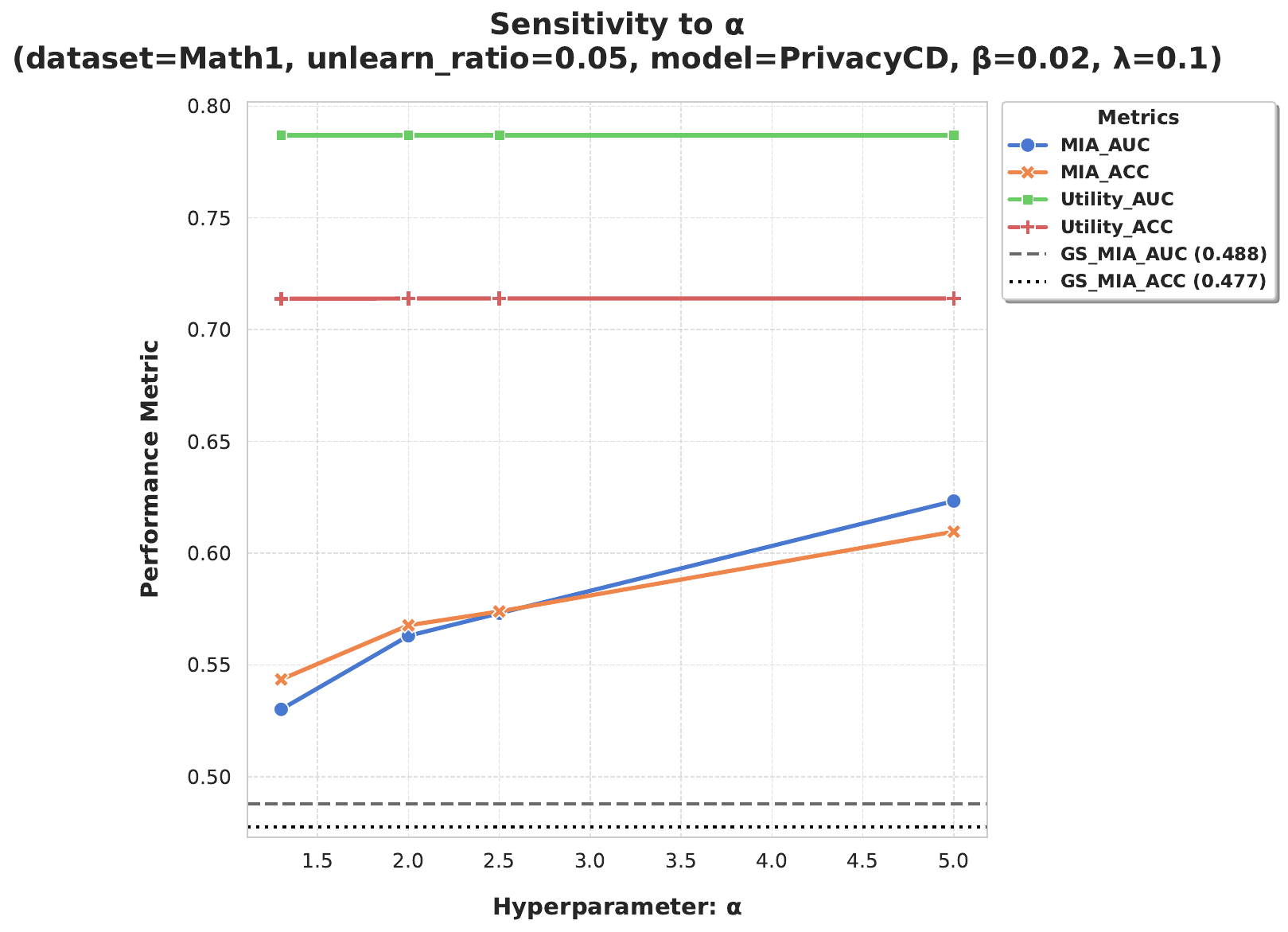}
			\subcaption{Sensitivity to $\alpha$}
		\end{minipage}
		\begin{minipage}[t]{0.31\textwidth}
			\centering
			\includegraphics[width=5.5cm]{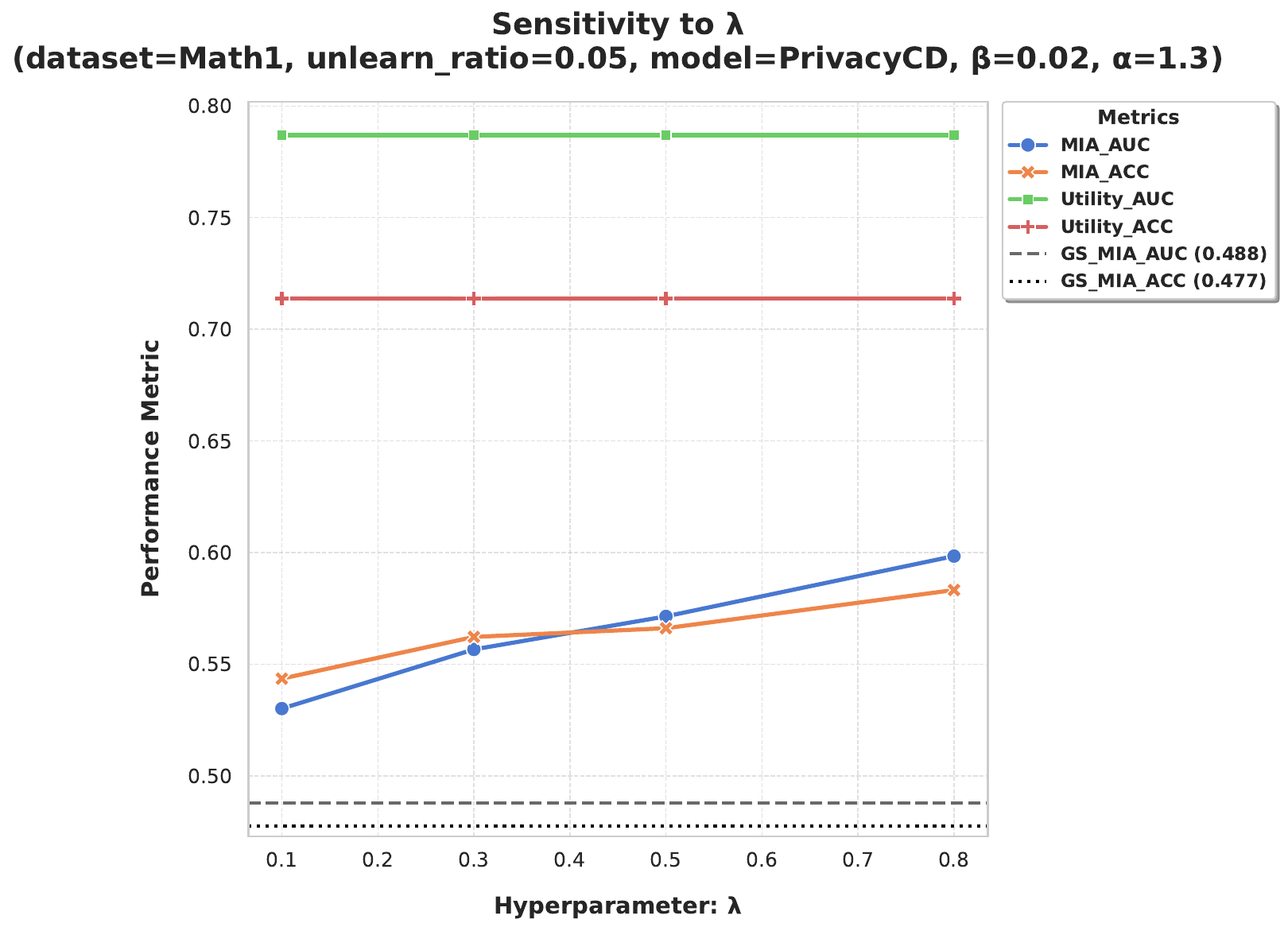}
			\subcaption{Sensitivity to $\lambda$}
		\end{minipage}
		\begin{minipage}[t]{0.31\textwidth}
			\centering
			\includegraphics[width=5.5cm]{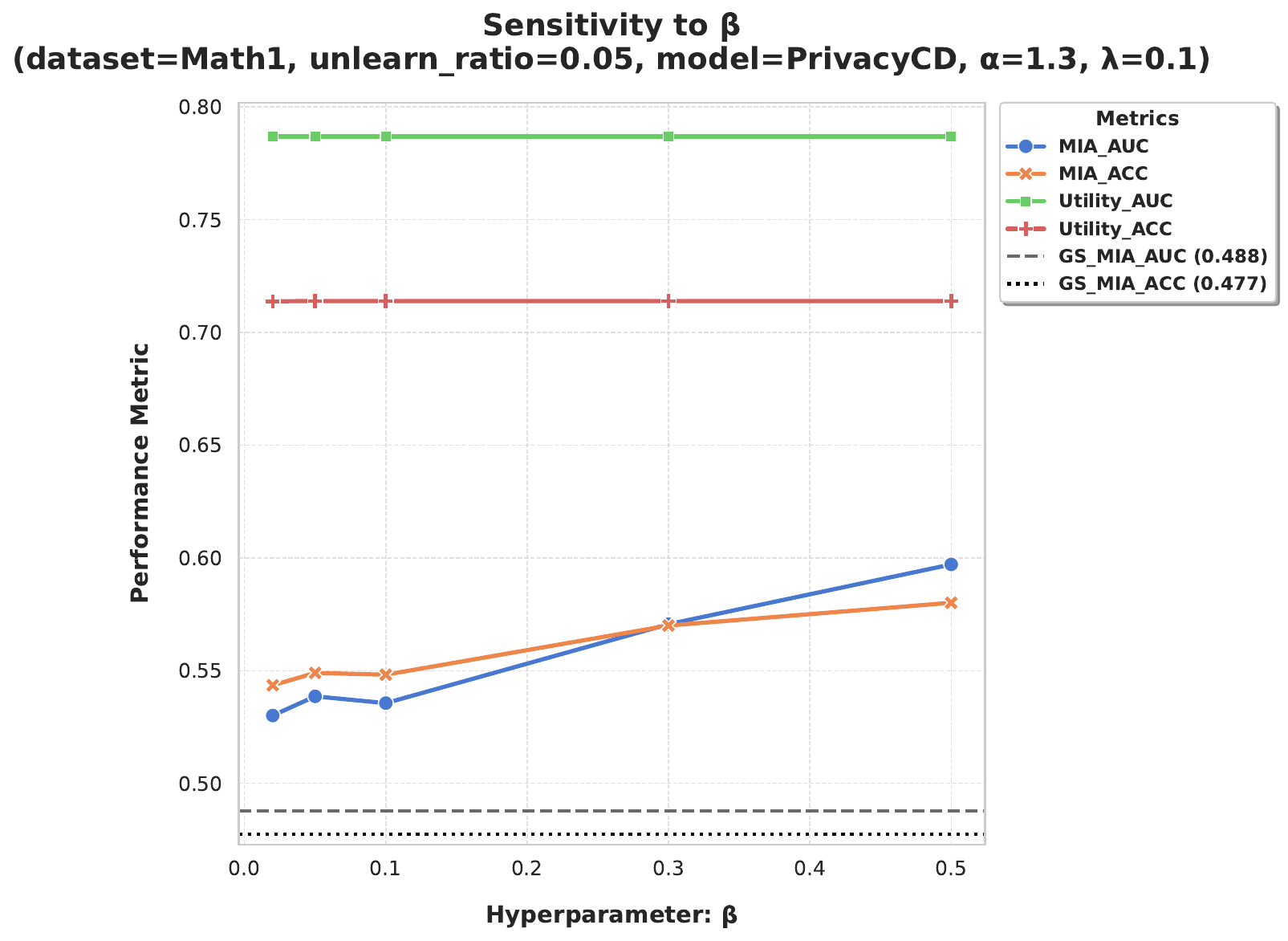}
			\subcaption{Sensitivity to $\beta$}
		\end{minipage}
		\caption{Parameter sensitivity analysis of our HIF algorithm on the Math1 dataset with a 5\% unlearning ratio.}
		\label{sa_math1_005}
\end{figure*}
\begin{figure*}[ht]
		\centering
		\begin{minipage}[t]{0.31\textwidth}
			\centering
			\includegraphics[width=5.5cm]{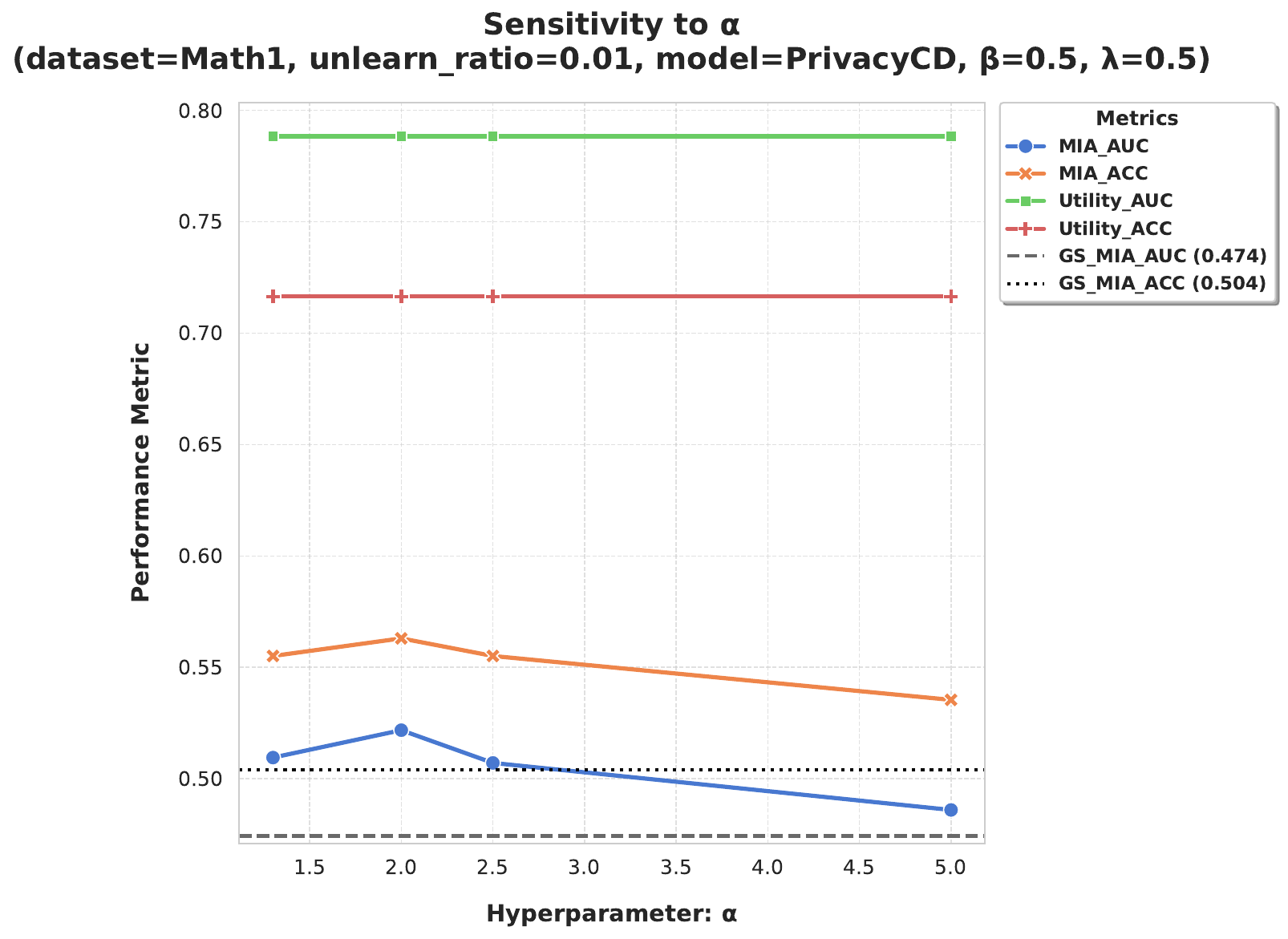}
			\subcaption{Sensitivity to $\alpha$}
		\end{minipage}
		\begin{minipage}[t]{0.31\textwidth}
			\centering
			\includegraphics[width=5.5cm]{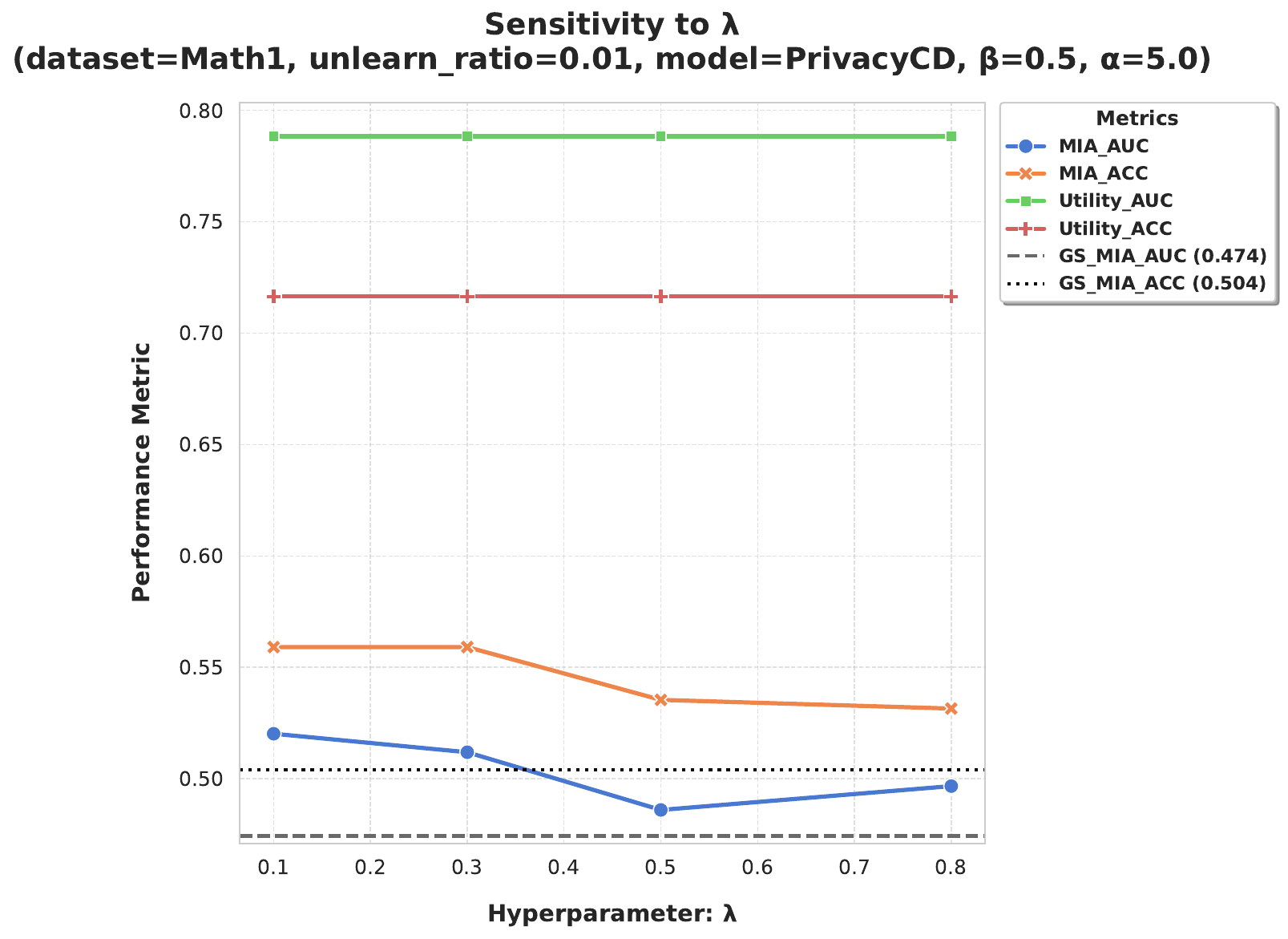}
			\subcaption{Sensitivity to $\lambda$}
		\end{minipage}
		\begin{minipage}[t]{0.31\textwidth}
			\centering
			\includegraphics[width=5.5cm]{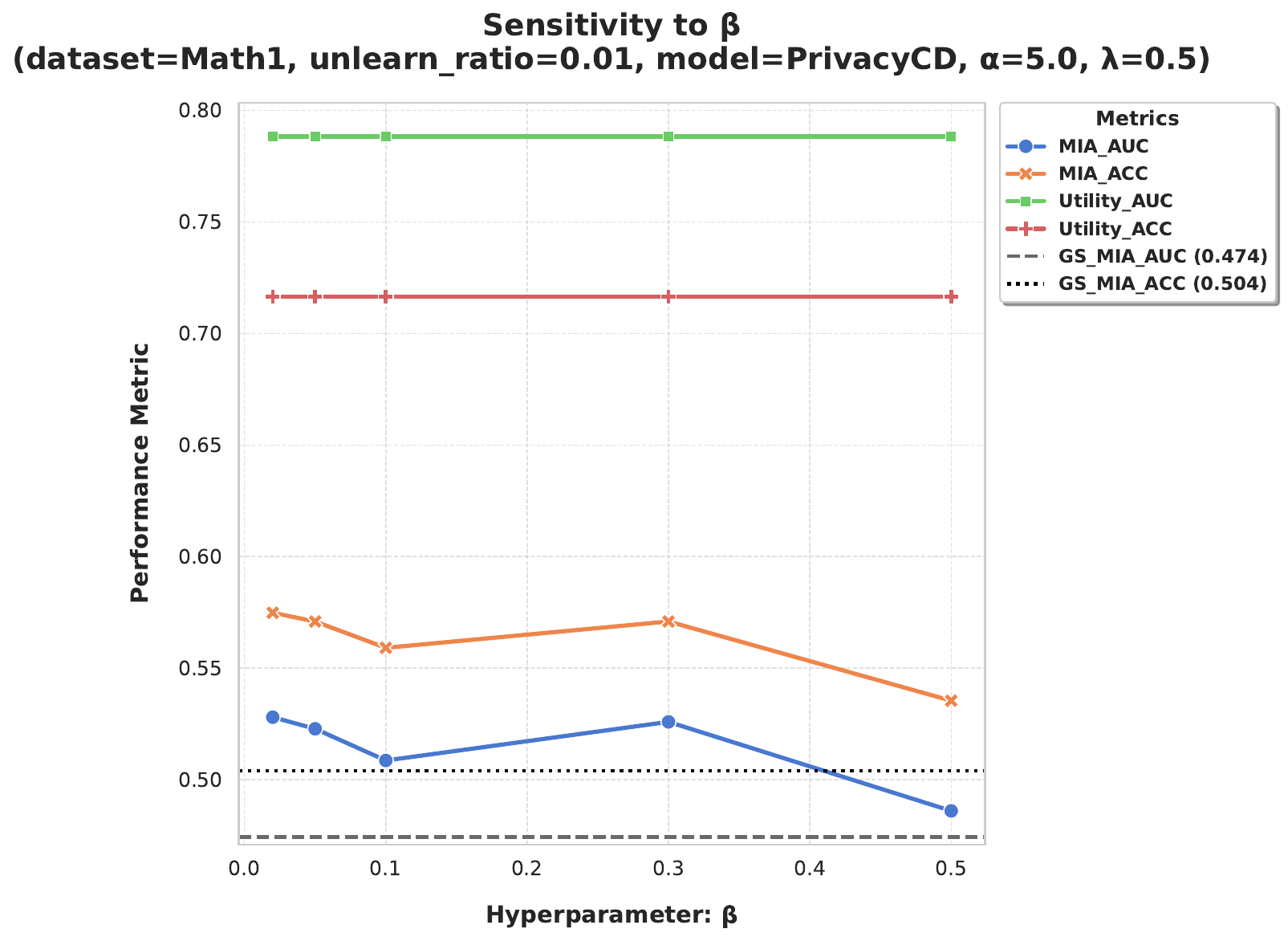}
			\subcaption{Sensitivity to $\beta$}
		\end{minipage}
		\caption{Parameter sensitivity analysis of our HIF algorithm on the Math1 dataset with a 1\% unlearning ratio.}
		\label{sa_math1_001}
\end{figure*}

\begin{figure*}[ht]
		\centering
		\begin{minipage}[t]{0.31\textwidth}
			\centering
			\includegraphics[width=5.5cm]{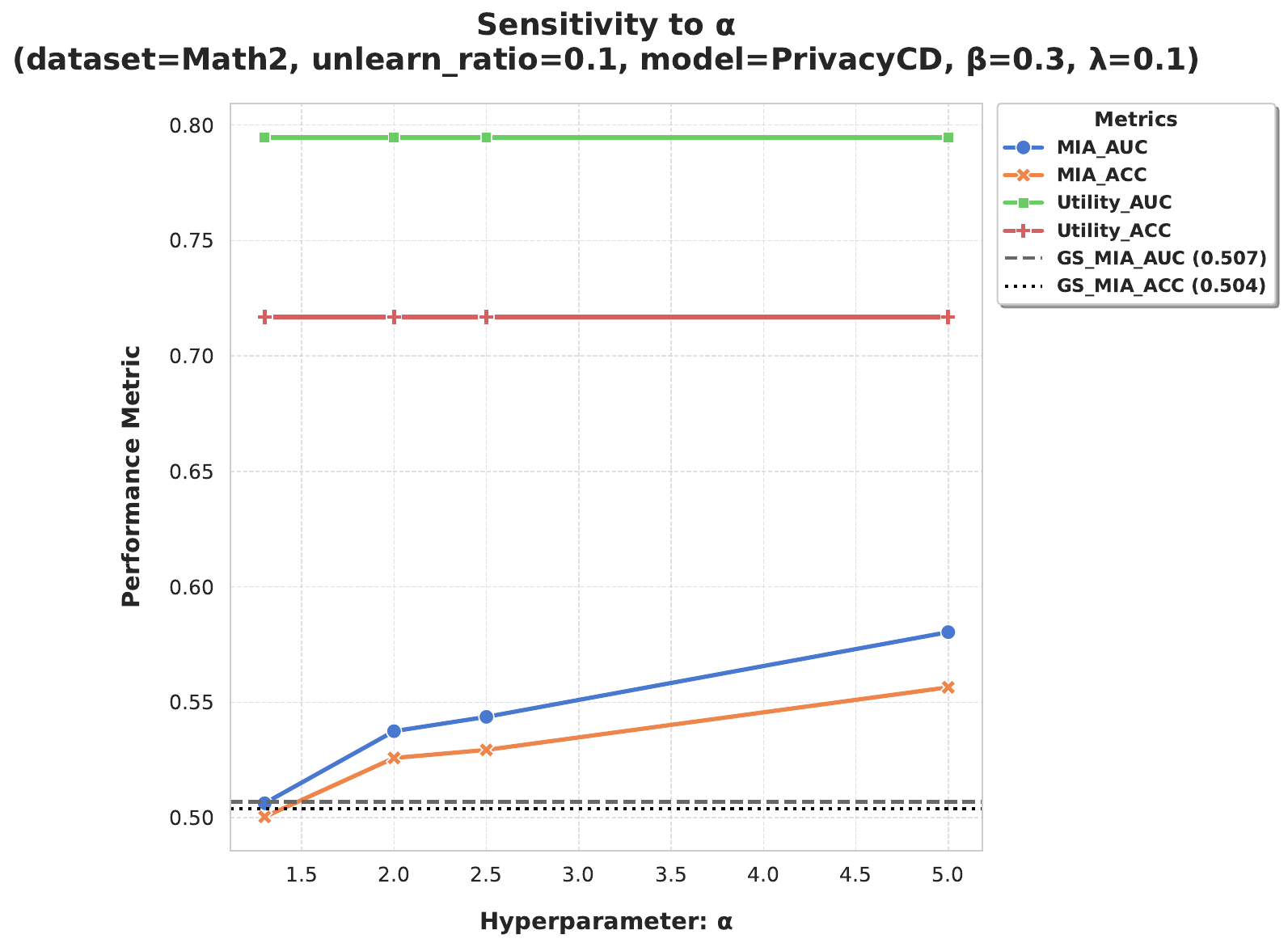}
			\subcaption{Sensitivity to $\alpha$}
		\end{minipage}
		\begin{minipage}[t]{0.31\textwidth}
			\centering
			\includegraphics[width=5.5cm]{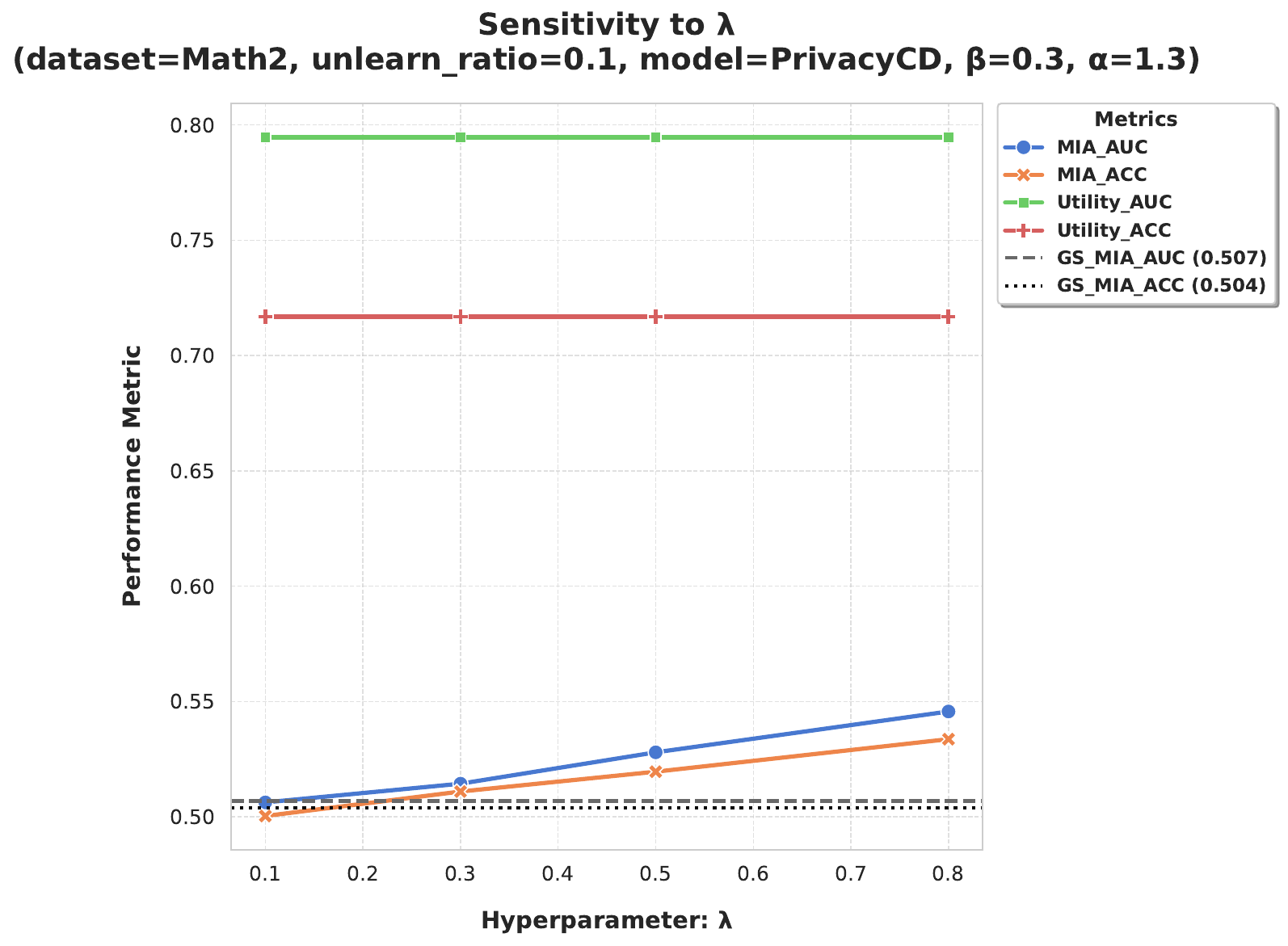}
			\subcaption{Sensitivity to $\lambda$}
		\end{minipage}
		\begin{minipage}[t]{0.31\textwidth}
			\centering
			\includegraphics[width=5.5cm]{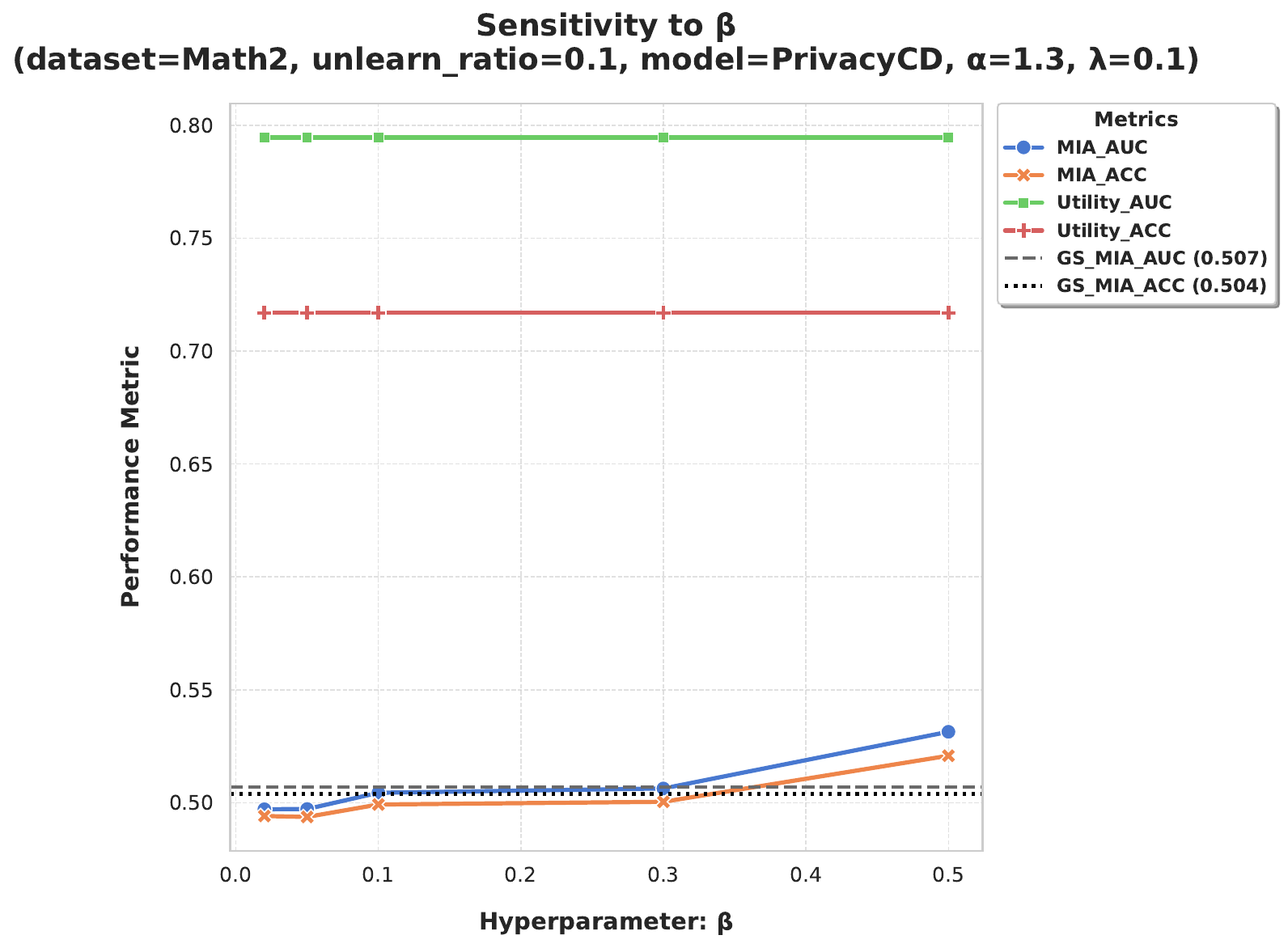}
			\subcaption{Sensitivity to $\beta$}
		\end{minipage}
		\caption{Parameter sensitivity analysis of our HIF algorithm on the Math2 dataset with a 10\% unlearning ratio.}
		\label{sa_math2_01}
\end{figure*}

\begin{figure*}[ht]
		\centering
		\begin{minipage}[t]{0.31\textwidth}
			\centering
			\includegraphics[width=5.5cm]{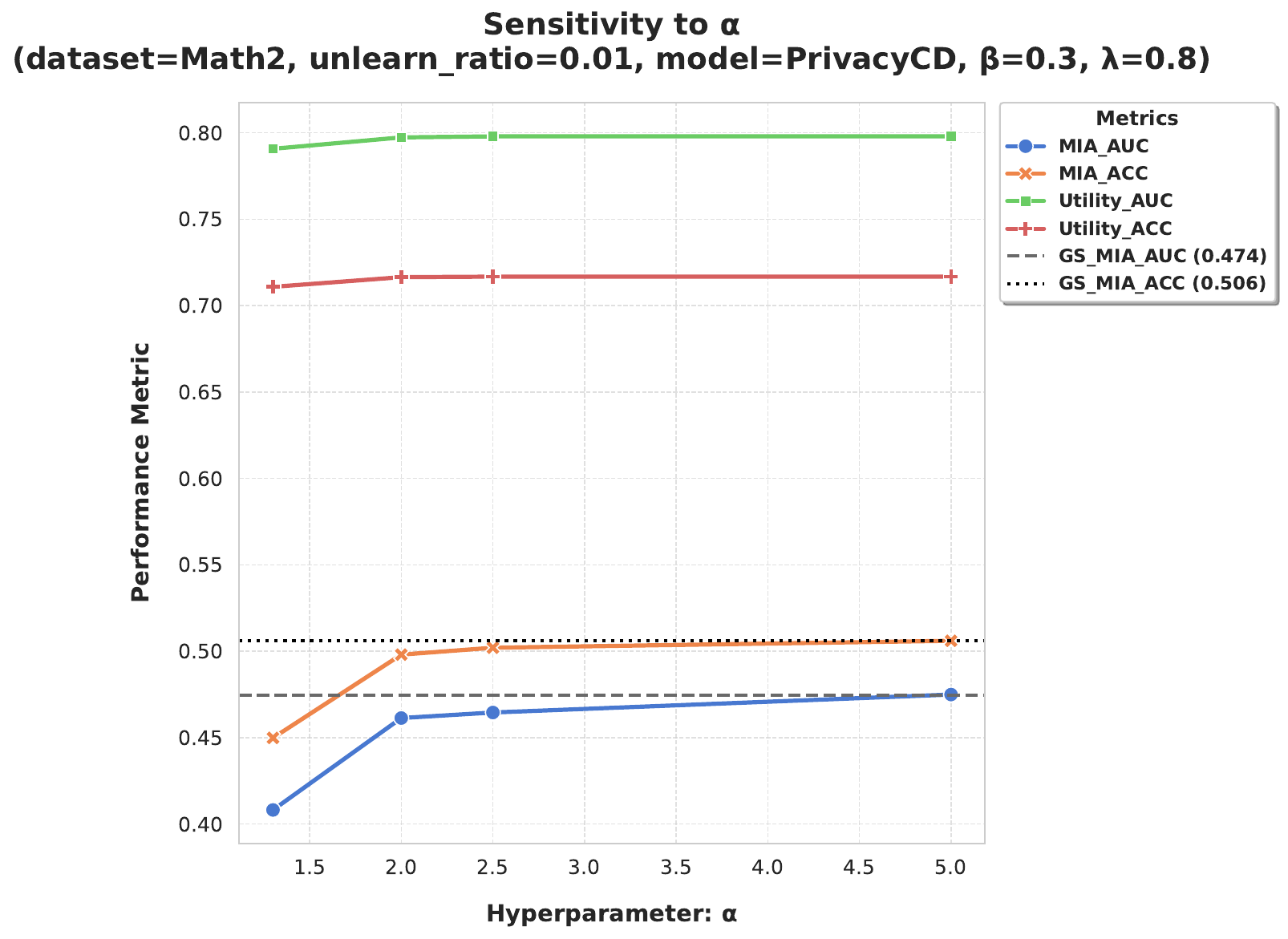}
			\subcaption{Sensitivity to $\alpha$}
		\end{minipage}
		\begin{minipage}[t]{0.31\textwidth}
			\centering
			\includegraphics[width=5.5cm]{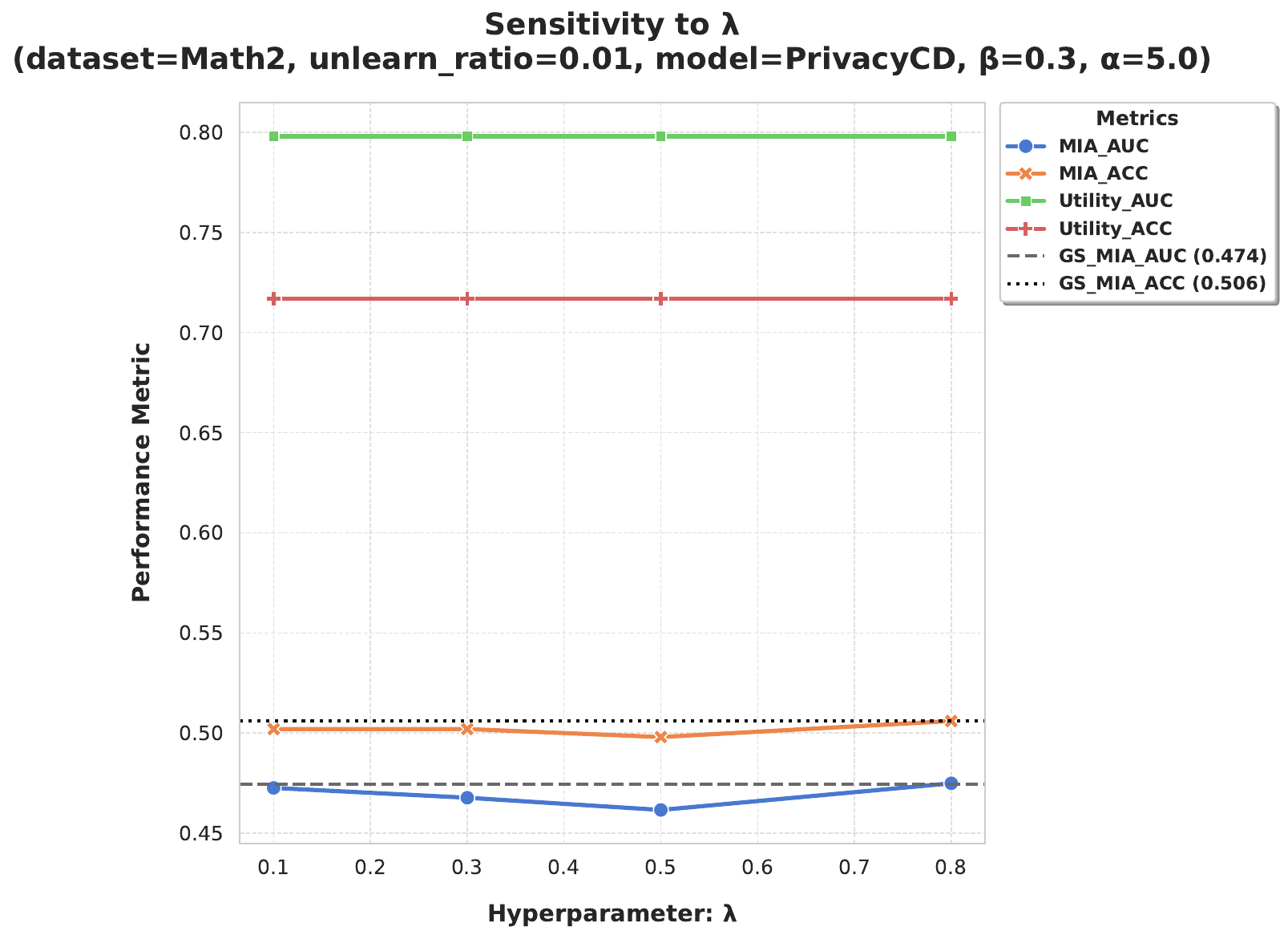}
			\subcaption{Sensitivity to $\lambda$}
		\end{minipage}
		\begin{minipage}[t]{0.31\textwidth}
			\centering
			\includegraphics[width=5.5cm]{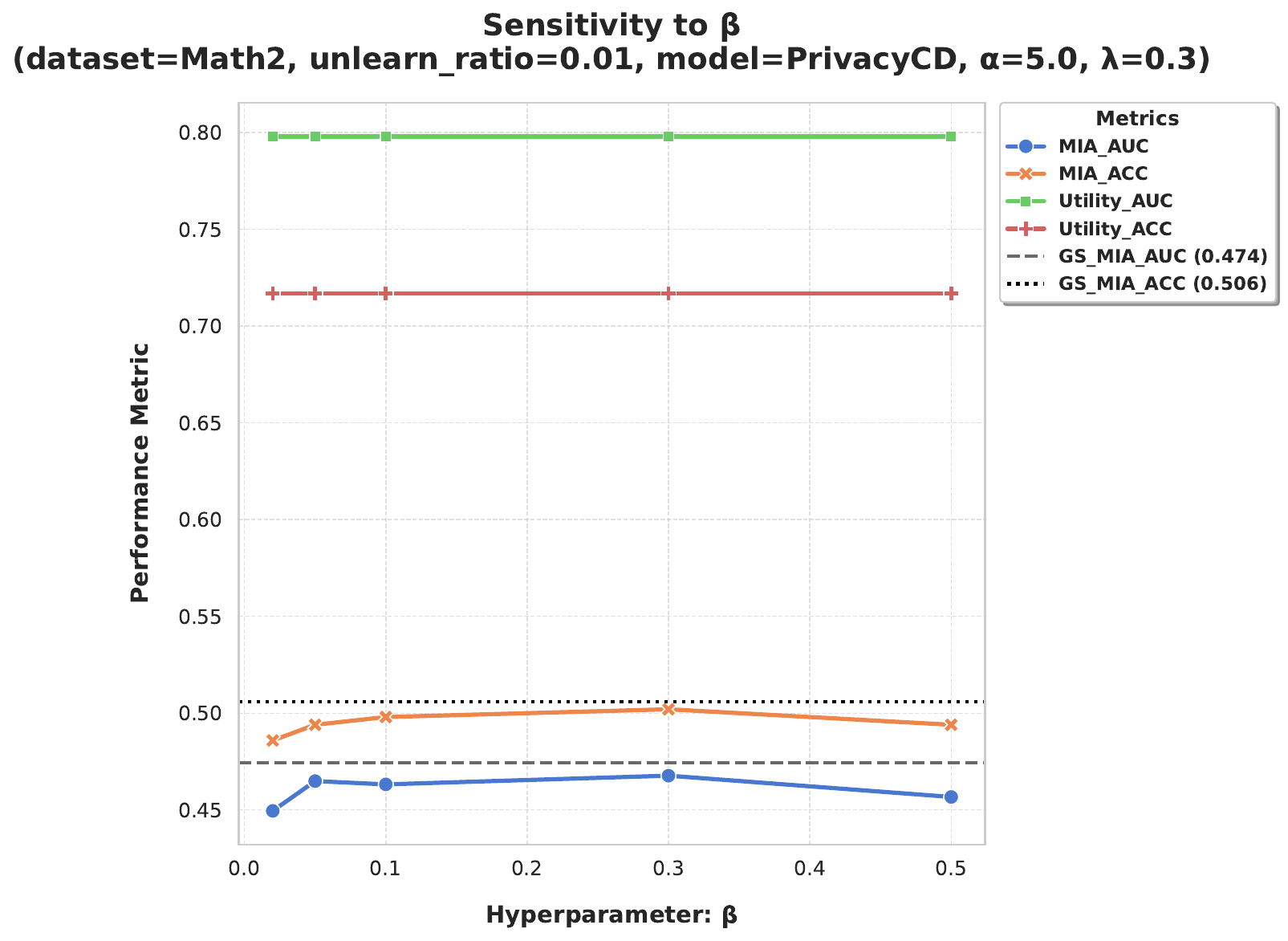}
			\subcaption{Sensitivity to $\beta$}
		\end{minipage}
		\caption{Parameter sensitivity analysis of our HIF algorithm on the Math2 dataset with a 1\% unlearning ratio.}
		\label{sa_math2_001}
\end{figure*}

\begin{figure*}[ht]
		\centering
		\begin{minipage}[t]{0.31\textwidth}
			\centering
			\includegraphics[width=5.5cm]{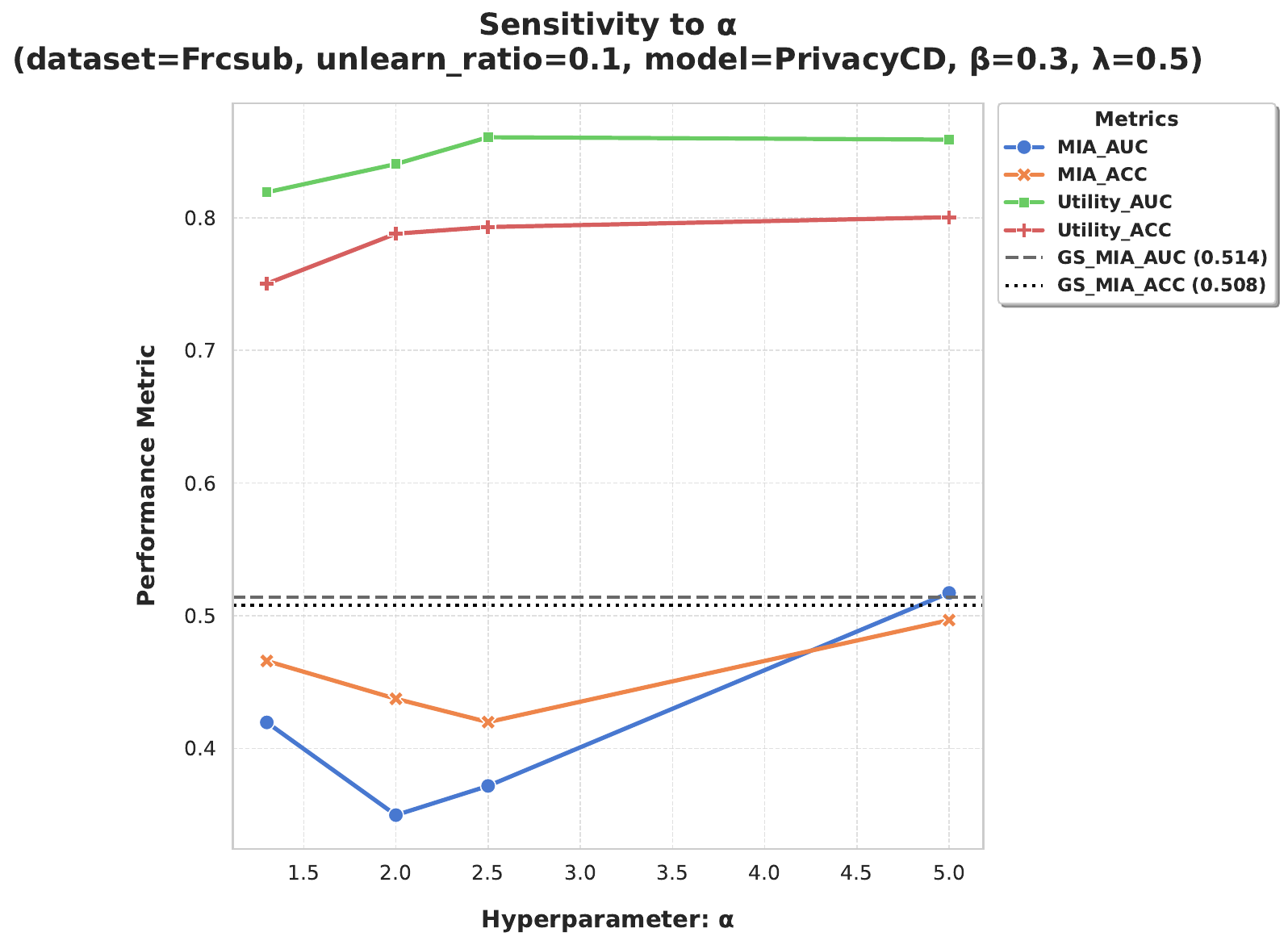}
			\subcaption{Sensitivity to $\alpha$}
		\end{minipage}
		\begin{minipage}[t]{0.31\textwidth}
			\centering
			\includegraphics[width=5.5cm]{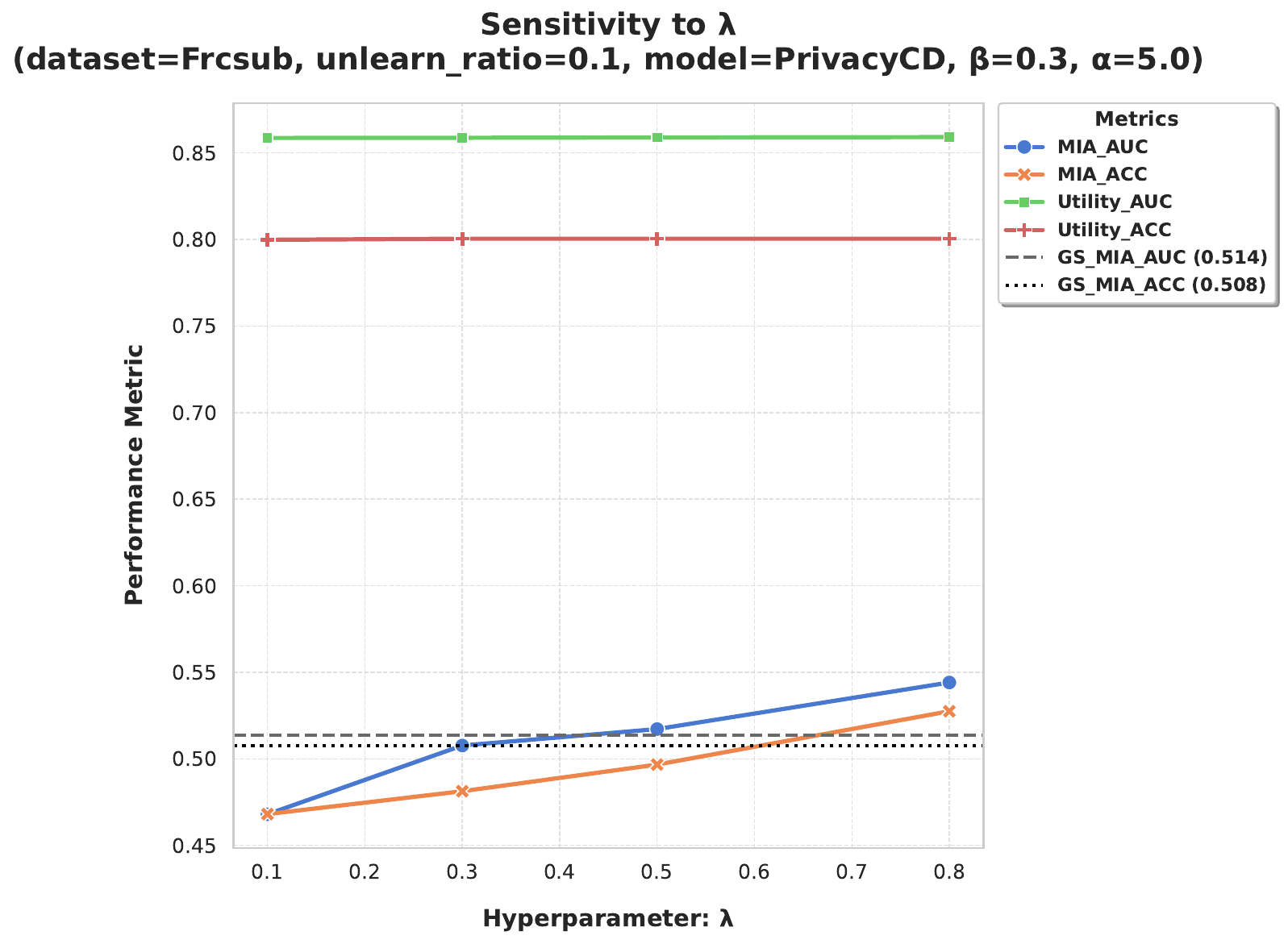}
			\subcaption{Sensitivity to $\lambda$}
		\end{minipage}
		\begin{minipage}[t]{0.31\textwidth}
			\centering
			\includegraphics[width=5.5cm]{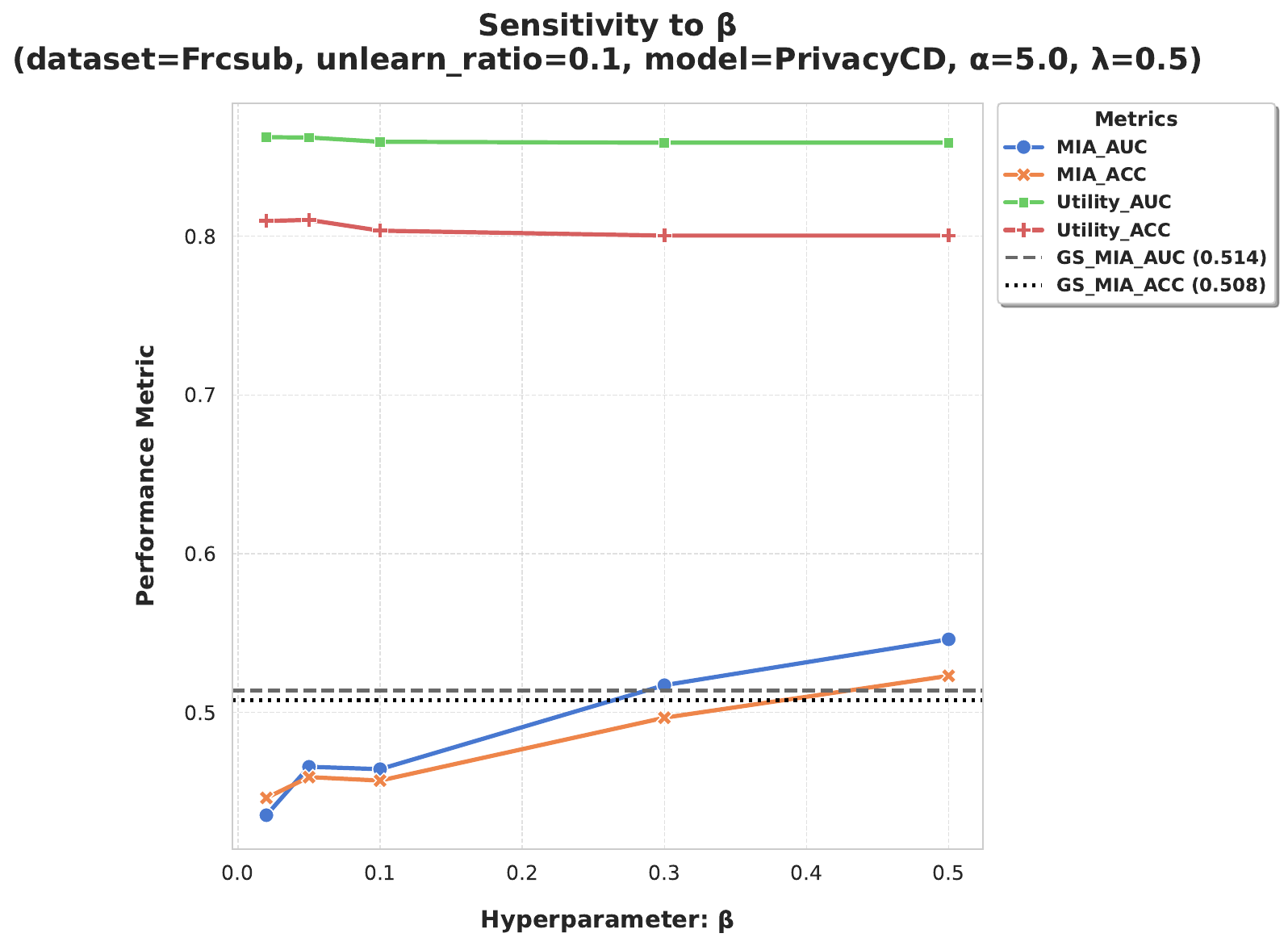}
			\subcaption{Sensitivity to $\beta$}
		\end{minipage}
		\caption{Parameter sensitivity analysis of our HIF algorithm on the Frcsub dataset with a 10\% unlearning ratio.}
		\label{sa_frcsub_01}
\end{figure*}
\begin{figure*}[ht]
		\centering
		\begin{minipage}[t]{0.31\textwidth}
			\centering
			\includegraphics[width=5.5cm]{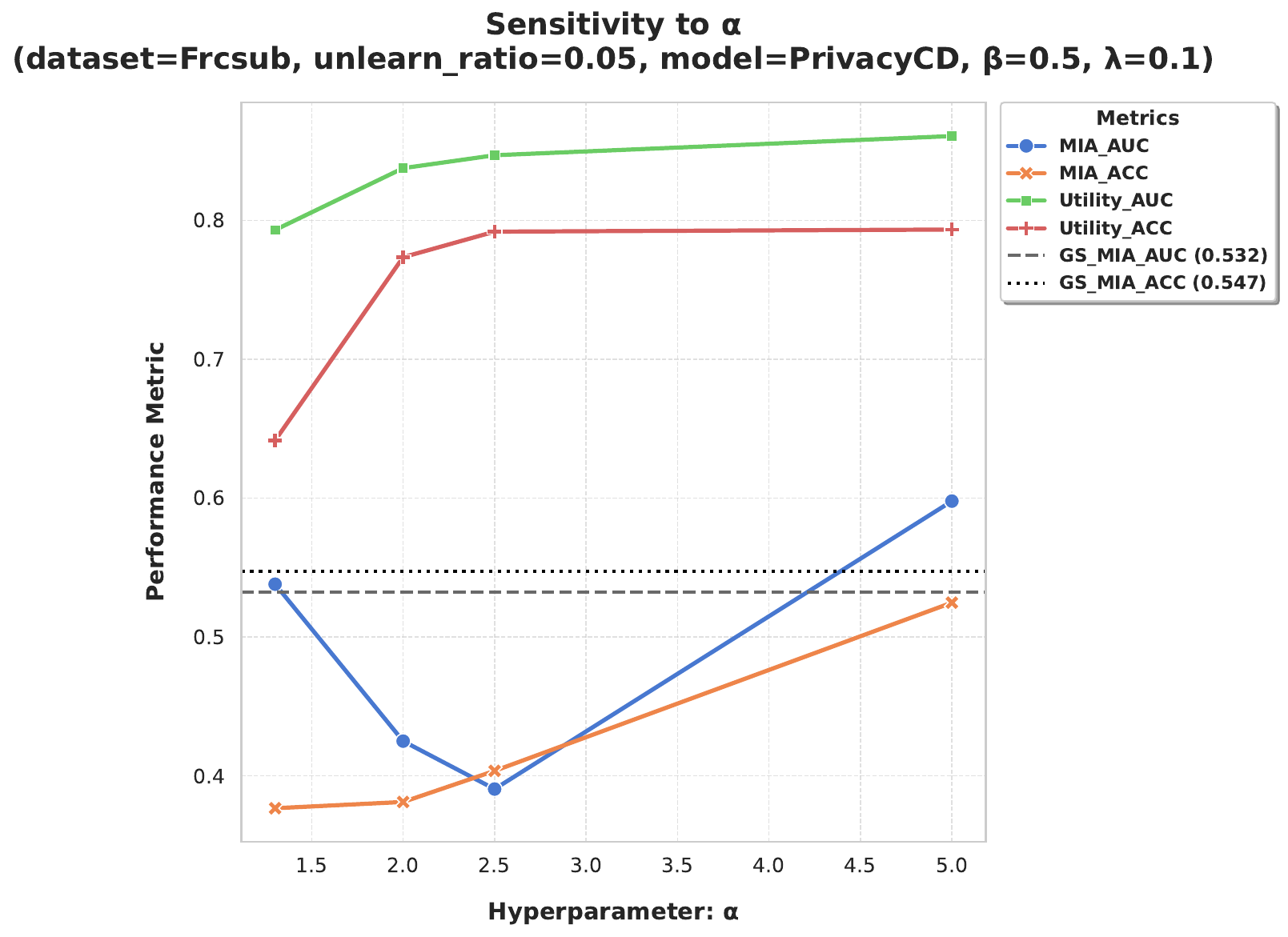}
			\subcaption{Sensitivity to $\alpha$}
		\end{minipage}
		\begin{minipage}[t]{0.31\textwidth}
			\centering
			\includegraphics[width=5.5cm]{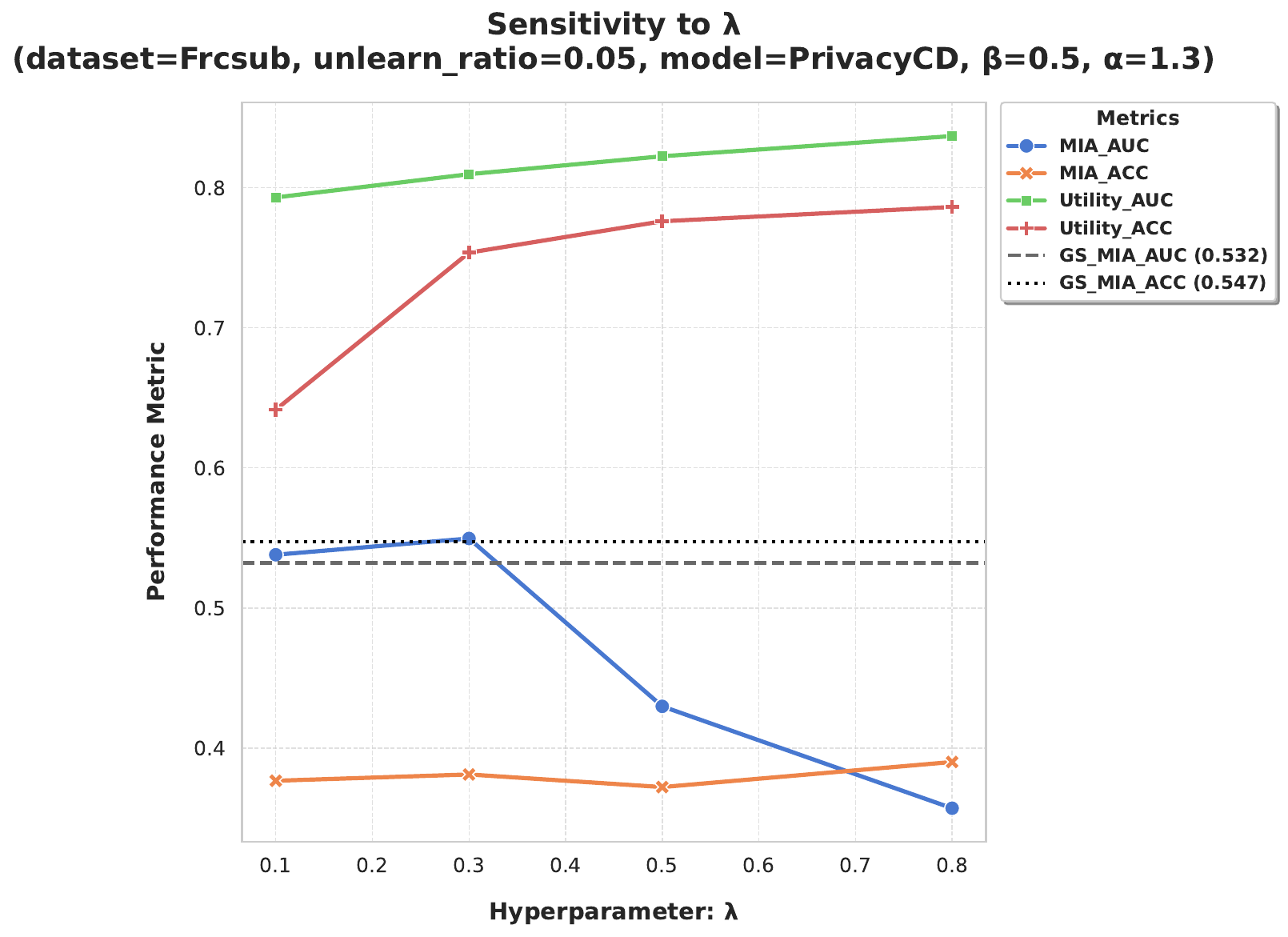}
			\subcaption{Sensitivity to $\lambda$}
		\end{minipage}
		\begin{minipage}[t]{0.31\textwidth}
			\centering
			\includegraphics[width=5.5cm]{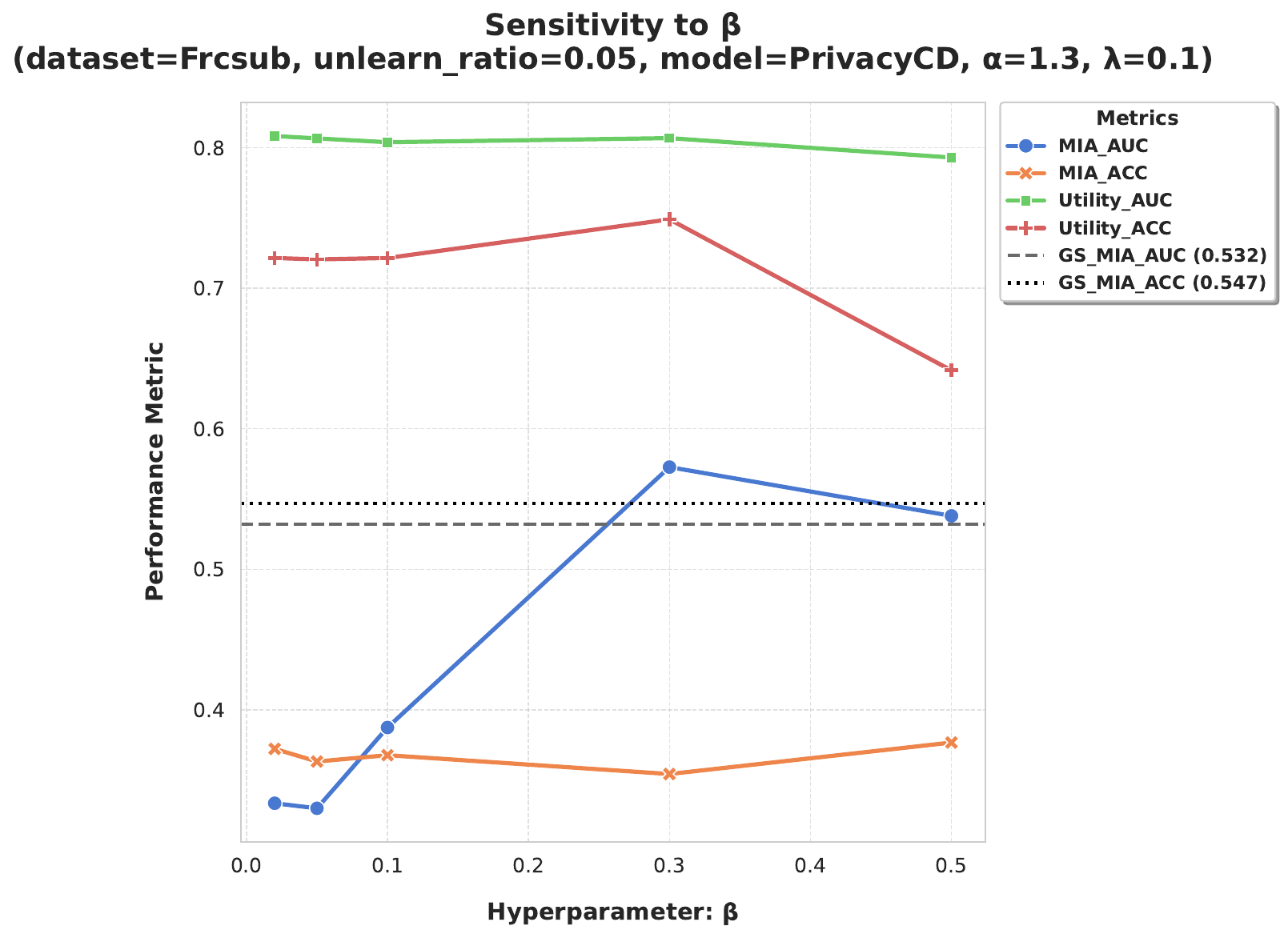}
			\subcaption{Sensitivity to $\beta$}
		\end{minipage}
		\caption{Parameter sensitivity analysis of our HIF algorithm on the Frcsub dataset with a 5\% unlearning ratio.}
		\label{sa_frcsub_005}
\end{figure*}
\begin{figure*}[ht]
		\centering
		\begin{minipage}[t]{0.31\textwidth}
			\centering
			\includegraphics[width=5.5cm]{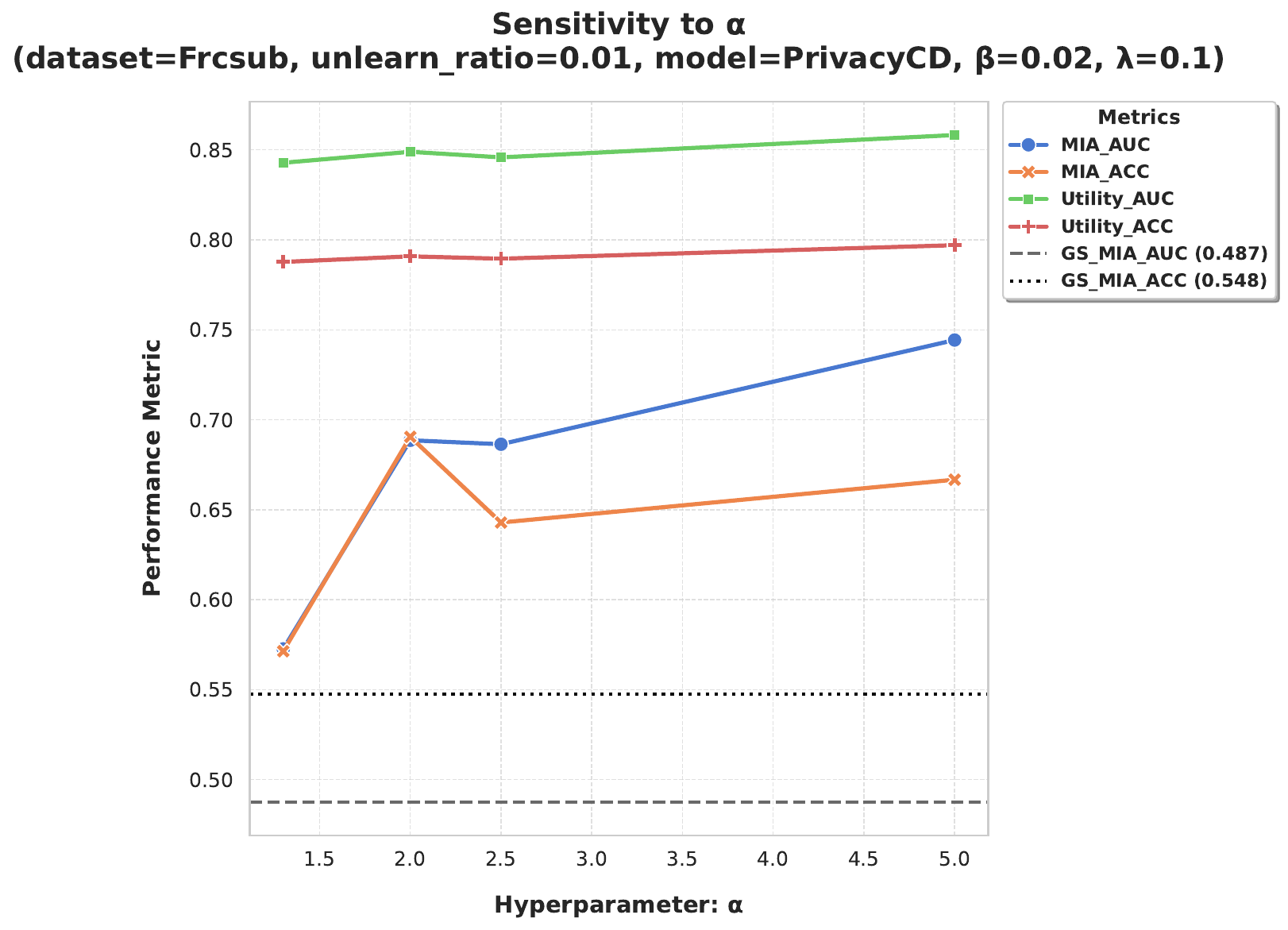}
			\subcaption{Sensitivity to $\alpha$}
		\end{minipage}
		\begin{minipage}[t]{0.31\textwidth}
			\centering
			\includegraphics[width=5.5cm]{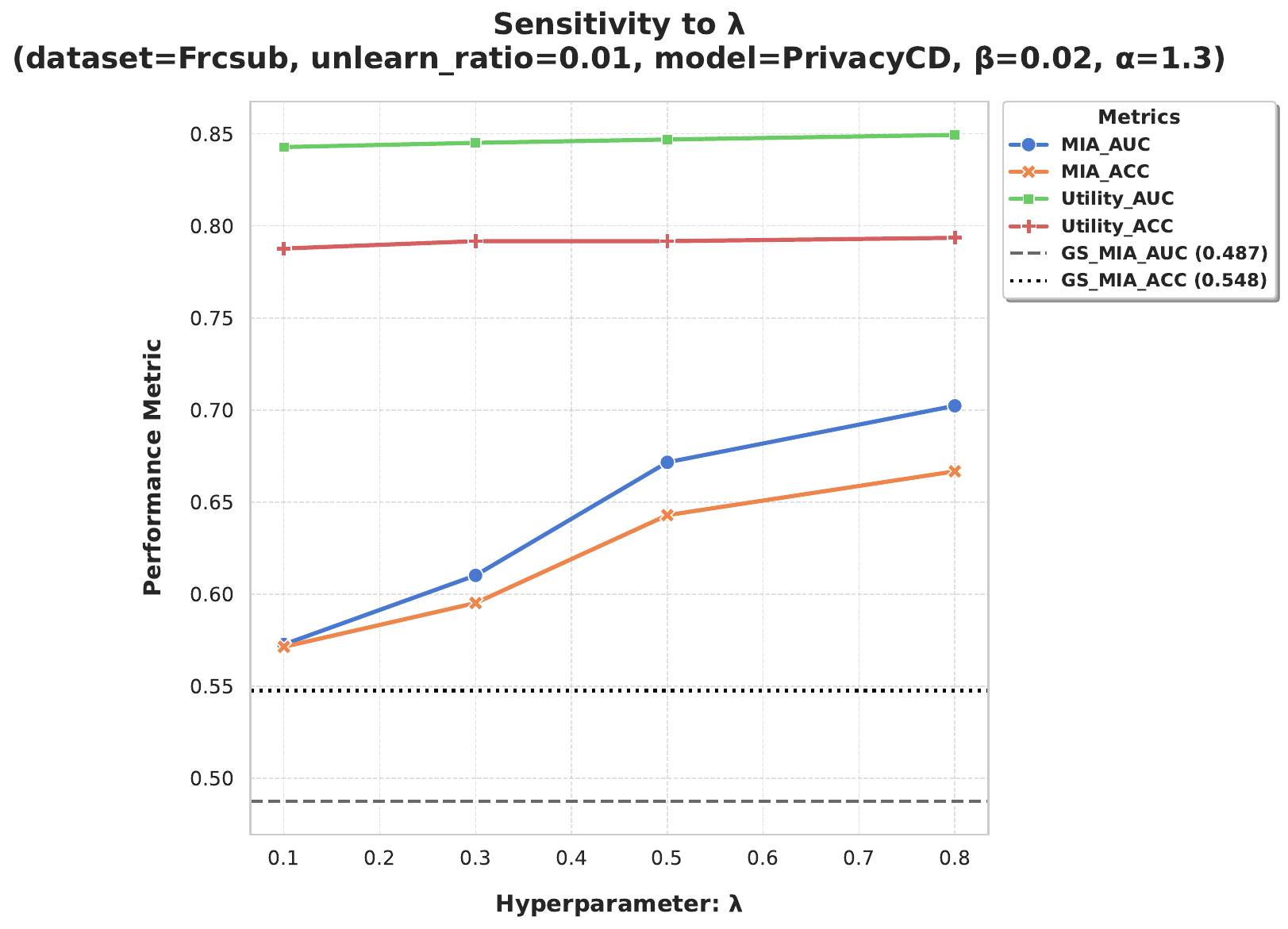}
			\subcaption{Sensitivity to $\lambda$}
		\end{minipage}
		\begin{minipage}[t]{0.31\textwidth}
			\centering
			\includegraphics[width=5.5cm]{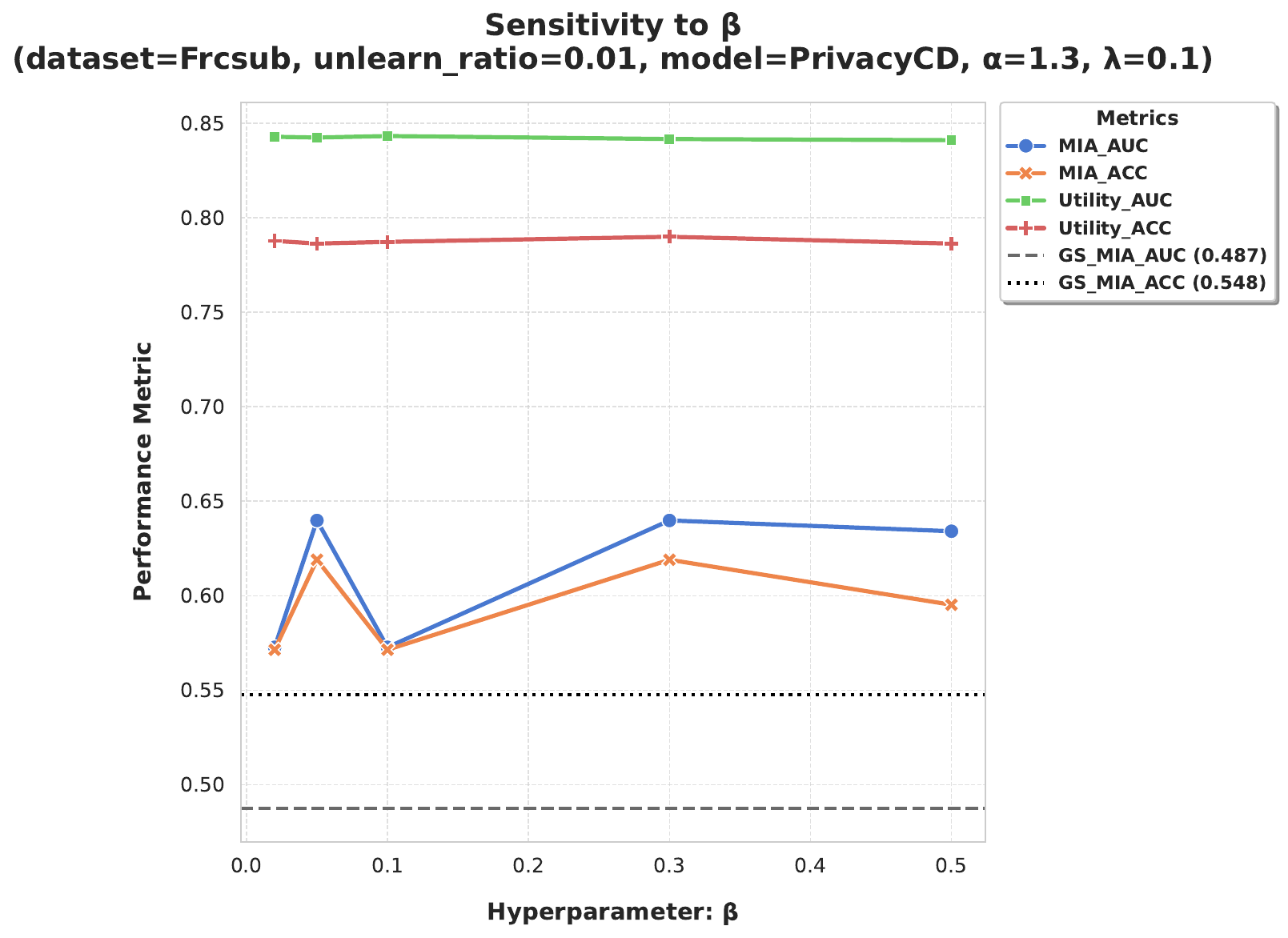}
			\subcaption{Sensitivity to $\beta$}
		\end{minipage}
		\caption{Parameter sensitivity analysis of our HIF algorithm on the Frcsub dataset with a 1\% unlearning ratio.}
		\label{sa_frcsub_001}
\end{figure*}
\section{Supplementary Information on Experimental Results}
\subsection{Detailed Ablation Study Results on Different Architectures}
This section provides the complete results for our ablation study, where the HIF algorithm was applied to two alternative neural CD architectures: NeuralCD and KSCD. The results across all datasets and unlearning ratios are presented in Table~\ref{ab_neuralcd} and Table~\ref{ab_kscd}, respectively.

The complete experimental results reinforce the two key findings discussed in the main paper.

First, the results consistently demonstrate the general effectiveness of the HIF algorithm across different underlying model architectures. In nearly all experimental settings shown in Table~\ref{ab_neuralcd} and Table~\ref{ab_kscd}, applying HIF to both NeuralCD and KSCD successfully maintains model utility at a level almost identical to the original model, while significantly reducing the MIA metrics. This confirms that HIF is a robust and broadly applicable unlearning strategy, not one that is over-tuned to a single specific architecture.

Second, and more importantly, these detailed results provide strong evidence for our central hypothesis that model architecture is a critical factor for unlearning performance. When comparing the unlearning efficacy of HIF across the three architectures (our primary decoupled model in Table~\ref{overallres}, NeuralCD in Table~\ref{ab_neuralcd}, and KSCD in Table~\ref{ab_neuralcd}), a clear trend emerges: HIF consistently achieves its best performance, i.e., MIA scores closest to the gold-standard retrain model, on our primary decoupled architecture. For example, on the Math1 dataset with a 1\% unlearning ratio, the MIA\_AUC for HIF is 0.486 on our main architecture, while it is 0.5175 on NeuralCD and 0.607 on KSCD. This performance gap highlights that the entangled representations in NeuralCD and the complex, multi-stage interactions in KSCD create a more challenging environment for importance-based unlearning. The signals of what to forget are more diffuse and harder to isolate in these architectures.

In conclusion, this comprehensive ablation study validates that while HIF is a potent unlearning algorithm in its own right, its full potential is unlocked when paired with an ``unlearning-friendly'' architecture that adheres to the principles of modularity and representation decoupling.
\subsection{Detailed Parameter Sensitivity Analysis}
This section provides a comprehensive analysis of the sensitivity of our algorithm to its three key hyperparameters: the selection threshold $\alpha$, the unlearning strength $\lambda$, and the hierarchical smoothing factor $\beta$. We present the results across all three datasets (Math1, Math2, and Frcsub) and all three unlearning ratios (10\%, 5\%, and 1\%) to offer a complete picture of their impact on model utility (Utility\_AUC, Utility\_ACC) and unlearning efficacy (MIA\_AUC, MIA\_ACC). The full set of plots is provided in Figures~\ref{sa_math1_01} -~\ref{sa_frcsub_001}.

\subsubsection{Effect of the Smoothing Factor $\beta$}The smoothing factor $\beta$ is the core component of our algorithm. Our analysis across all nine experimental settings consistently and robustly validates its effectiveness.
\begin{itemize}
    \item \textbf{Universal Effectiveness:} In every single case, we observe the existence of an optimal $\beta > 0$ that yields a better unlearning efficacy (lower MIA scores) than the naive FIM baseline ($\beta=0$). As shown in Figure~\ref{fig:sensitivity_analysis}(c) and the appendix figures, as $\beta$ increases from 0 to a small positive value (typically in the range of 0.02 to 0.1), the MIA metrics consistently exhibit a significant drop, often reaching their minimum in this range.
    
    \item \textbf{Utility Preservation:} Critically, this significant improvement in unlearning efficacy is achieved with minimal to no degradation in model utility. The Utility\_AUC and Utility\_ACC curves remain largely flat or even slightly increase in the optimal $\beta$ range, providing strong empirical evidence for our theoretical argument that hierarchical smoothing reduces estimation noise without harming generalizable knowledge.
    
    \item \textbf{Robustness:} While the exact optimal value of $\beta$ varies slightly with the dataset and unlearning ratio, the beneficial effect of a small, positive $\beta$ is a universal phenomenon across all our experiments. This demonstrates the robustness of the ``hierarchical wisdom'' concept.
\end{itemize}
\subsubsection{Effect of the Selection Threshold $\alpha$}The parameter $\alpha$ controls the strictness of the criterion for selecting parameters to be modified.
\begin{itemize}
    \item \textbf{General Trend:} In most scenarios (e.g., on the Math1 and Math2 datasets), we observe a clear and predictable trend: as $\alpha$ increases, the MIA scores also tend to increase. This is intuitive, as a higher $\alpha$ means only parameters deemed highly important to the forget set are selected, potentially leaving behind the influence of moderately important ones and thus resulting in a less thorough unlearning.
    
    \item \textbf{Utility Stability:} Across all experiments, the model's utility remains remarkably stable and is almost entirely insensitive to the value of $\alpha$. This is a highly desirable property, as it suggests that the parameters selected by our criterion are indeed primarily related to the forget set and not critical for the model's overall performance on the retain set.
    
    \item \textbf{Anomalous Behavior on Frcsub:} On the Frcsub dataset, the trend is less monotonic. For instance, at the 10\% and 5\% ratios, the MIA\_AUC first decreases and then increases as $\alpha$ grows. This suggests that on datasets with different characteristics, an intermediate level of selection strictness might be optimal, possibly by filtering out noisy, low-importance parameters more effectively.
\end{itemize}
\subsubsection{Effect of the Unlearning Strength $\lambda$}The parameter $\lambda$ directly controls the magnitude of attenuation applied to the selected parameters. The analysis of $\lambda$ reveals the most complex and context-dependent behavior.
\begin{itemize}
    \item \textbf{Context-Dependent Behavior:} We identified two distinct behavioral regimes for $\lambda$. When the unlearning ratio is small (1\% on Math1 and Math2), increasing $\lambda$ generally leads to better unlearning (lower MIA scores), which aligns with the intuition that stronger attenuation leads to more forgetting.
    
    \item \textbf{Counter-intuitive Trend at Higher Ratios:} However, for larger unlearning ratios (5\% and 10\% on Math1 and Math2), we consistently observe a counter-intuitive trend: increasing $\lambda$ beyond a certain point leads to worse unlearning efficacy (higher MIA scores). We hypothesize this is because when a large number of parameters are being modified, an overly aggressive attenuation ($\lambda \gg 0.5$) might drastically distort the parameter space, causing the unlearned model to behave in an anomalous way that, while different from the original model, is still easily distinguishable from the retrained model by the MIA attacker. This suggests that for larger-scale unlearning, a more moderate attenuation is preferable.
    
    \item \textbf{Utility-Efficacy Trade-off on Frcsub:} On the Frcsub dataset, the behavior of $\lambda$ is different again. Increasing $\lambda$ consistently improves unlearning efficacy (MIA scores drop) but at the cost of a noticeable decrease in model utility. This highlights a classic trade-off that is more pronounced on this specific dataset.
\end{itemize}

In conclusion, our comprehensive sensitivity analysis demonstrates that while the effects of $\alpha$ and $\lambda$ can be context-dependent, the benefit of our core hierarchical smoothing mechanism, controlled by $\beta$, is universal and robust. It consistently provides a significant boost to unlearning performance, substantiating its role as the key innovation of our algorithm.
\begin{figure*}[ht]
		\centering
		\begin{minipage}[t]{0.31\textwidth}
			\centering
			\includegraphics[width=5.5cm]{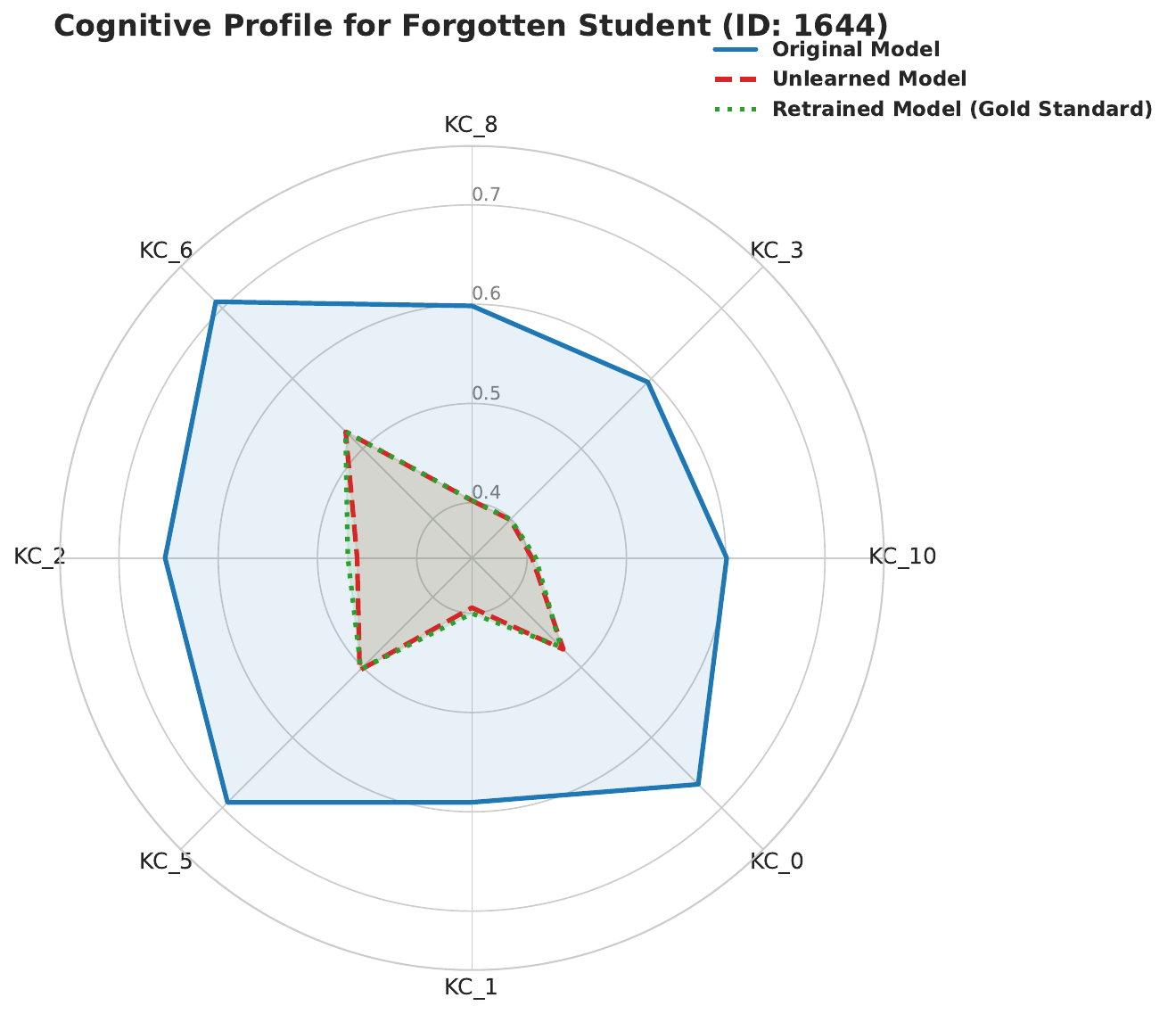}
			\subcaption{Student ID: 1644}
		\end{minipage}
		\begin{minipage}[t]{0.31\textwidth}
			\centering
			\includegraphics[width=5.5cm]{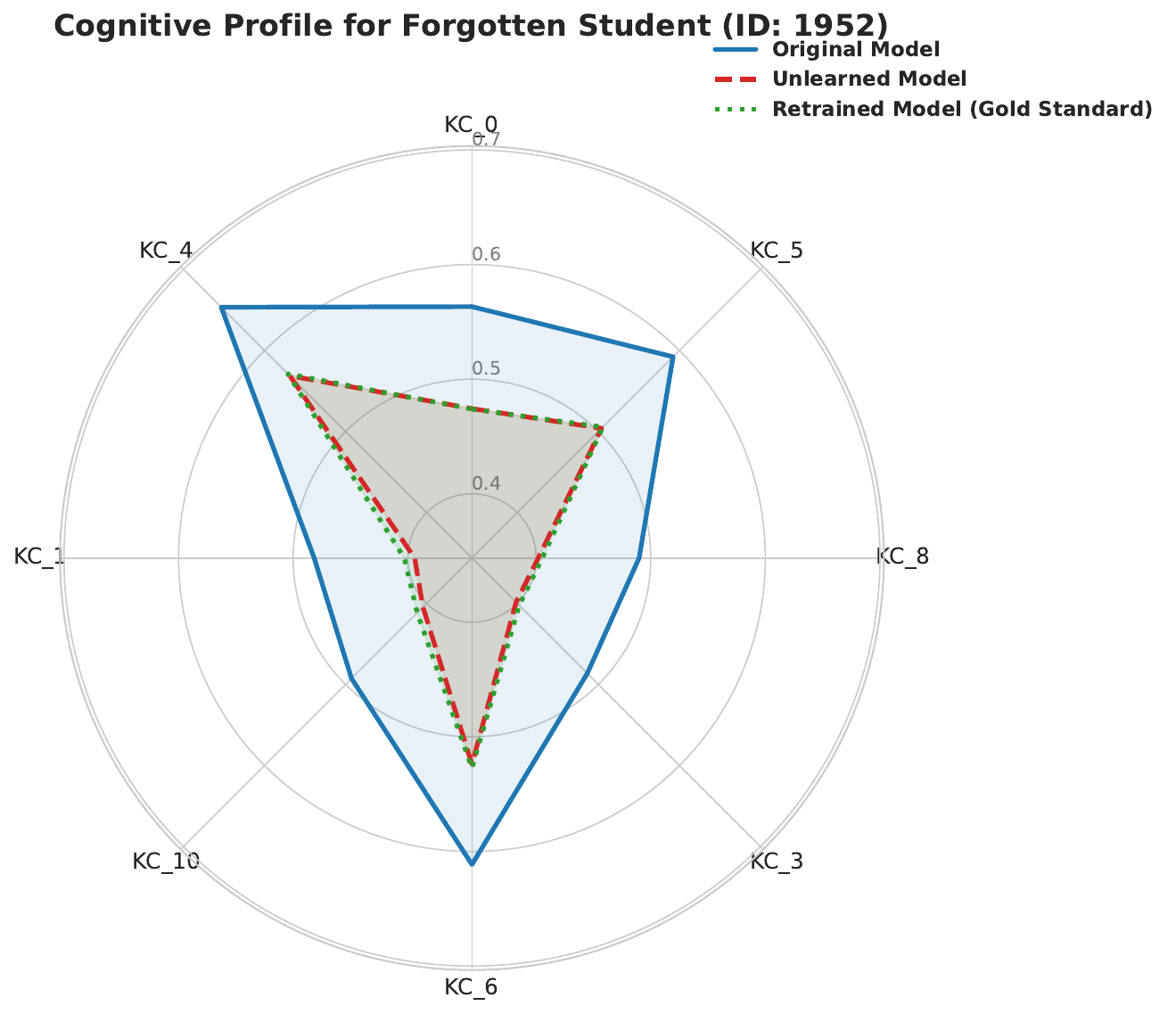}
			\subcaption{Student ID: 1952}
		\end{minipage}
		\begin{minipage}[t]{0.31\textwidth}
			\centering
			\includegraphics[width=5.5cm]{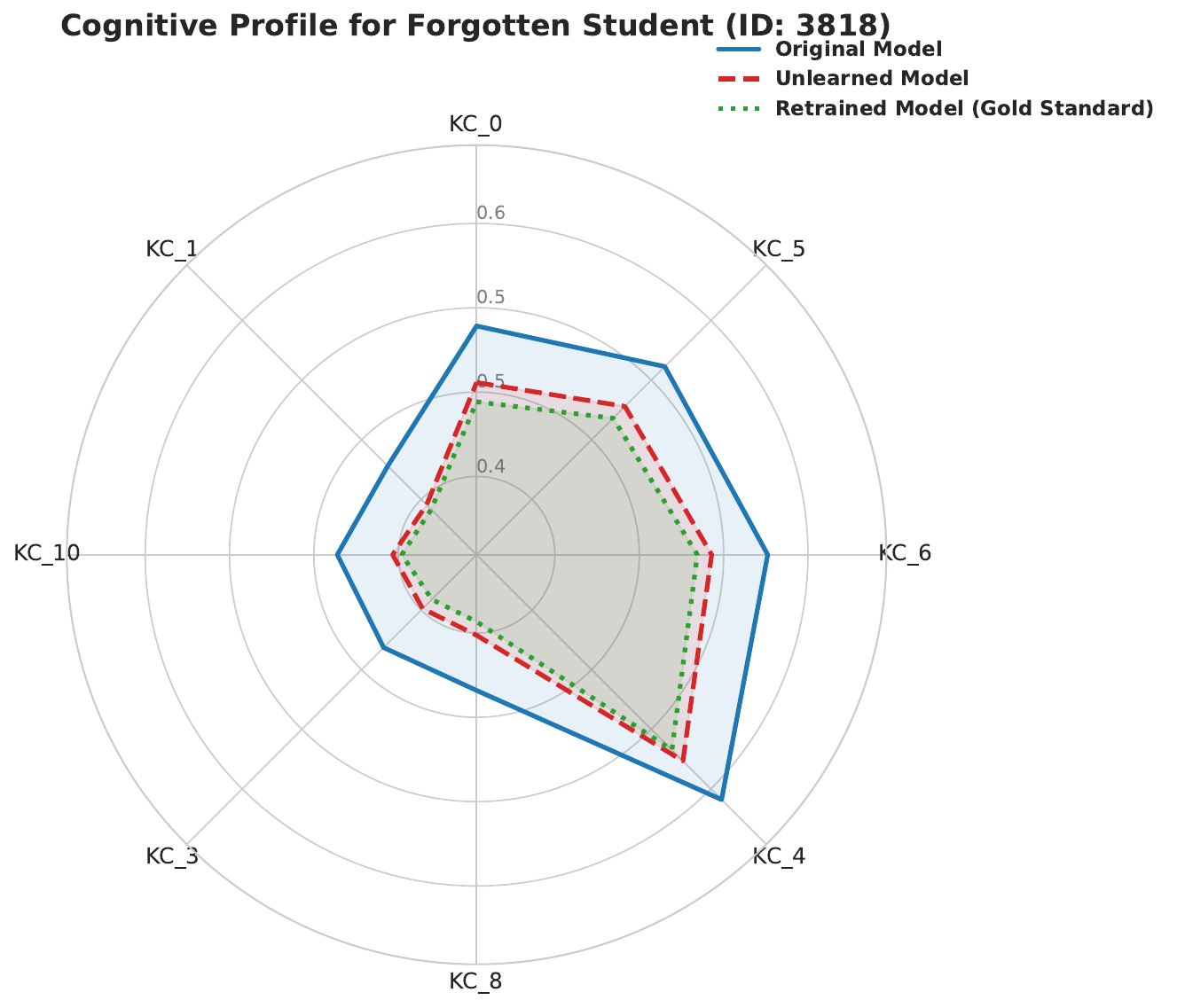}
			\subcaption{Student ID: 3818}
	   \end{minipage}
		\caption{Case study of cognitive profiles for three randomly selected forgotten students from the Math2 dataset (10\% unlearning ratio).}
		\label{qa_math1_01}
\end{figure*}
\begin{figure*}[ht]
		\centering
		\begin{minipage}[t]{0.31\textwidth}
			\centering
			\includegraphics[width=5.5cm]{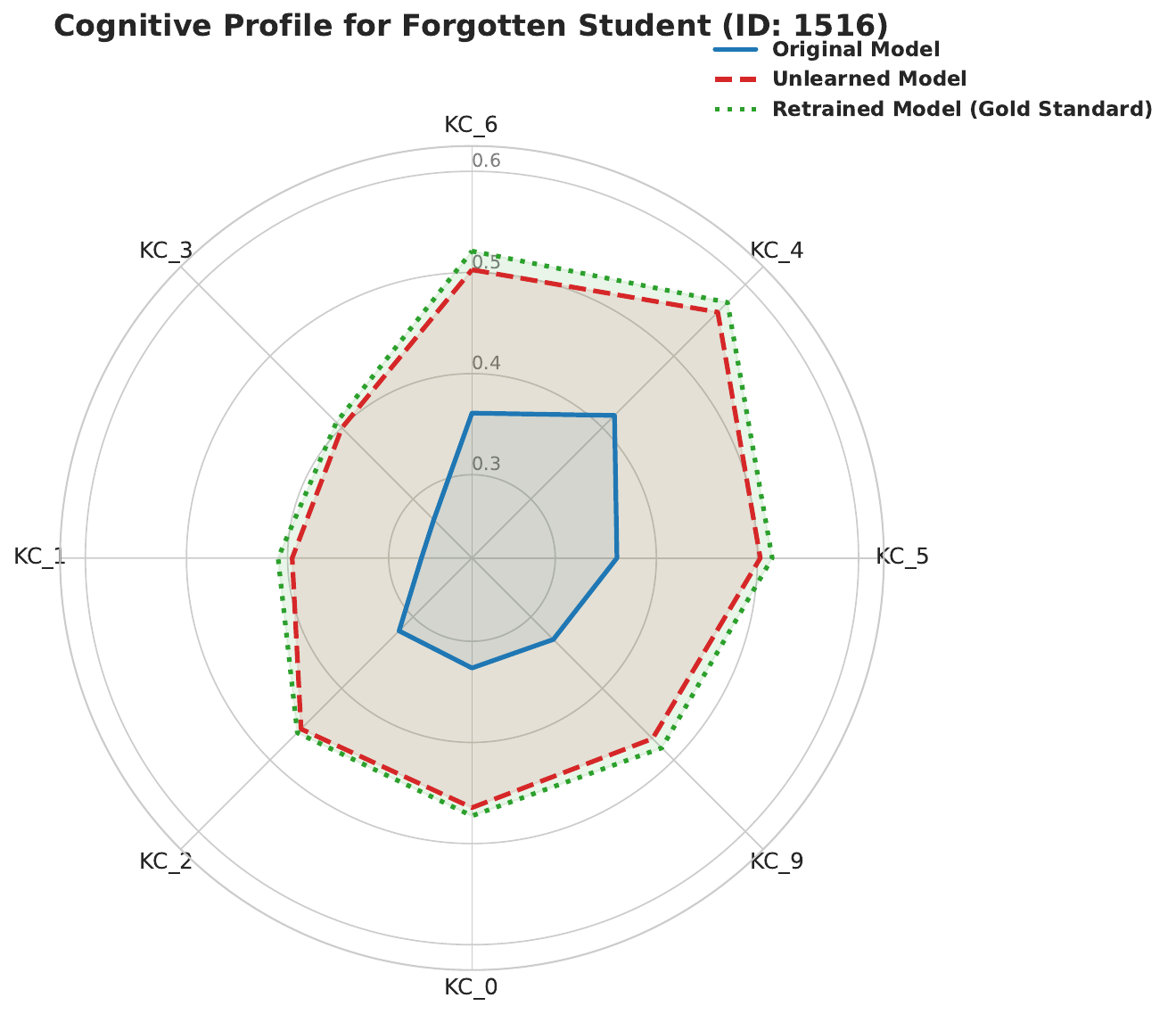}
			\subcaption{Student ID: 1516}
		\end{minipage}
		\begin{minipage}[t]{0.31\textwidth}
			\centering
			\includegraphics[width=5.5cm]{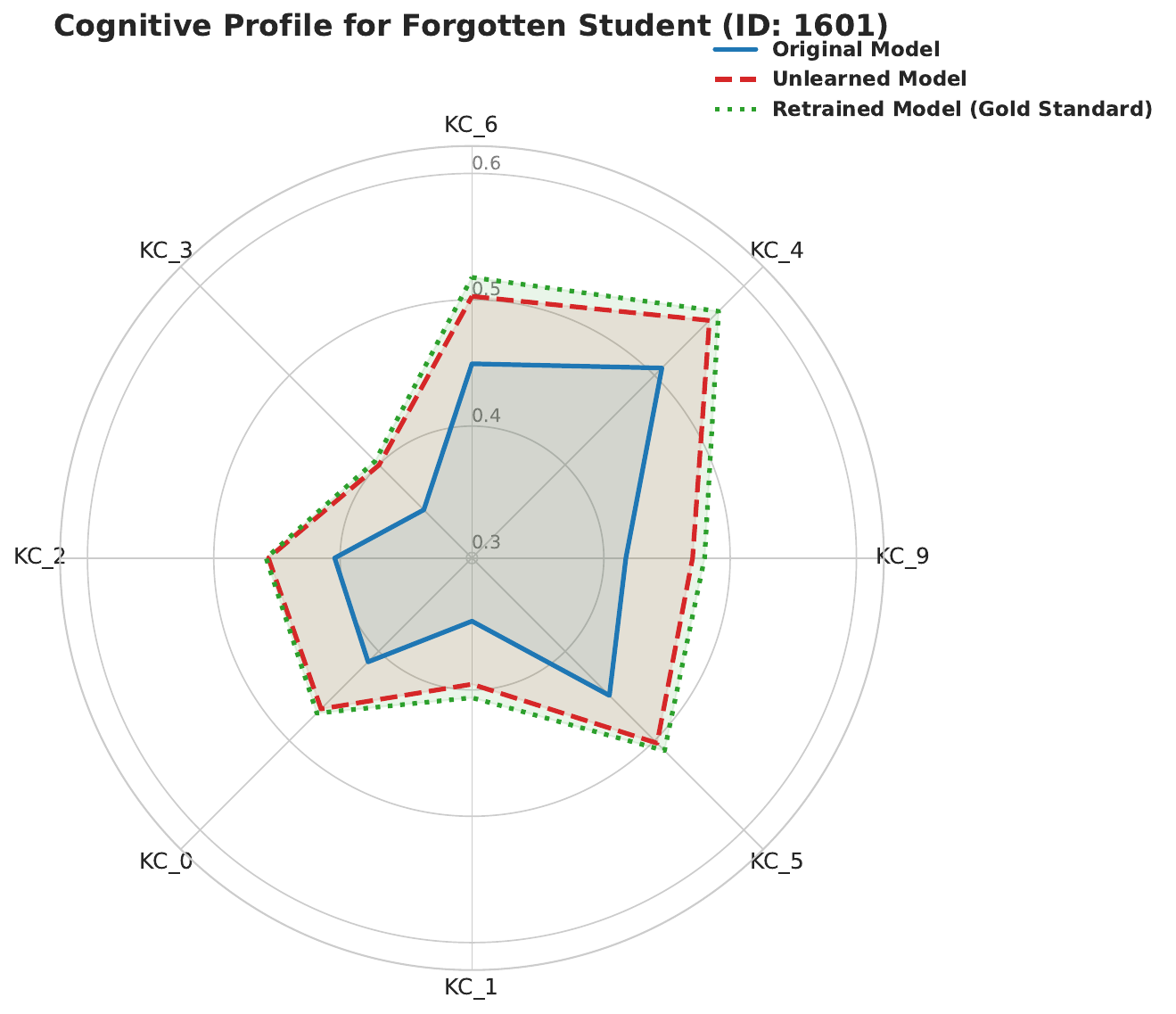}
			\subcaption{Student ID: 1601}
		\end{minipage}
		\begin{minipage}[t]{0.31\textwidth}
			\centering
			\includegraphics[width=5.5cm]{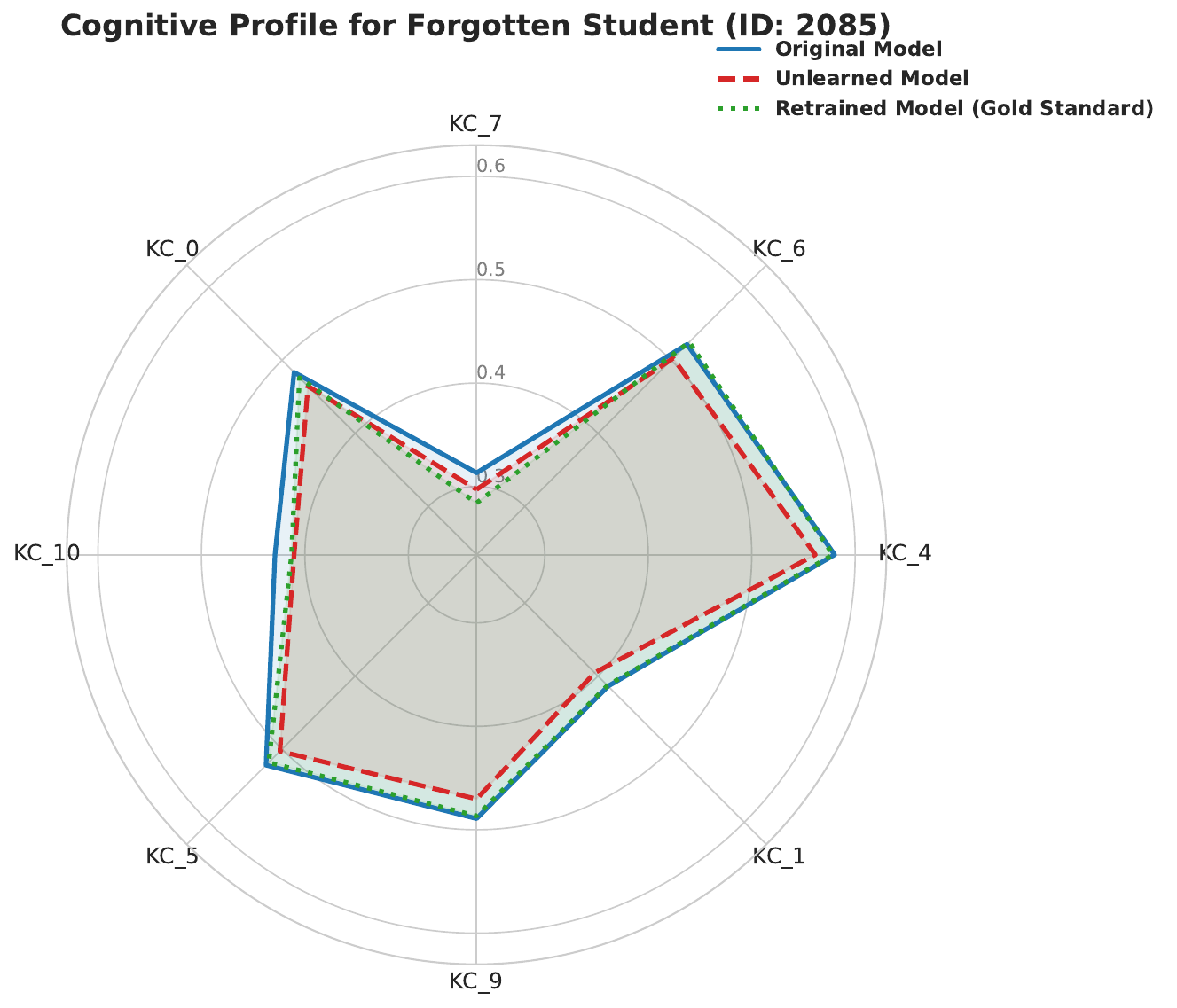}
			\subcaption{Student ID: 2085}
	   \end{minipage}
		\caption{Case study of cognitive profiles for three randomly selected forgotten students from the Math2 dataset (5\% unlearning ratio).}
		\label{qa_math1_005}
\end{figure*}
\begin{figure*}[ht]
		\centering
		\begin{minipage}[t]{0.31\textwidth}
			\centering
			\includegraphics[width=5.5cm]{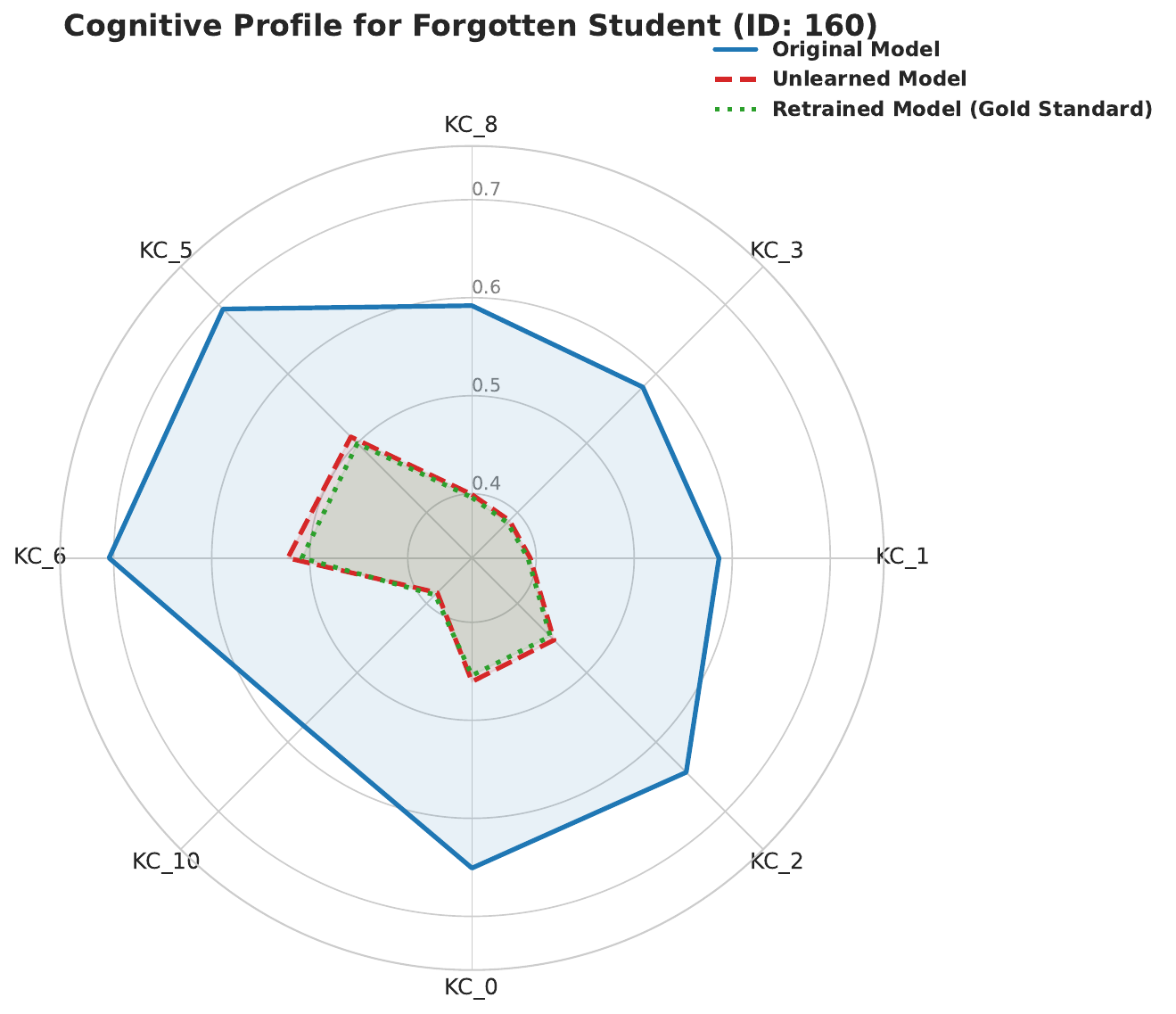}
			\subcaption{Student ID: 160}
		\end{minipage}
		\begin{minipage}[t]{0.31\textwidth}
			\centering
			\includegraphics[width=5.5cm]{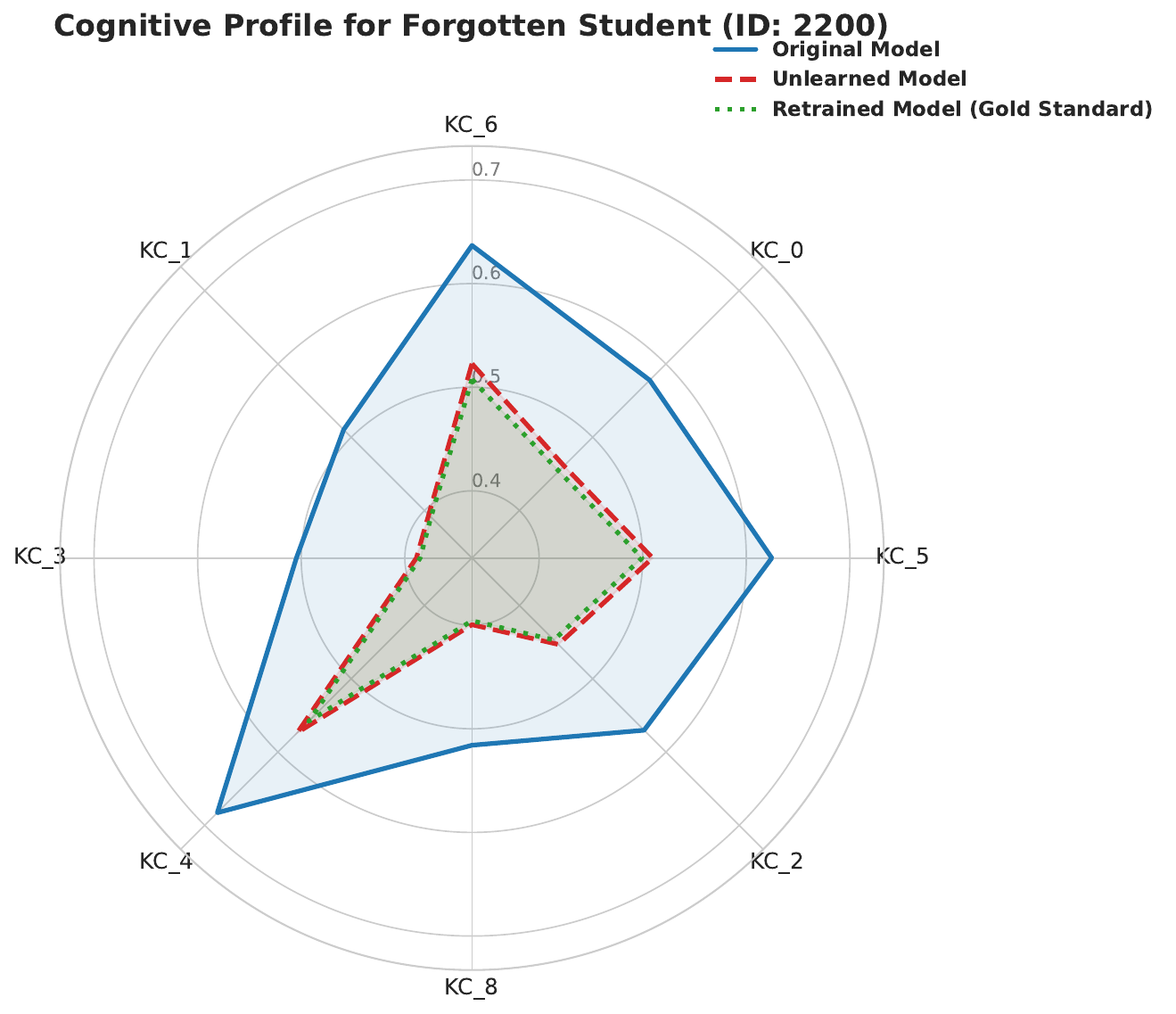}
			\subcaption{Student ID: 2200}
		\end{minipage}
		\begin{minipage}[t]{0.31\textwidth}
			\centering
			\includegraphics[width=5.5cm]{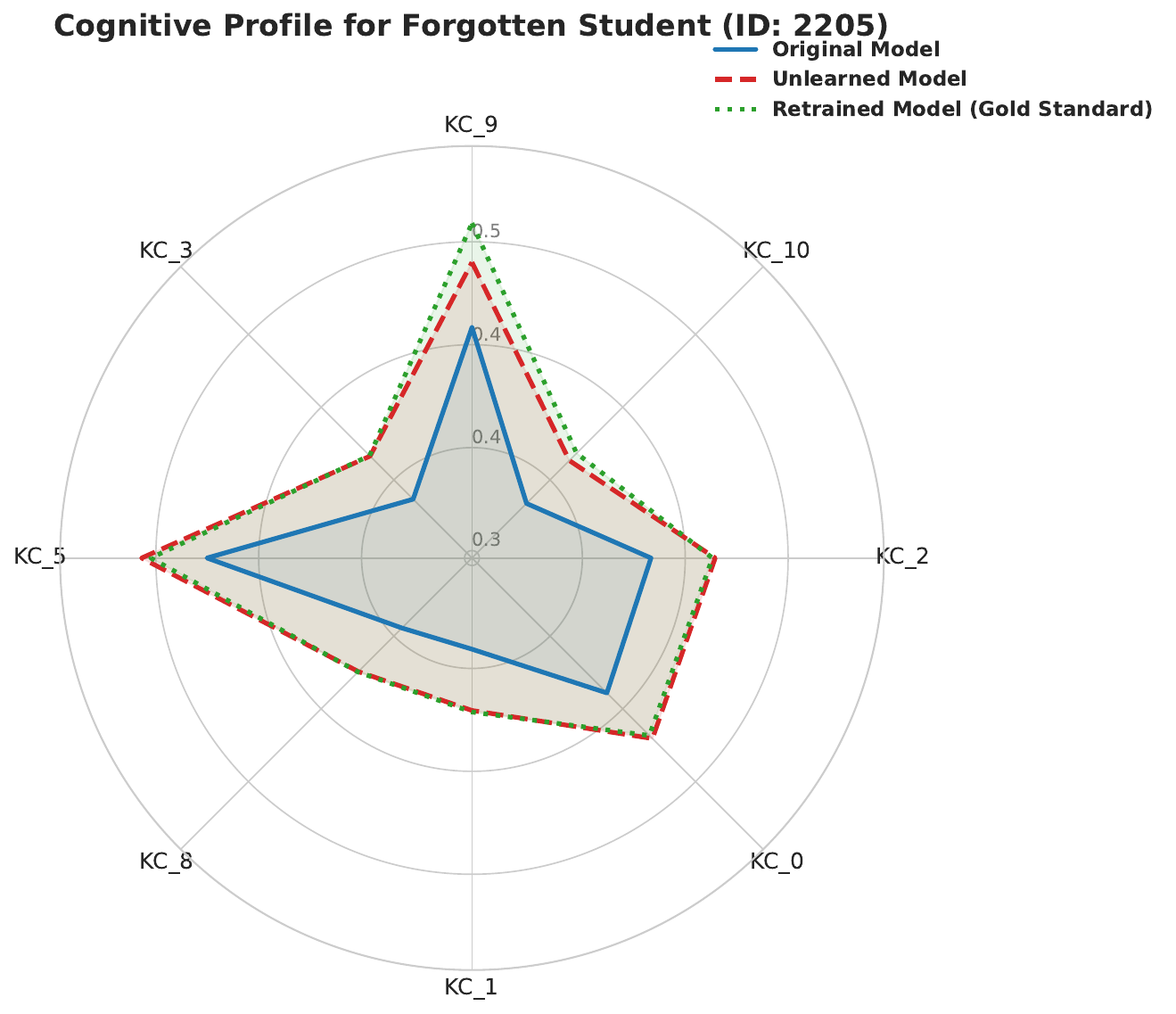}
			\subcaption{Student ID: 2205}
	   \end{minipage}
		\caption{Case study of cognitive profiles for three randomly selected forgotten students from the Math2 dataset (1\% unlearning ratio).}
		\label{qa_math1_001}
\end{figure*}

\begin{figure*}[ht]
		\centering
		\begin{minipage}[t]{0.31\textwidth}
			\centering
			\includegraphics[width=5.5cm]{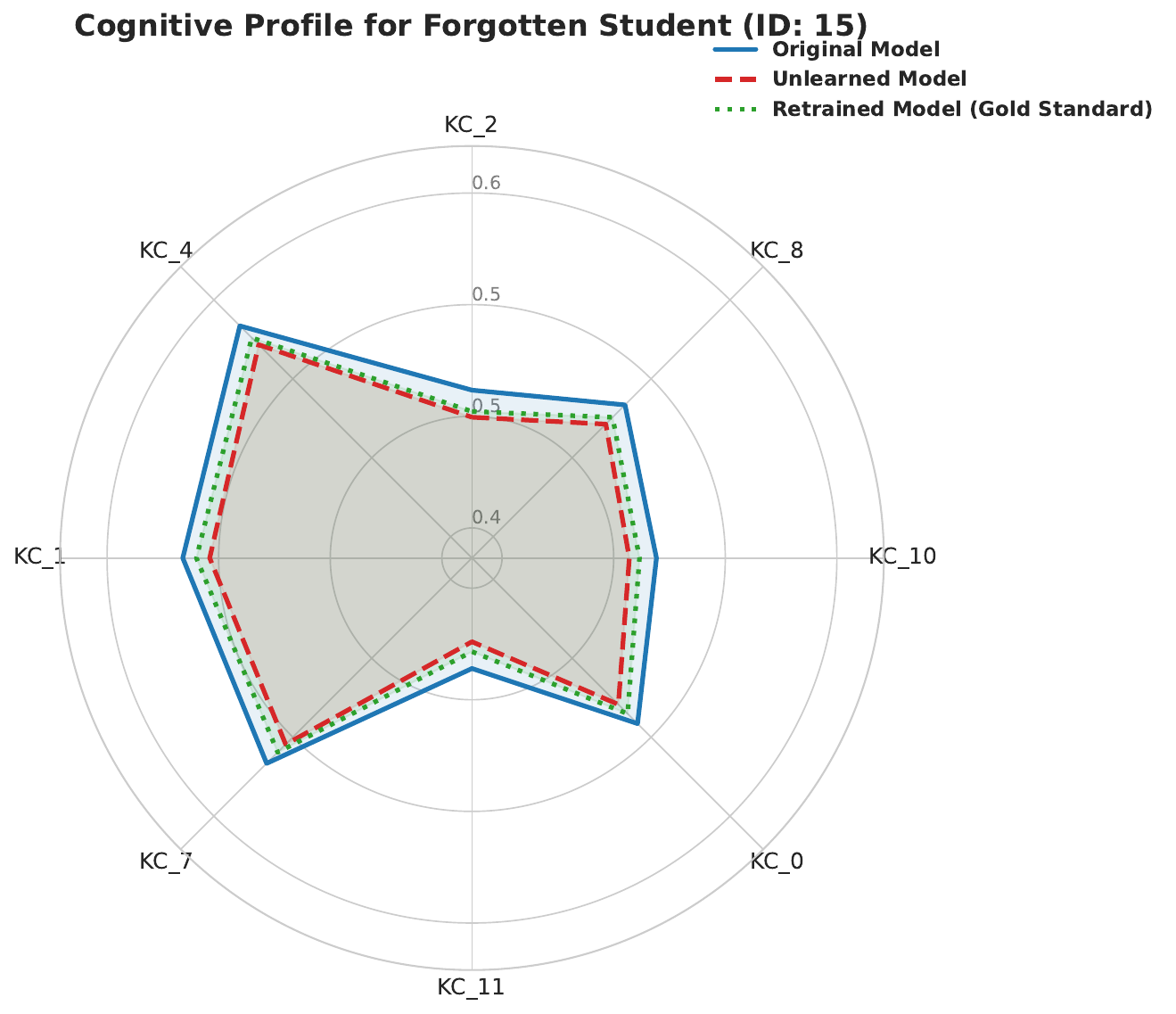}
			\subcaption{Student ID: 15}
		\end{minipage}
		\begin{minipage}[t]{0.31\textwidth}
			\centering
			\includegraphics[width=5.5cm]{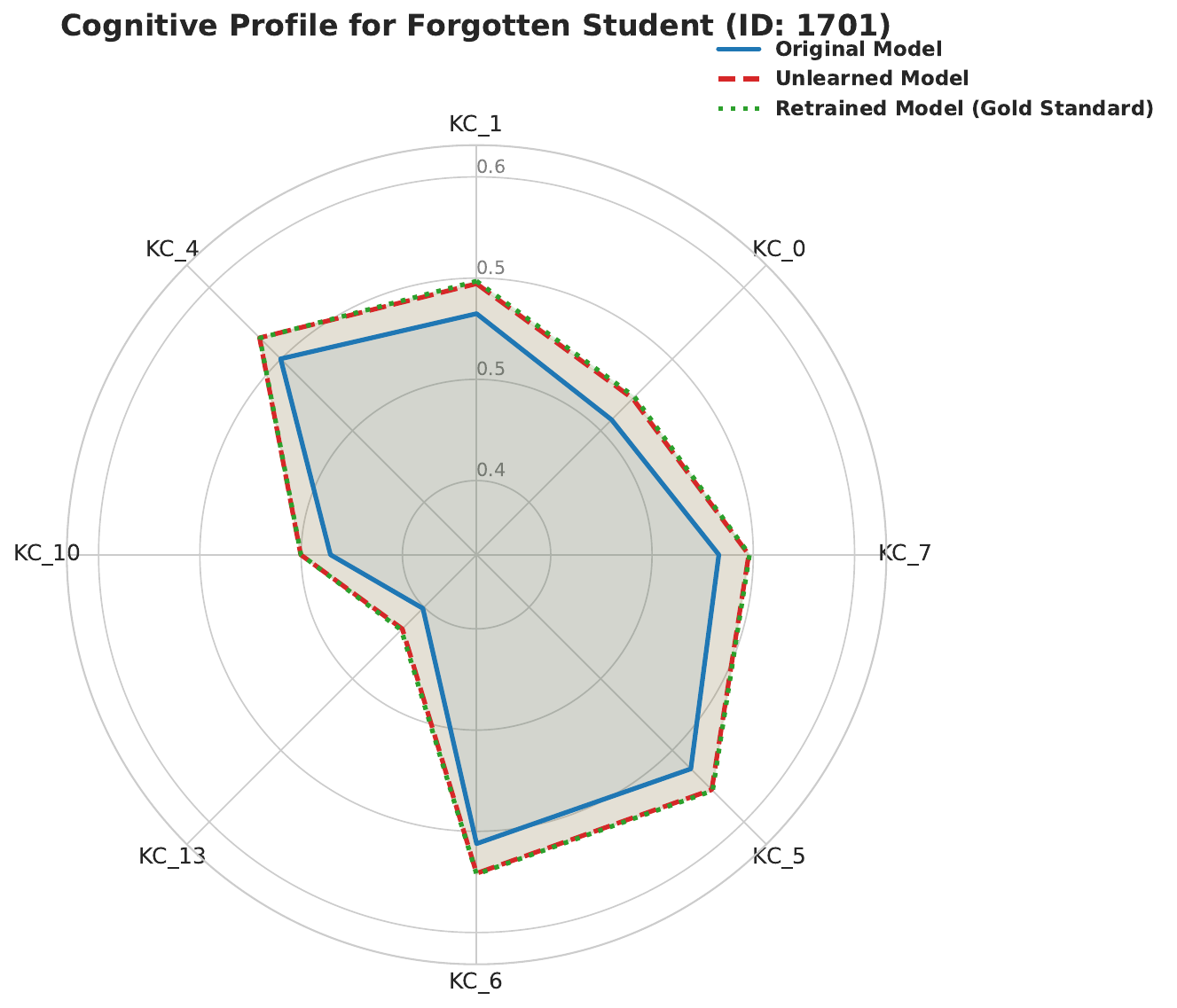}
			\subcaption{Student ID: 1701}
		\end{minipage}
		\begin{minipage}[t]{0.31\textwidth}
			\centering
			\includegraphics[width=5.5cm]{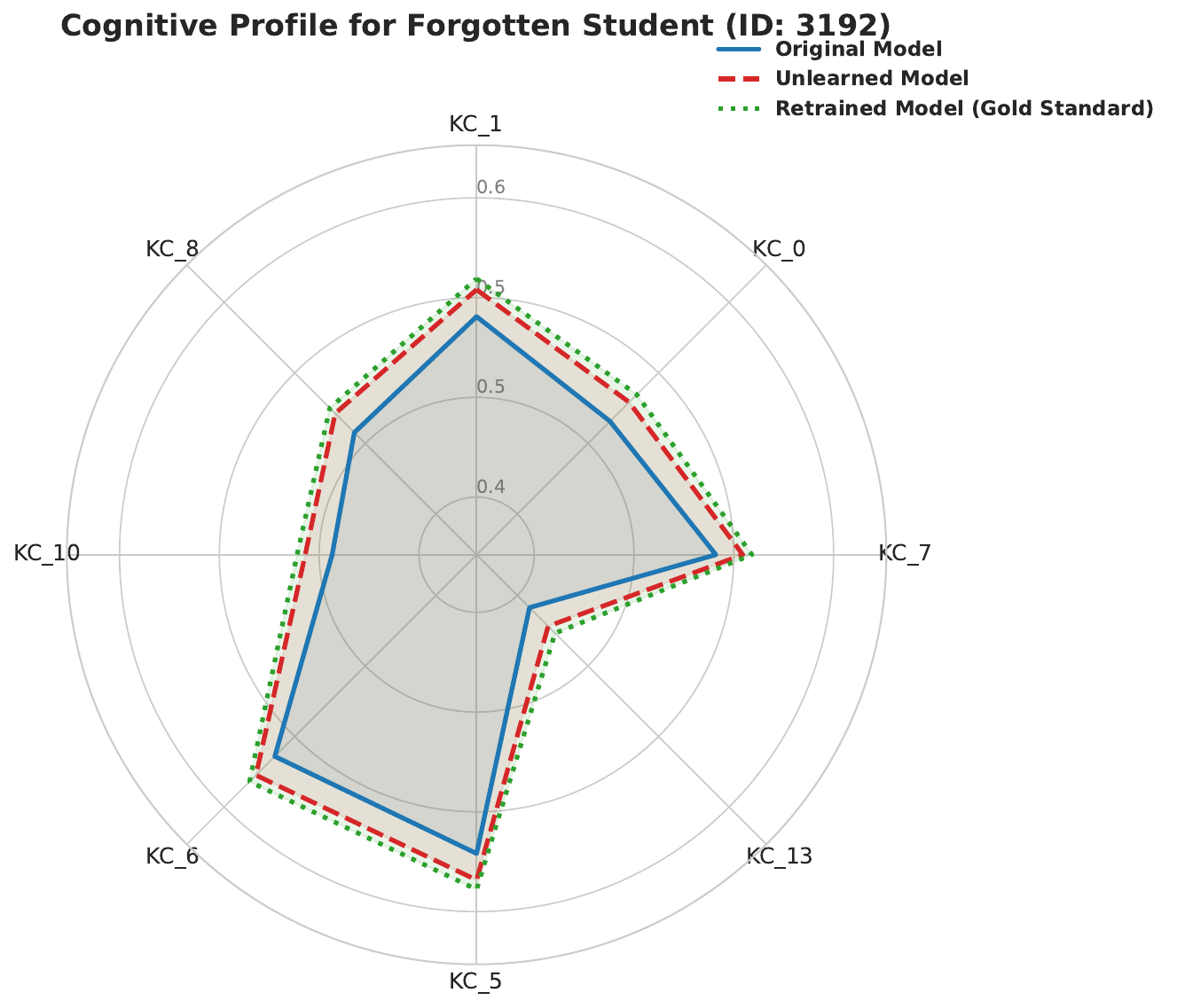}
			\subcaption{Student ID: 3192}
	   \end{minipage}
		\caption{Case study of cognitive profiles for three randomly selected forgotten students from the Math2 dataset (10\% unlearning ratio).}
		\label{qa_math2_01}
\end{figure*}
\begin{figure*}[ht]
		\centering
		\begin{minipage}[t]{0.31\textwidth}
			\centering
			\includegraphics[width=5.5cm]{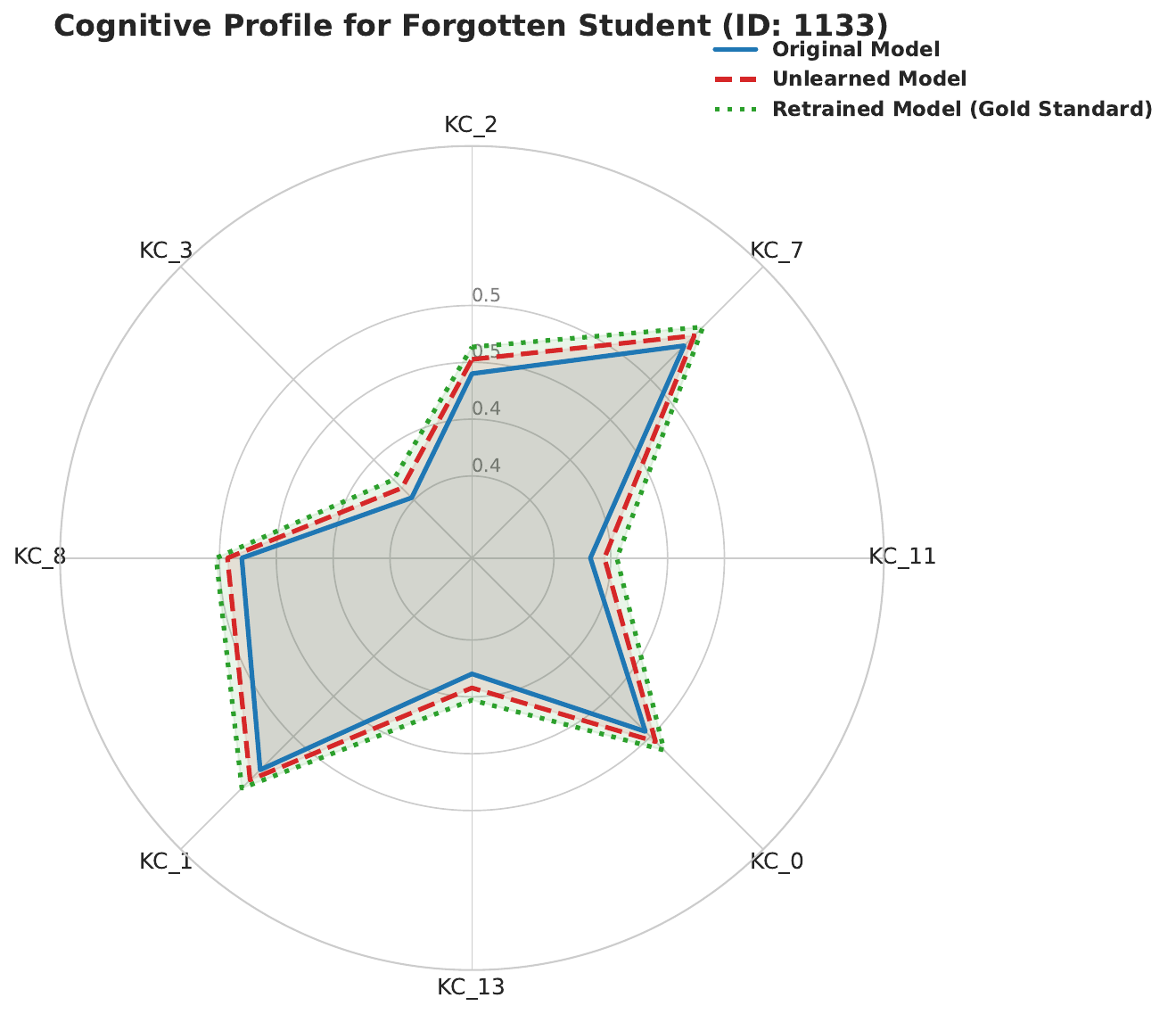}
			\subcaption{Student ID: 1133}
		\end{minipage}
		\begin{minipage}[t]{0.31\textwidth}
			\centering
			\includegraphics[width=5.5cm]{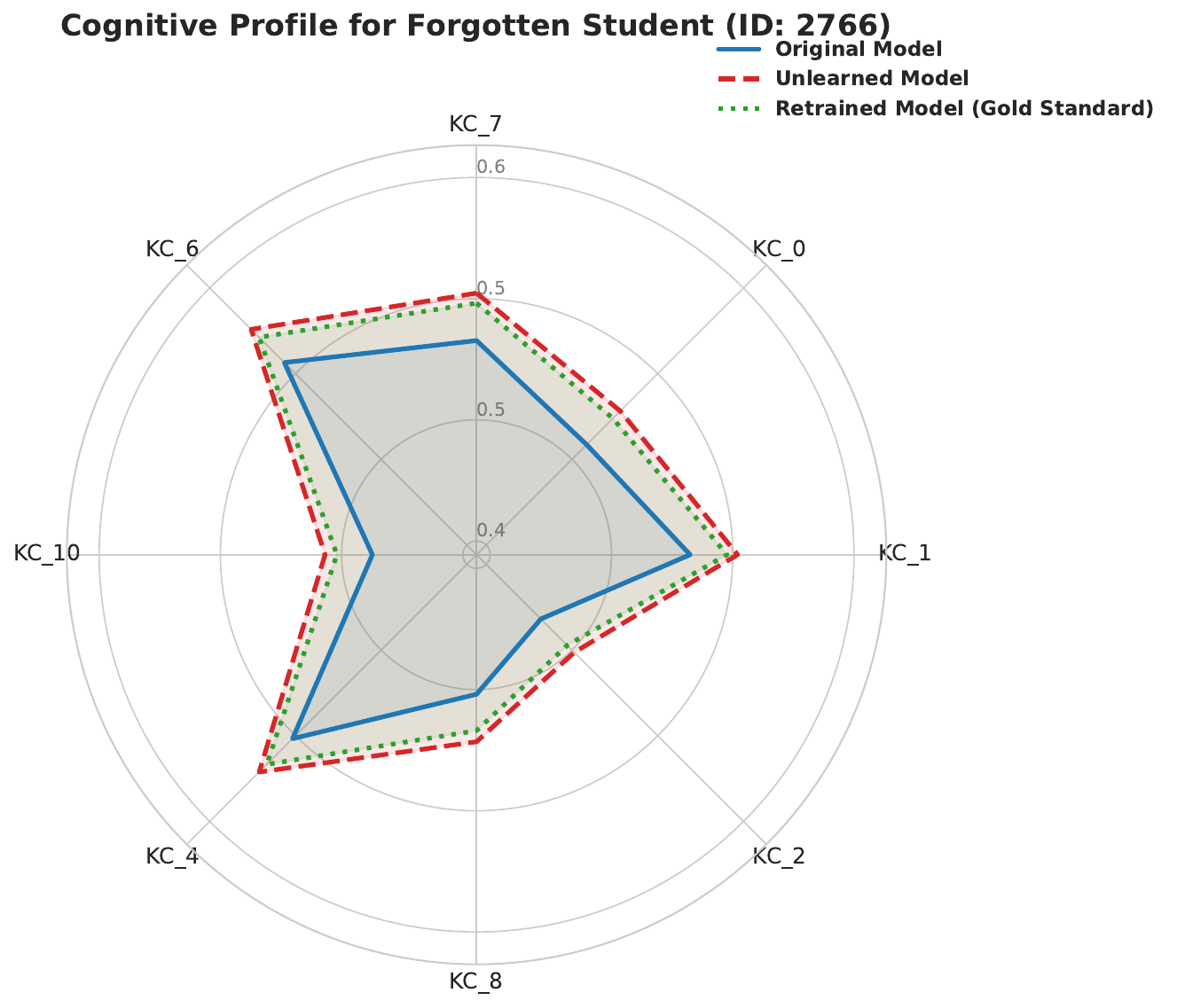}
			\subcaption{Student ID: 2766}
		\end{minipage}
		\begin{minipage}[t]{0.31\textwidth}
			\centering
			\includegraphics[width=5.5cm]{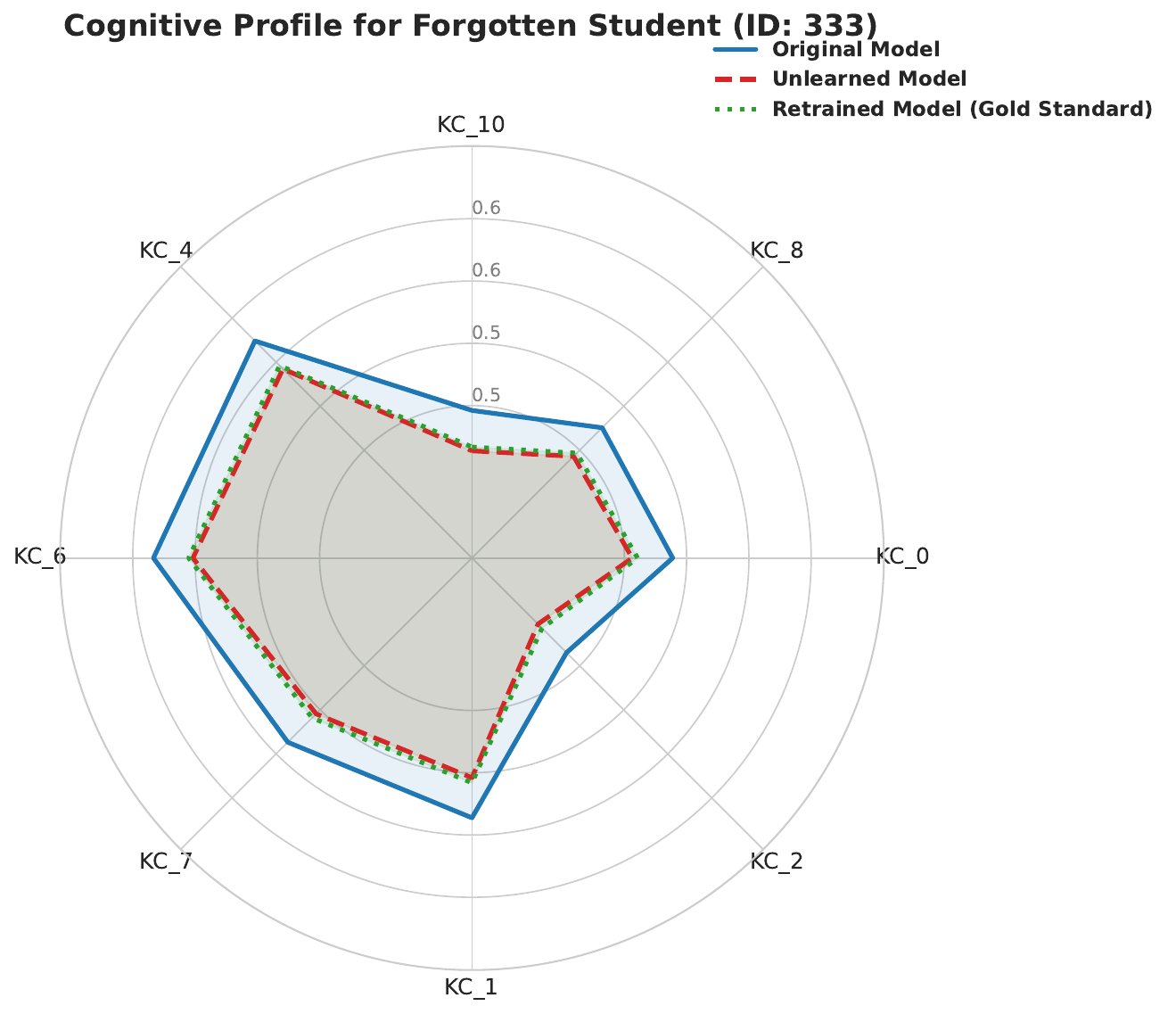}
			\subcaption{Student ID: 333}
	   \end{minipage}
		\caption{Case study of cognitive profiles for three randomly selected forgotten students from the Math2 dataset (1\% unlearning ratio).}
		\label{qa_math2_001}
\end{figure*}

\begin{figure*}[ht]
		\centering
		\begin{minipage}[t]{0.31\textwidth}
			\centering
			\includegraphics[width=5.5cm]{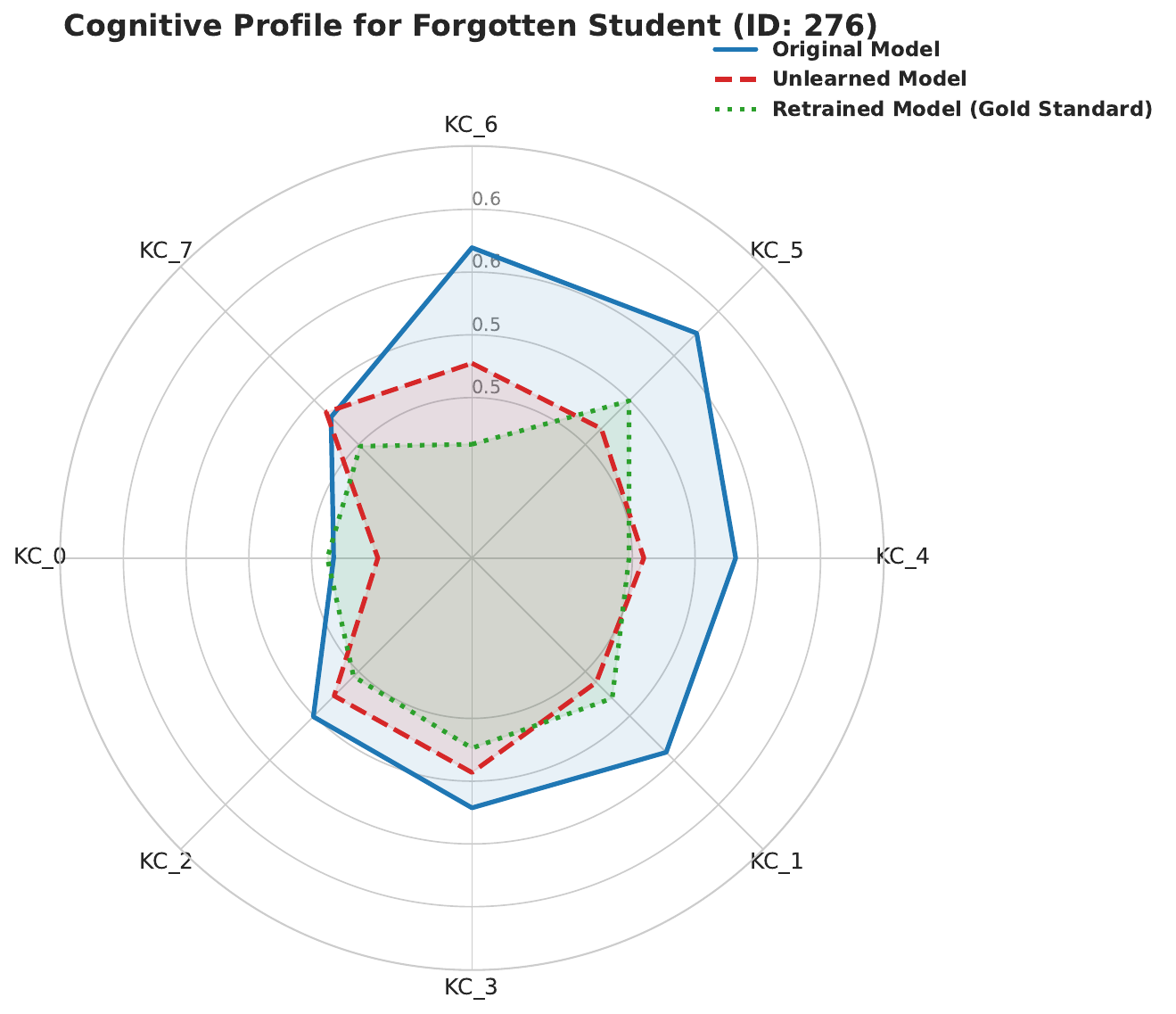}
			\subcaption{Student ID: 276}
		\end{minipage}
		\begin{minipage}[t]{0.31\textwidth}
			\centering
			\includegraphics[width=5.5cm]{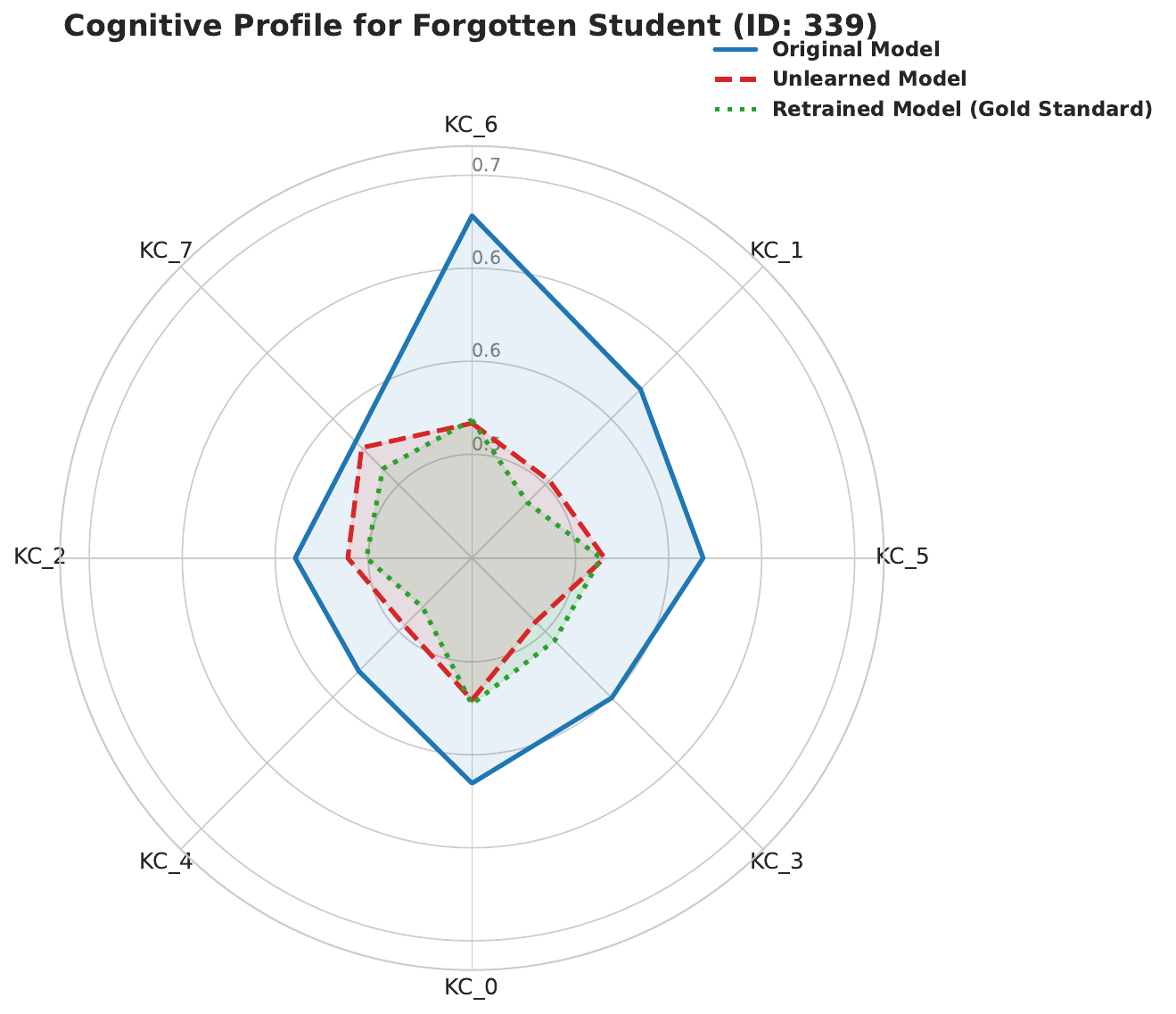}
			\subcaption{Student ID: 339}
		\end{minipage}
		\begin{minipage}[t]{0.31\textwidth}
			\centering
			\includegraphics[width=5.5cm]{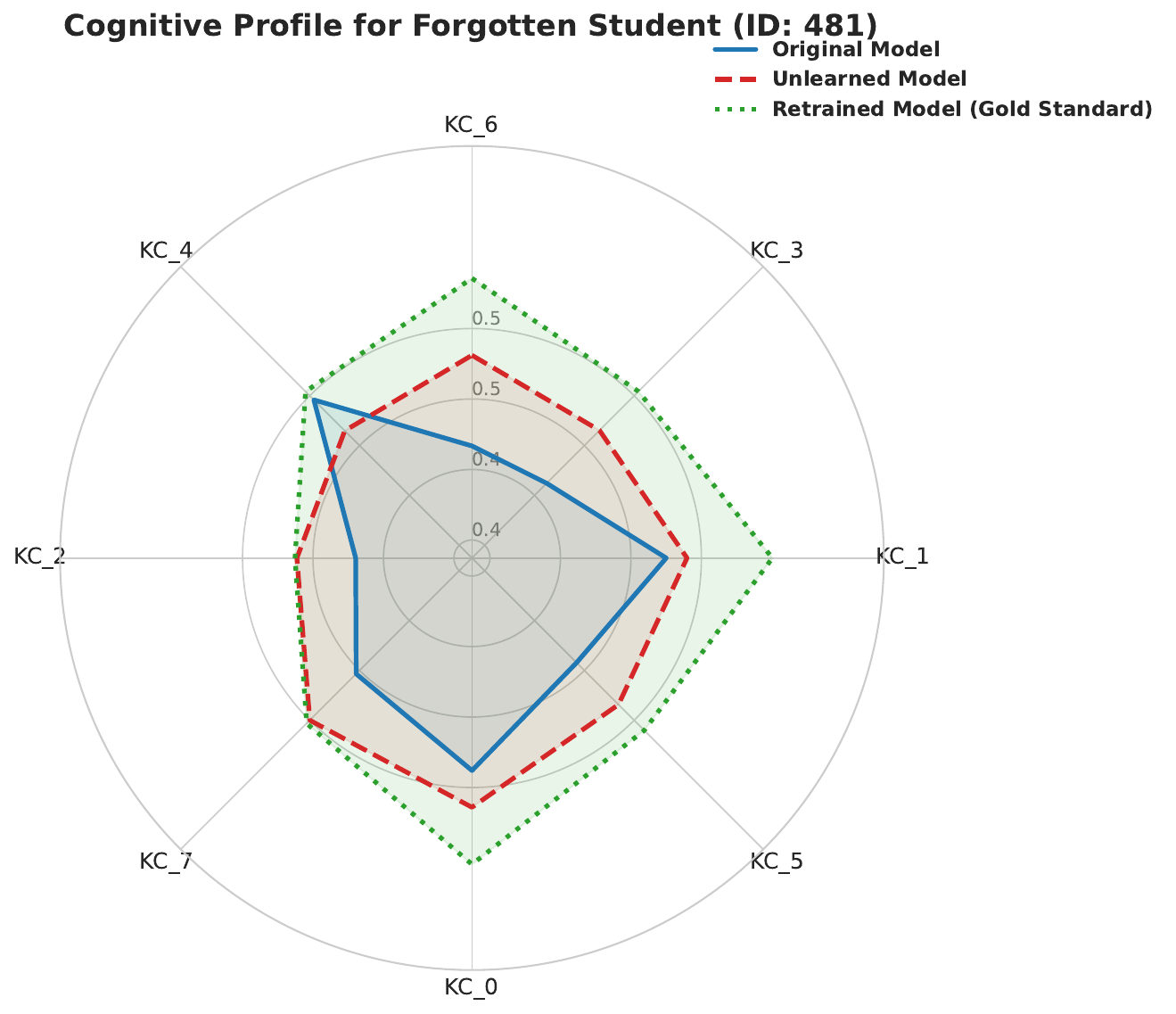}
			\subcaption{Student ID: 481}
	   \end{minipage}
		\caption{Case study of cognitive profiles for three randomly selected forgotten students from the Frcsub dataset (10\% unlearning ratio).}
		\label{qa_frcsub_01}
\end{figure*}
\begin{figure*}[ht]
		\centering
		\begin{minipage}[t]{0.31\textwidth}
			\centering
			\includegraphics[width=5.5cm]{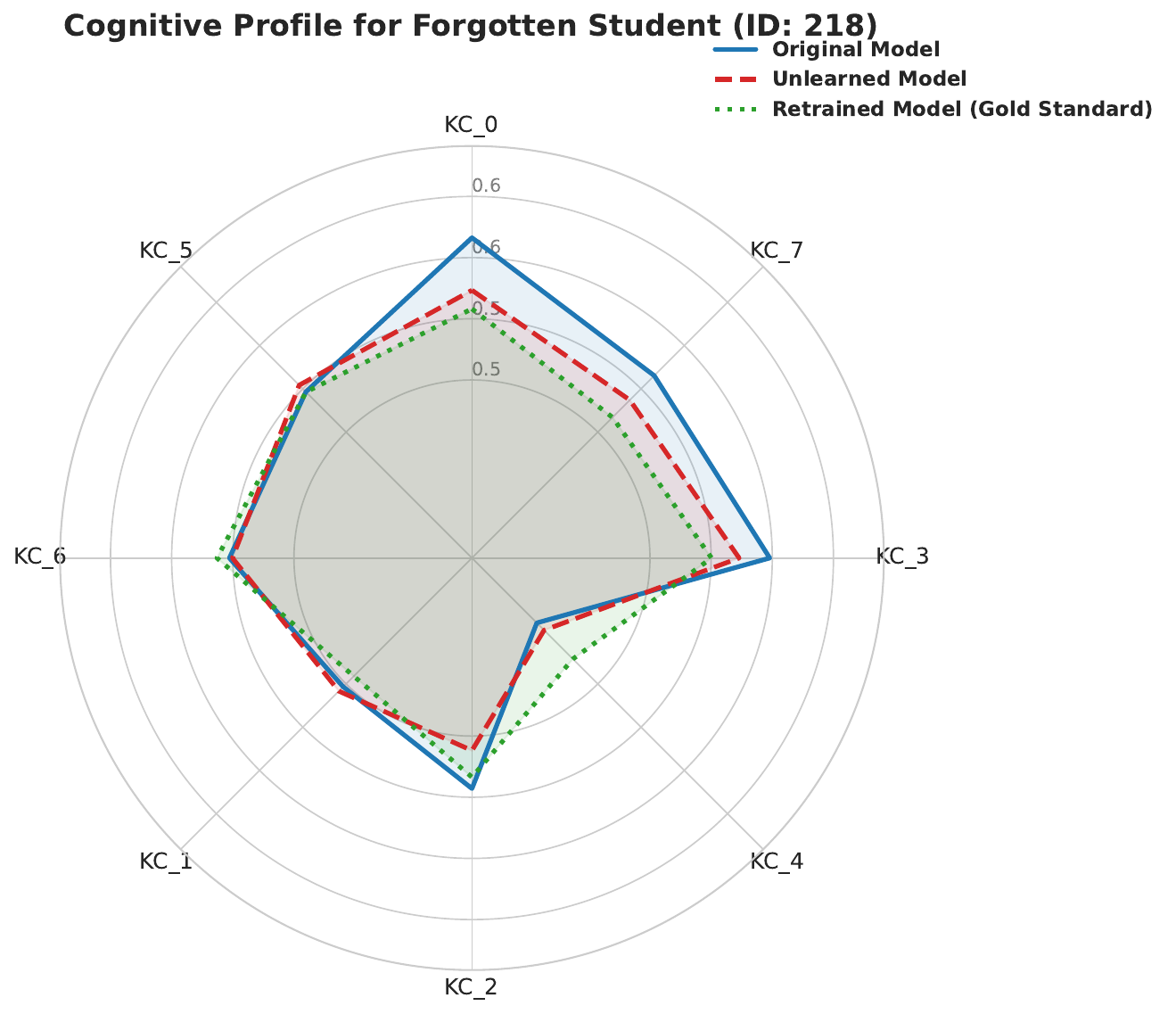}
			\subcaption{Student ID: 218}
		\end{minipage}
		\begin{minipage}[t]{0.31\textwidth}
			\centering
			\includegraphics[width=5.5cm]{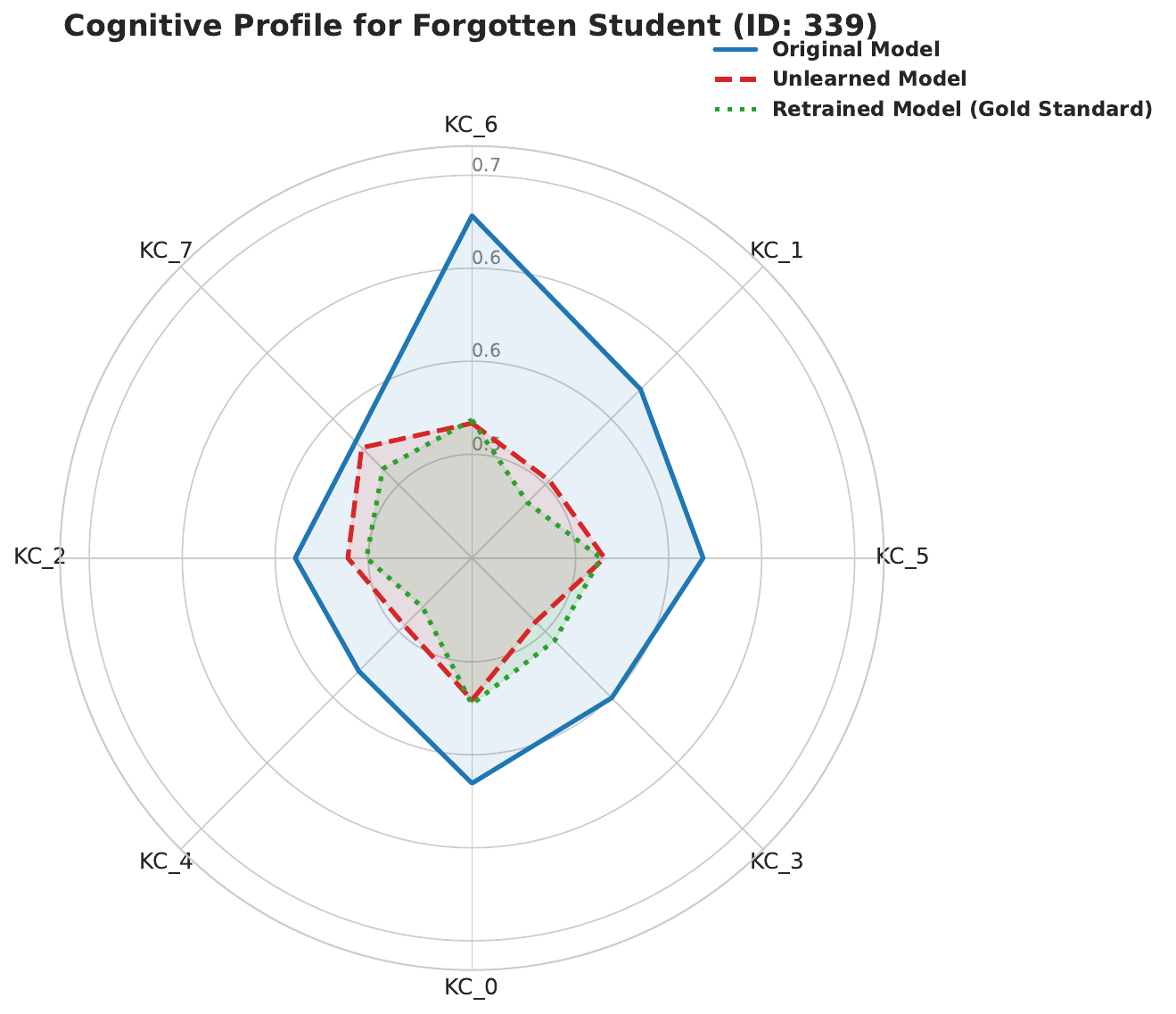}
			\subcaption{Student ID: 339}
		\end{minipage}
		\begin{minipage}[t]{0.31\textwidth}
			\centering
			\includegraphics[width=5.5cm]{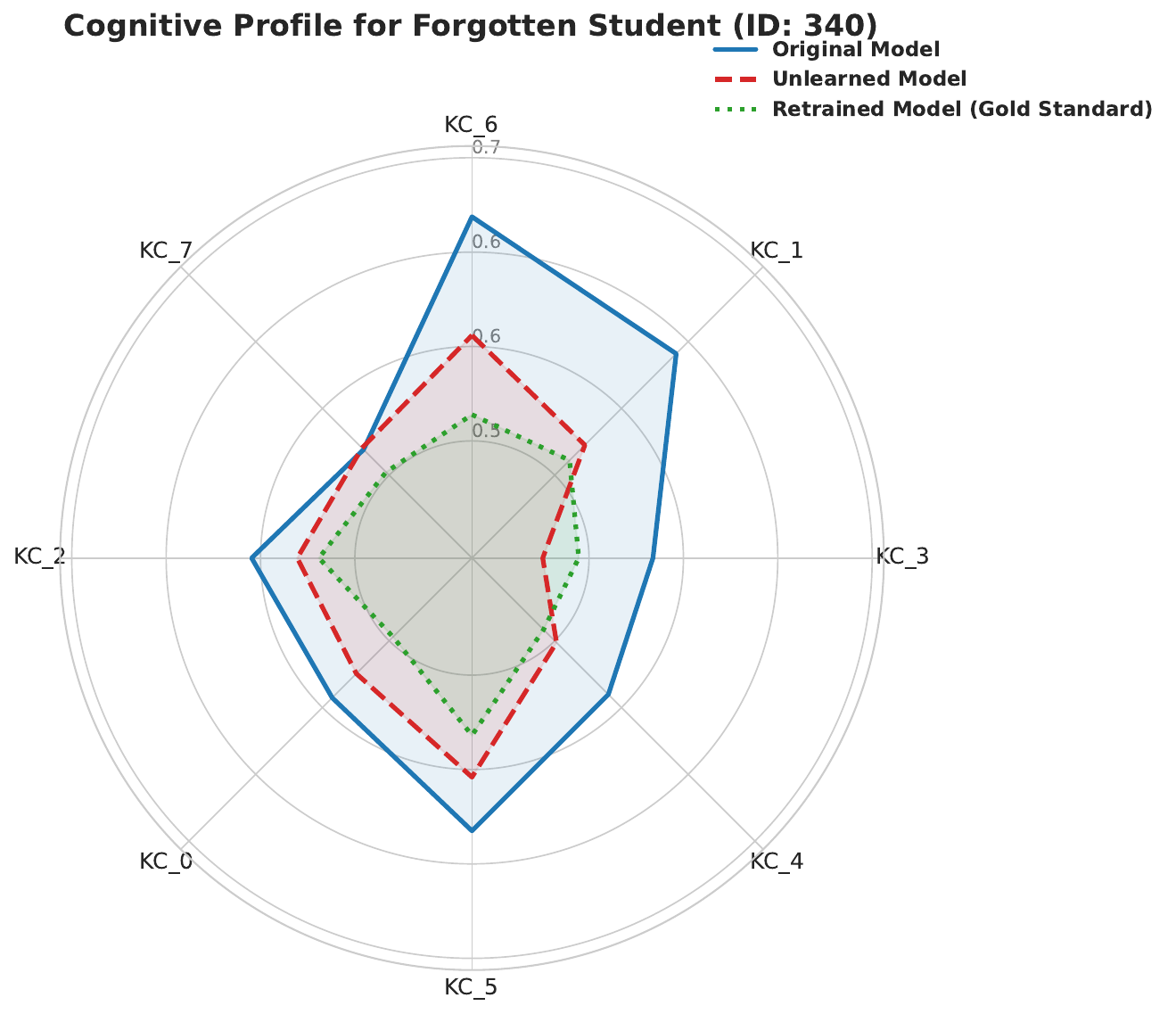}
			\subcaption{Student ID: 340}
	   \end{minipage}
		\caption{Case study of cognitive profiles for three randomly selected forgotten students from the Frcsub dataset (5\% unlearning ratio).}
		\label{qa_frcsub_005}
\end{figure*}
\begin{figure*}[ht]
		\centering
		\begin{minipage}[t]{0.31\textwidth}
			\centering
			\includegraphics[width=5.5cm]{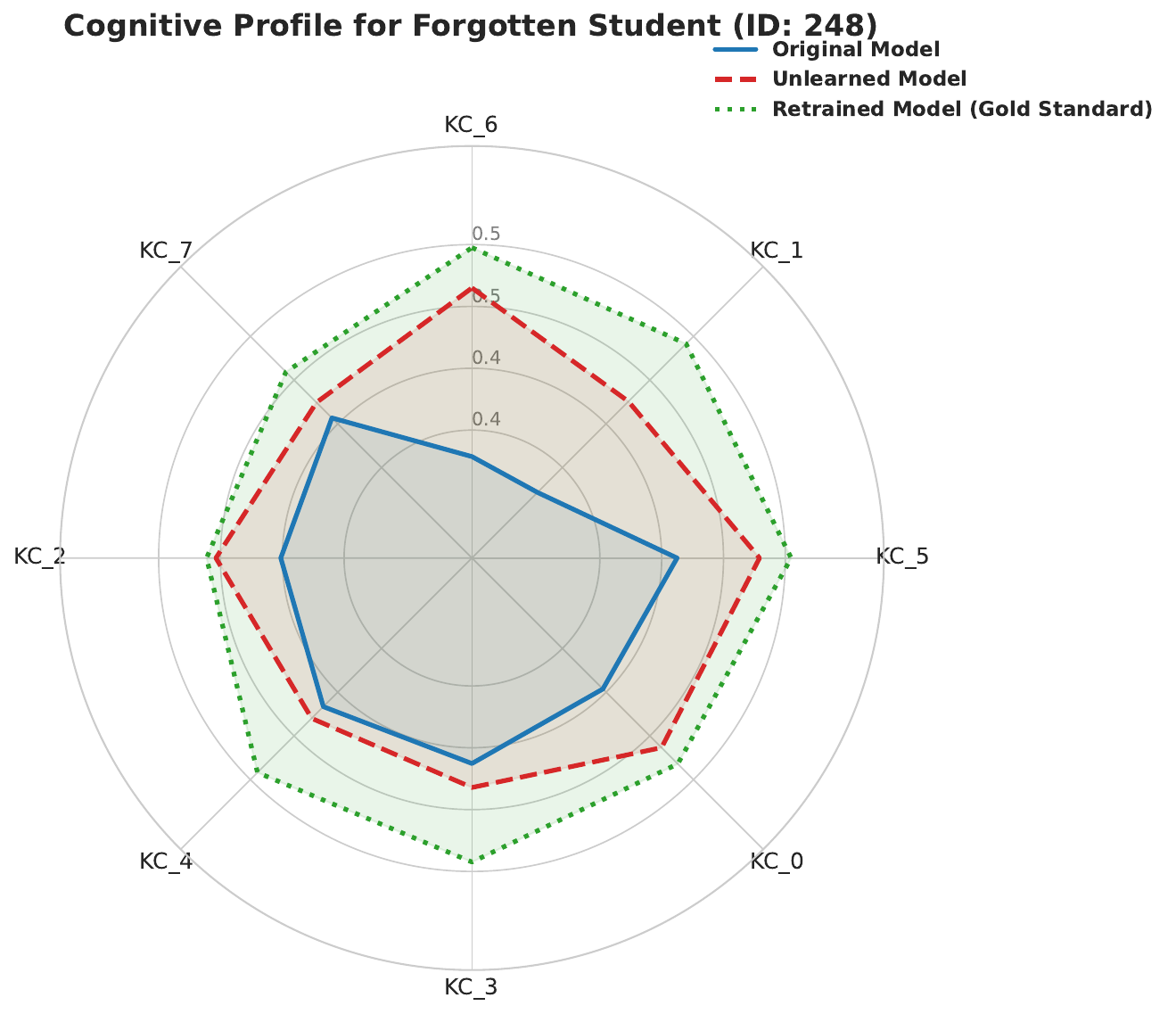}
			\subcaption{Student ID: 248}
		\end{minipage}
		\begin{minipage}[t]{0.31\textwidth}
			\centering
			\includegraphics[width=5.5cm]{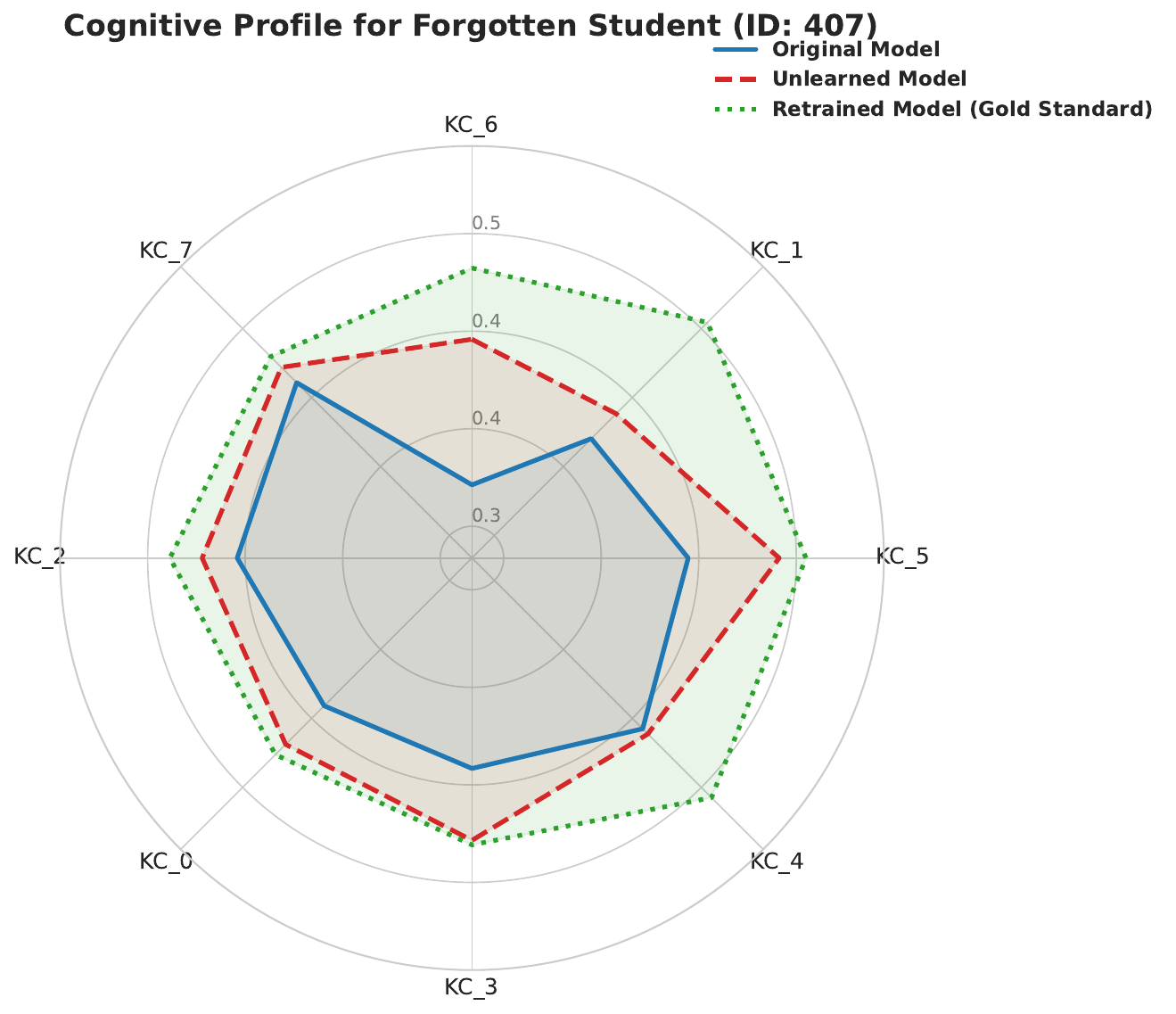}
			\subcaption{Student ID: 407}
		\end{minipage}
		\begin{minipage}[t]{0.31\textwidth}
			\centering
			\includegraphics[width=5.5cm]{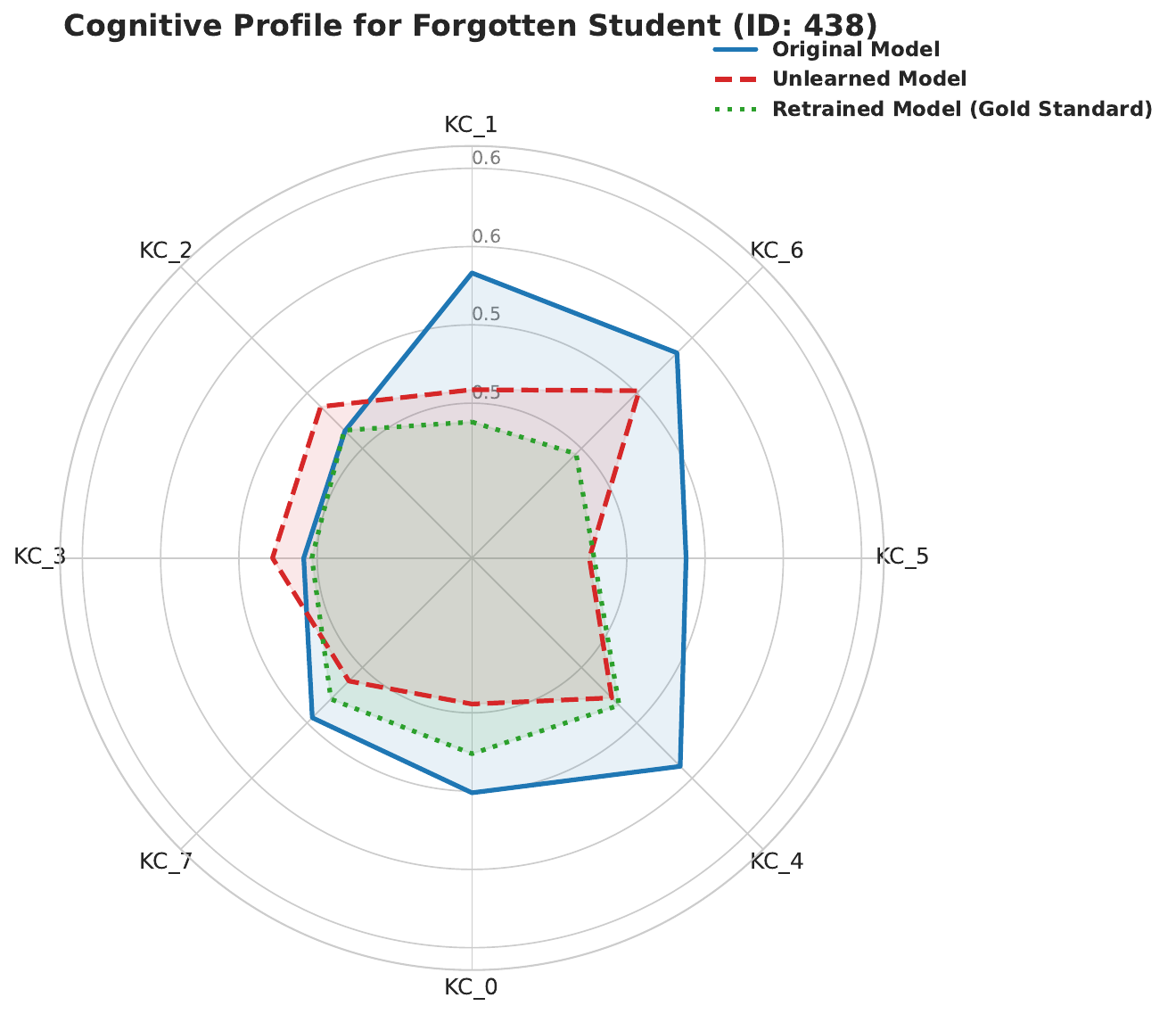}
			\subcaption{Student ID: 438}
	   \end{minipage}
		\caption{Case study of cognitive profiles for three randomly selected forgotten students from the Frcsub dataset (1\% unlearning ratio).}
		\label{qa_frcsub_001}
\end{figure*}
\subsection{Detailed Case Studies of Cognitive Profiles}
This section provides a detailed, case-by-case qualitative analysis of our algorithm's unlearning effectiveness, presenting cognitive profiles for randomly selected forgotten students across all three datasets and unlearning ratios. These visualizations offer intuitive support for our quantitative findings. The complete set of case study figures is presented in Figures~\ref{qa_math1_01} -~\ref{qa_frcsub_001}.

\subsubsection{Analysis on Math1 and Math2 Datasets}The case studies on the Math1 and Math2 datasets consistently and compellingly demonstrate the success of PrivacyCD. Across all tested unlearning ratios (10\%, 5\%, and 1\%), a clear and repeated pattern emerges:
\begin{itemize}
    \item The \textbf{Original Model} (blue, solid line) consistently produces a large, distinct, and often irregular cognitive profile for the forgotten student. This reflects a state of strong ``memory'' or even overfitting to the student's specific response patterns during the initial training.
    \item The \textbf{Retrained Model} (green, dotted line), which serves as the gold standard, diagnoses the same student with a significantly more contracted and conservative profile. This represents the ideal ``forgotten'' state, where the model, having no prior information, provides a less certain, near-average assessment.
    \item Critically, in nearly every case on these two datasets, the profile generated by our \textbf{Unlearned Model} (red, dashed line) after applying PrivacyCD almost perfectly overlaps with or closely traces the profile of the Retrained Model. This visual evidence strongly suggests that PrivacyCD successfully restores the model to a state that is statistically and behaviorally almost identical to one that was never exposed to the student's data.
\end{itemize}

\subsubsection{Analysis on the Frcsub Dataset and Discussion}The results on the Frcsub dataset, while still demonstrating the effectiveness of PrivacyCD, reveal a more nuanced and interesting dynamic.
\begin{itemize}
    \item In several cases (e.g., Student IDs 248, 481, 407), we observe a different pattern: the profile of the \textbf{Retrained Model} (green, dotted line) is noticeably larger and more expansive than that of our \textbf{Unlearned Model} (red, dashed line). At first glance, this might seem like imperfect unlearning.
    \item However, a deeper analysis reveals that this is not a failure but rather a testament to PrivacyCD's ability to achieve a more favorable outcome. The Frcsub dataset is smaller and more focused on a specific domain (fraction subtraction). Retraining on the remaining data after removing a student can sometimes lead the retrained model to become over-confident or biased based on the remaining student population, resulting in a potentially distorted ``uninformative'' profile.
    \item In these scenarios, our PrivacyCD, by precisely modifying the original, more robustly trained model, achieves a state of forgetting that is not only close to the retrained model but is arguably \textbf{more stable and less biased}. The Unlearned Model's profile represents a more plausible ``neutral'' state than the one produced by a full retrain on a smaller, potentially less diverse dataset. For instance, in the case of Student ID 276, the Unlearned Model's profile is a tight, conservative shape, while the Original model's profile is much larger, indicating successful forgetting.
\end{itemize}

In conclusion, the comprehensive qualitative analysis across more than twenty individual cases robustly validates the effectiveness of our PrivacyCD framework. The results on Math1 and Math2 demonstrate PrivacyCD's capability to precisely replicate the gold-standard retrained state. The nuanced results on Frcsub further suggest that in certain contexts, PrivacyCD's surgical modification approach may even lead to a more stable and desirable unlearned state than a full retrain, highlighting the sophistication and practical advantages of our method.
\end{document}